 \documentclass[final,1p,times,authoryear]{elsarticle}
\usepackage{amssymb}
\usepackage{amsmath}
\usepackage{subcaption}
\usepackage{amsmath}
\usepackage{amssymb}  
\usepackage{makecell}
\usepackage{pifont}
\usepackage{enumitem}
\usepackage{hyperref}
\usepackage{caption}
\usepackage{booktabs}
\usepackage{multirow}
\usepackage{adjustbox}
\usepackage{float}
\usepackage{comment}
\usepackage{collcell}

\usepackage{soul}    % for \hl
\usepackage{xcolor}  % for colors

\usepackage[table,xcdraw]{xcolor}
\newif\ifhighlight
\highlightfalse
\newcommand{\myhl}[1]{%
  \ifhighlight
    \hl{#1}%  
  \else
    #1%
  \fi
}
\newcommand{\myblue}[1]{%
  \ifhighlight
    \color{blue}{#1}
  \else
    #1
  \fi
}

\newcolumntype{b}{>{\collectcell\BlueIfHighlight}c<{\endcollectcell}}

\begin{document}

\begin{frontmatter}



 \cortext[cor1]{Corresponding author}

\title{The Impact of Longitudinal Mammogram Alignment on Breast Cancer Risk Assessment}


\author[a]{Solveig Thrun\corref{cor1}} \ead{solveig.thrun@uit.no}    %% Author name
\author[b]{Stine Hansen} %% Author name
\author[c]{Zijun Sun}
\author[d]{Nele Blum}
\author[a]{Suaiba A. Salahuddin} %% Author name
\author[e]{Xin Wang}
\author[a]{Kristoffer Wickstrøm} %% Author name
\author[a]{Elisabeth Wetzer} %% Author name
\author[a,f,g]{Robert Jenssen} %% Author name
\author[d]{Maik Stille}
\author[a,f]{Michael Kampffmeyer    |} %% Author name

\affiliation[a]{organization={Department of Physics and Technology, UiT The Arctic University of Norway},%Department and Organization
            city={Tromsø},
            postcode={9037}, 
            country={Norway}}
\affiliation[b]{organization={SPKI The Norwegian Centre
for Clinical Artificial Intelligence, University Hospital of North Norway},%Department and Organization
            city={Tromsø},
            postcode={9019}, 
            country={Norway}}    
\affiliation[c]{organization={Department of Computer Science and Engineering, University of Bologna}, 
 postcode={40126},
city={Bologna}, 
country={Italy}}
\affiliation[d]{organization={Fraunhofer Research Institution for Individualized and Cell-Based Medical Engineering IMTE}, 
 postcode={23562},
city={Lübeck}, 
country={Germany}}
\affiliation[e]{organization={Department of Radiology, Netherlands Cancer Institute (NKI)}, city={Amsterdam}, postcode={1066 CX}, country={The Netherlands}}
\affiliation[f]{organization={Norwegian Computing Center},%Department and Organization
            city={Oslo},
            postcode={0373}, 
            country={Norway}}
\affiliation[g]{organization={Pioneer Centre for AI, University of Copenhagen},%Department and Organization
            city={Copenhagen},
            postcode={1350}, 
            country={Denmark}}              
%% Abstract
\begin{abstract}

Regular mammography screening is crucial for early breast cancer detection. By leveraging deep learning-based risk models, screening intervals can be personalized, especially for high-risk individuals. While recent methods increasingly incorporate longitudinal information from prior mammograms, accurate spatial alignment across time points remains a key challenge. Misalignment can obscure meaningful tissue changes and degrade model performance. In this study, we provide insights into various alignment strategies, image-based registration, feature-level (representation space) alignment with and without regularization, and implicit alignment methods, for their effectiveness in longitudinal deep learning-based risk modeling. Using two large-scale mammography datasets, we assess each method across key metrics, including predictive accuracy, precision, recall, and deformation field quality.

Our results show that image-based registration consistently outperforms the more recently favored feature-based and implicit approaches across all metrics, enabling more accurate, temporally consistent predictions and generating smooth, anatomically plausible deformation fields. Although regularizing the deformation field improves deformation quality, it reduces the risk prediction performance of feature-level alignment. Applying image-based deformation fields within the feature space yields the best risk prediction performance.

These findings underscore the importance of image-based deformation fields for spatial alignment in longitudinal risk modeling, offering improved prediction accuracy and robustness. This approach has strong potential to enhance personalized screening and enable earlier interventions for high-risk individuals. The code is available at \url{https://github.com/sot176/Mammogram_Alignment_Study_Risk_Prediction.git}, allowing full reproducibility of the results. 

\end{abstract}

%% Keywords

\end{frontmatter}




\section{Introduction}\label{intro}
Breast cancer remains the most prevalent cancer type among women worldwide, posing a significant public health challenge across diverse populations~\citep{cancerstat}. Mammography is the current gold standard for breast cancer screening and is typically performed in two standard views, craniocaudal (CC) and mediolateral oblique (MLO), to provide comprehensive visualization of breast tissue~\citep{pharmaceutics13050723}. Widespread implementation of regular screening programs has led to a substantial reduction in breast cancer mortality, primarily through early detection and intervention~\citep{zielonke2020evidence}. However, despite this progress, there remains significant room for improvement, particularly for individuals at higher risk~\citep{rubio2024risk}.  

One of the key challenges in the detection of breast cancer is breast density, which not only increases the risk of developing cancer, but also reduces the sensitivity of mammography~\citep{bodewes2022density}. Dense breast tissue can obscure malignant lesions, complicating early detection and increasing the likelihood of missed diagnoses during routine screening~\citep{cancerriskdensity}. As a result, women with high breast density face a dual burden: an increased risk of cancer and a reduced screening effectiveness. Although an increase in screening frequency could mitigate these risks, such measures are often constrained by the capacity of the healthcare system~\citep{mammoscreen}, creating a trade-off between the benefits of early detection and the practical limitations of routine imaging. These challenges underscore the need for a more personalized approach to screening. In this context, personalized risk prediction models offer a promising solution, enabling tailored screening intervals that improve results while optimizing resource allocation~\citep{eriksson2023risk}.

Recent advances in deep learning have shown considerable promise in improving breast cancer risk prediction using mammography images~\citep{2024riskreview}. However, while radiologists routinely compare current mammograms with previous exams to distinguish between benign and malignant changes and improve diagnostic precision~\citep{akwo2024access, prioirmammo}, early deep learning models primarily analyzed current images in isolation, overlooking valuable temporal information~\citep{mirai,2024riskasymmirai}. Incorporating longitudinal data, such as prior screenings, can capture subtle tissue changes over time, which are often critical for accurate risk assessment~\citep{changesrisk}. This highlights the growing recognition that the temporal context is essential for robust breast cancer risk prediction, especially in cases where changes evolve gradually over multiple years.

Based on this insight, recent studies have shown that models that incorporate longitudinal mammography outperform those based solely on single-time-point images~\citep{miccai2023,Wang2023,dadsetan2022deep, Kar_Longitudinal_MICCAI2024,Wan_Ordinal_MICCAI2024,zijunmiccai}. By integrating both prior and current mammograms, these models more effectively capture temporal dynamics, thereby improving the accuracy of risk assessments. However, precise alignment of mammograms across time points is critical for reliably evaluating tissue changes. Misalignment can obscure or distort relevant features, leading to inaccurate predictions. This challenge is compounded by the non-rigid nature of breast tissue~\citep{guo2006breast} and variations in breast positioning and compression during image acquisition. To address these issues, image registration, both through explicit and implicit techniques, plays a vital role in aligning sequential mammograms and improving the performance of longitudinal risk models.
 
Various alignment strategies have been proposed for mammography, yet it remains unclear which approach is most effective for downstream tasks such as breast cancer risk prediction. Direct comparisons between these methods are scarce, and the relationship between alignment quality and predictive performance has not been systematically explored. In general, alignment methods can be categorized as explicit or implicit. Explicit alignment can be applied at the image level (input space) or the feature level (representation space), offering anatomically precise transformations, but potentially introducing artifacts or interpolation errors. In contrast, implicit alignment is learned jointly with feature extraction and may yield more robust representations, though often at the expense of anatomical precision. While each approach has its advantages and limitations, their comparative impact on predictive performance remains underexplored. Notably, there are currently no deep learning-based registration models developed for explicit longitudinal alignment of mammograms directly in the image space. Traditional registration methods, while effective, are often computationally intensive and time-consuming. To overcome these limitations, we introduce MammoRegNet, a deep learning model specifically designed for accurate and efficient registration of longitudinal mammograms, leveraging advances in explicit alignment techniques from other imaging domains. 

This work presents the first comprehensive evaluation of both alignment paradigms, analyzing their trade-offs and exploring whether a hybrid strategy can optimize spatial alignment and predictive accuracy. Importantly, all alignment approaches are evaluated within a unified risk prediction framework to ensure a consistent and fair comparison of their impact on breast cancer risk prediction.
    
The main contributions of this study are summarized as follows:
\begin{itemize}
  \item We present the first systematic evaluation of explicit (image-based and feature-based) and implicit alignment strategies for longitudinal breast cancer risk prediction, within a unified risk prediction framework to ensure consistency across comparisons.
\item We demonstrate that spatial alignment, especially through image-based methods, consistently improves predictive accuracy and robustness over time.
\item We show that enhancing deformation quality via regularization in feature alignment reduces risk prediction performance, revealing a key trade-off.
\end{itemize}

A preliminary version of this work appeared in~\citep{ownpaper}. In this extended study, we significantly enhance our previous work through the following key improvements:
\begin{enumerate}
    \item We expand the scope of the analysis to include implicit alignment methods, enabling a more comprehensive comparison across alignment strategies. Our earlier work focused exclusively on explicit approaches.
    \item We adopt higher-resolution mammography images ($1664 \times 2048$ vs. $512 \times 1024$) to align with recent state-of-the-art studies~\citep{mirai,2024riskasymmirai,Kar_Longitudinal_MICCAI2024}. Although image-based alignment previously showed superior performance at lower resolutions, we find that applying image-based deformation fields in feature space yields improved results at higher resolutions, likely due to richer and more semantically meaningful representations.
    \item We enhance model performance by incorporating a temporal self-attention block that effectively captures dependencies across time between mammographic feature representations.
    \item We validate our findings using the OA-BreaCR framework~\citep{Wan_Ordinal_MICCAI2024}, demonstrating consistency of alignment effects across different longitudinal risk prediction models. 
    \item We include a more extensive literature review to better situate our work within the broader research landscape.
    \item We provide a substantially expanded methodology section, with detailed descriptions of alignment strategies, risk prediction models, and a thorough evaluation of their effects on predictive performance.
\end{enumerate}

The paper is structured as follows. Section~\ref{relwork} provides an overview of relevant related work, with a focus on breast cancer risk prediction and image registration. Section~\ref{methods} presents a detailed description of the risk prediction pipeline incorporating various alignment methods. In Section~\ref{expsetup}, we describe the experimental setup and the evaluation metrics used to assess the performance of the different alignment strategies. Sections~\ref{result} and~\ref{discussion} present and discuss the results, respectively.  Finally, Sections~\ref{limitationsoutlook} and~\ref{conclusion} discuss the limitations of the proposed approach, outline future research directions, and summarize the key findings of the study.

\section{Related Work}\label{relwork}
Deep learning has substantially advanced the fields of breast cancer risk prediction and image registration, driving improvements in early detection and longitudinal analysis. In the following, we first review recent developments in deep learning-based breast cancer risk prediction (Section~\ref{relworkriskpred}), with an emphasis on models that incorporate temporal information across multiple screenings. We then discuss deep learning approaches to medical image registration (Section~\ref{relworkreg}), highlighting emerging efforts to adapt these methods to the unique challenges of mammography, particularly in longitudinal settings.

\subsection{Breast Cancer Risk Prediction} \label{relworkriskpred}

Breast cancer risk prediction is a well-established field, with early models relying on radiomic-based characteristics such as mammographic density and parenchymal texture extracted through hand-crafted algorithms~\citep{Anandarajah2022,libra,Tan2016}. Mammographic density is a well-known independent risk factor and remains a cornerstone of clinical risk assessment. 

More recent evidence, however, emphasizes that breast cancer risk extends beyond density alone and can be characterized by a diverse set of imaging markers across multiple modalities. In this context, AI-driven image features have emerged as key enablers of personalized, risk-adapted breast cancer screening strategies, reflecting a shift from hand-crafted radiomic descriptors toward deep learning–based representations learned directly from digital mammography and tomosynthesis~\citep{BeyondDensityReview,AIMammoPhenotypingReview}. Recent advances have thus increasingly focused on deep learning approaches that automatically learn predictive patterns from mammograms~\citep{mirai,guo2006breast,miccai2023,Wan_Ordinal_MICCAI2024,Kar_Longitudinal_MICCAI2024,risk2019,dadsetan2022deep,2023risk}. Current models typically fall into two categories: single-timepoint models and those that incorporate temporal information from sequential screenings. They may use imaging data alone or combine them with clinical risk factors, such as age, family history, or hormonal status. In both types, extracted features, whether hand-crafted or learned, tend to focus on critical indicators such as tissue density, mass morphology, and textural characteristics~\citep{evalAIrisk,2024riskreview}.

 Mirai~\citep{mirai} is a deep learning model designed for breast cancer risk prediction at multiple future timepoints, using standard mammographic views and, when available, non-imaging risk factors. To improve interpretability, AsymMirai~\citep{2024riskasymmirai} captures bilateral asymmetries in mammograms to approximate Mirai’s predictions while providing more transparent and explainable outputs.

A key limitation of many existing models, including Mirai and AsymMirai, is their reliance on a single mammographic timepoint, which ignores the temporal dynamics of breast tissue evolution. Recently, several approaches have emerged that incorporate longitudinal imaging data to improve predictive accuracy. These methods follow two strategies: explicit alignment of images across timepoints or implicit modeling of temporal changes without direct spatial alignment.

Some approaches rely on recurrent neural networks, particularly Long Short-Term Memory (LSTM)–based~\citep{lstm} models, to capture temporal dependencies across exams and to emphasize the importance of modeling change patterns for effective risk stratification~\citep{2023riskimpli}. Other methods employ transformer-based~\citep{attentionpaper} architectures. PRIME+~\citep{miccai2023} integrates prior mammograms using a transformer decoder and outperforms single-timepoint models in both short- and long-term risk prediction. Similarly, TRINet~\citep{trinet} combines sequential mammograms with time-aware attention mechanisms to predict short-term risk. Most recently, LoMaR~\citep{Kar_Longitudinal_MICCAI2024} extended Mirai by introducing a transformer-based model that processes multiple prior mammograms, implicitly modeling temporal changes without relying on non-imaging risk factors. Similarly, VMRA-MaR~\citep{zijunmiccai} presents a framework for predicting breast cancer risk that takes advantage of both longitudinal imaging data and bilateral asymmetry in mammograms.

While many models incorporate temporal information through feature fusion or sequential modeling, some take a more explicit approach by aligning either mammographic images or extracted feature maps to better capture localized changes over time. LRP-NET~\citep{dadsetan2022deep} tracks spatiotemporal breast tissue changes across four prior exams by applying affine image registration before training. The results demonstrate the benefits of spatial alignment in longitudinal modeling, showing improved performance over single-timepoint baselines. In contrast, OA-BreaCR~\citep{Wan_Ordinal_MICCAI2024} spatially aligns feature maps extracted from sequential mammograms using an attention-based mechanism. By modeling temporal changes at the feature level and incorporating ordinal learning to account for time-to-event outcomes, OA-BreaCR enhances risk assessment while maintaining interpretability, supporting more personalized screening strategies.

Together, these studies highlight the growing importance of temporal modeling in the prediction of breast cancer risk, since sequential mammograms reveal dynamic patterns that are often missed by single-timepoint analysis. However, the alignment of temporal information, whether at the image or feature level, can significantly impact the model's ability to capture meaningful changes. Therefore, a systematic evaluation of different alignment strategies is essential to understand their influence on downstream risk prediction performance. Although prior work has demonstrated the benefits of incorporating longitudinal information into breast cancer risk models, the role of alignment has remained underexplored, particularly in the context of mammography. Our work addresses this gap by providing insights into various alignment methods, including image-level registration, feature-level alignment, and implicit approaches, and by evaluating their effects on risk prediction performance.

\subsection{Registration} \label{relworkreg}
Image registration plays a critical role in longitudinal medical imaging by enabling the spatial alignment of scans acquired at different time points. In the context of breast cancer screening, explicit alignment, whether at the image or feature level, is essential for accurately tracking tissue changes across visits, such as evolving asymmetries or subtle structural shifts, which may signal early disease development~\citep{regchangetrack}. By improving spatial correspondence over time, explicit registration enhances the interpretability and reliability of risk prediction models that leverage sequential imaging data. However, most deep learning-based image registration methods have primarily been developed and evaluated on volumetric imaging modalities such as magnetic resonance imaging (MRI) and computed tomography (CT), with a strong focus on anatomical regions like the brain and abdomen~\citep{regsurvey}.

Among these, convolutional neural network (CNN)-based models have been foundational in medical image registration. VoxelMorph~\citep{voxelmorph} pioneered deep learning-based deformable registration by learning spatial transformations directly from image pairs in an unsupervised manner. Building on this foundation, models such as AMNet~\citep{AMNet} and PC-Reg~\citep{pcreg} introduced architectural innovations to enhance performance, AMNet through its adaptive multi-level design, and PC-Reg with a pyramidal prediction–correction approach to handle large deformations. More recently, transformer-based architectures like TransMorph~\citep{transmorph} and TransMatch~\citep{transmatch} have been proposed for unsupervised deformable registration, with the latter incorporating a dual-stream feature matching strategy. NICE-Trans~\citep{nicetrans} further advances this trend by introducing a non-iterative coarse-to-fine transformer network that jointly learns both affine and deformable transformations, enabling efficient and accurate alignment without multi-stage optimization.

While deep learning-based image registration has made significant strides in volumetric imaging domains, its application to mammography remains relatively limited. A few studies have explored registering the two standard mammographic views, CC and MLO, within a single screening episode using CNN-based methods to enhance feature correspondence~\citep{walton2022,li2020mammography}. In contrast, longitudinal mammography registration, which involves aligning images from the same view acquired at different timepoints, has largely relied on conventional image processing techniques. Traditional registration algorithms have, for instance, been used to subtract prior images from recent ones, thereby improving the visibility of new abnormalities~\citep{Loizidou2021,loizidou2022}.

Although conventional registration methods have been applied to longitudinal mammography to align images from different timepoints, these approaches are often time-consuming and computationally expensive. We present MammoRegNet, a deep learning-based image registration approach tailored for longitudinal mammography. Building on recent advances in image registration for brain MRI, MammoRegNet efficiently aligns mammograms from the same view (e.g. CC or MLO) acquired at different timepoints. This enables high-quality spatial correspondence across longitudinal exams. By applying MammoRegNet, we provide new insights into how precise image-level alignment over time can enhance risk model performance, highlighting the potential of deep learning-based registration to improve breast cancer screening outcomes.

\section{Methods}\label{methods}
This section presents our unified framework for analyzing different alignment strategies and their impact on breast cancer risk prediction performance. We first formalize the core problem in Section~\ref{problemdef}. Section~\ref{riskhead} then introduces the shared components of the risk prediction head, including temporal attention, the cumulative probability layer, and multilevel joint learning with the associated loss functions.
We subsequently describe the single-timepoint model in Section~\ref{singletimepoint}, followed by the no-alignment baseline in Section~\ref{noalign}. The implicit alignment approach is detailed in Section~\ref{impalign}. Finally, Section~\ref{expalign} presents two explicit alignment strategies: feature-level alignment (Section~\ref{featalign}) and image-level alignment (Section~\ref{imgalign}).

\subsection{Problem definition}\label{problemdef}
The risk prediction task involves estimating the probability that an individual will develop breast cancer within a specified time frame. In this study, we focus on predicting the likelihood of breast cancer occurrence within a five-year period following a given mammogram. Accurate alignment is essential in this context, as it ensures consistent localization of anatomical structures over time and facilitates the interpretation of temporal changes. This consistency enables the model to better capture subtle longitudinal patterns, such as progressive tissue changes or emerging abnormalities, which may signal an increased risk of cancer.

To effectively leverage temporal information from longitudinal mammography scans, it is crucial to model how breast tissue evolves over time. One approach is implicit alignment, which relies on deep learning models to learn temporal dependencies directly from the data, without explicitly aligning the images or their corresponding features. These models capture the underlying temporal relationships between scans by learning patterns of progression or change across time points. However, to provide a more structured temporal integration, we can also incorporate explicit alignment, where the goal is to align a current mammogram \(\mathbf{I}^{\text{cur}} \in \mathbb{R}^{H \times W}\) or its corresponding feature map \(\mathbf{f}^{\text{cur}} \in \mathbb{R}^{C \times h \times w}\) with a prior mammogram \(\mathbf{I}^{\text{pri}} \in \mathbb{R}^{H \times W}\) or its feature map \(\mathbf{f}^{\text{pri}} \in \mathbb{R}^{C \times h \times w}\). Here, \(H\) and \(W\) denote the spatial resolution of the input images, while \(h < H\) and \(w < W\) indicate the reduced spatial resolution of the feature maps due to downsampling in the backbone network. \(C\) denotes the number of feature channels. In this case, we aim to compute a deformation field \(\boldsymbol{\phi} \in \mathbb{R}^{2 \times H \times W}\) that minimizes a cost  function $E$, which measures the dissimilarity between the current image and the spatially transformed prior image. The formulation is given by the following optimization problem
\begin{equation}
\widehat{\boldsymbol{\phi}} = \arg \min_{\boldsymbol{\phi}}  E( \mathbf{I}^{\text{cur}}, \mathbf{I}^{\text{pri}} \circ \boldsymbol{\phi}) + \lambda R(\boldsymbol{\phi}) .
\end{equation}
Here, $E$ quantifies the image dissimilarity, while $ R(\boldsymbol{\phi})$ is a regularization term that enforces desirable properties such as spatial smoothness and anatomical plausibility. The hyperparameter $ \lambda$  balances image similarity with regularization to prevent unrealistic deformations (e.g., folding). This formulation can also be extended to feature-level alignment, where feature maps are registered by minimizing the same cost function.

\begin{figure}[t]
    \centering  
    \begin{subfigure}{1.0\textwidth}
            \centering  
        \includegraphics[height=1.5cm]{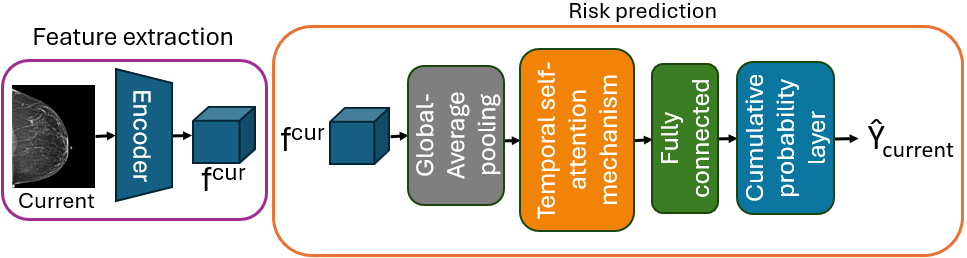}
        \caption{SingleTimePoint}
        \label{fig:singletimepoint}
    \end{subfigure} \\
       \begin{subfigure}{0.43\textwidth}
            \centering  
        \includegraphics[height=3.1cm]{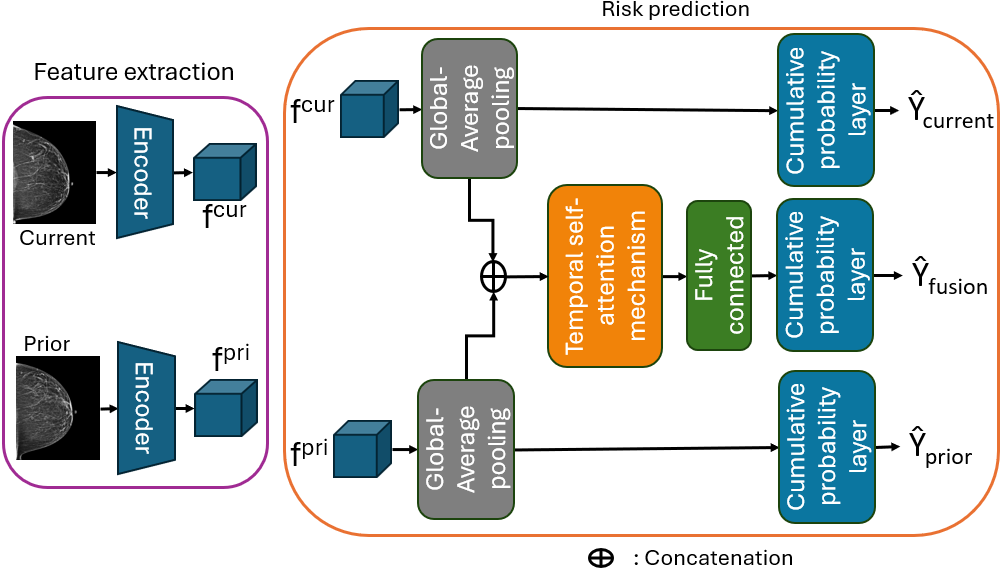}
        \caption{NoAlign}
        \label{fig:noalign}
    \end{subfigure} 
     \begin{subfigure}{0.43\textwidth}
            \centering  
        \includegraphics[height=3.1cm]{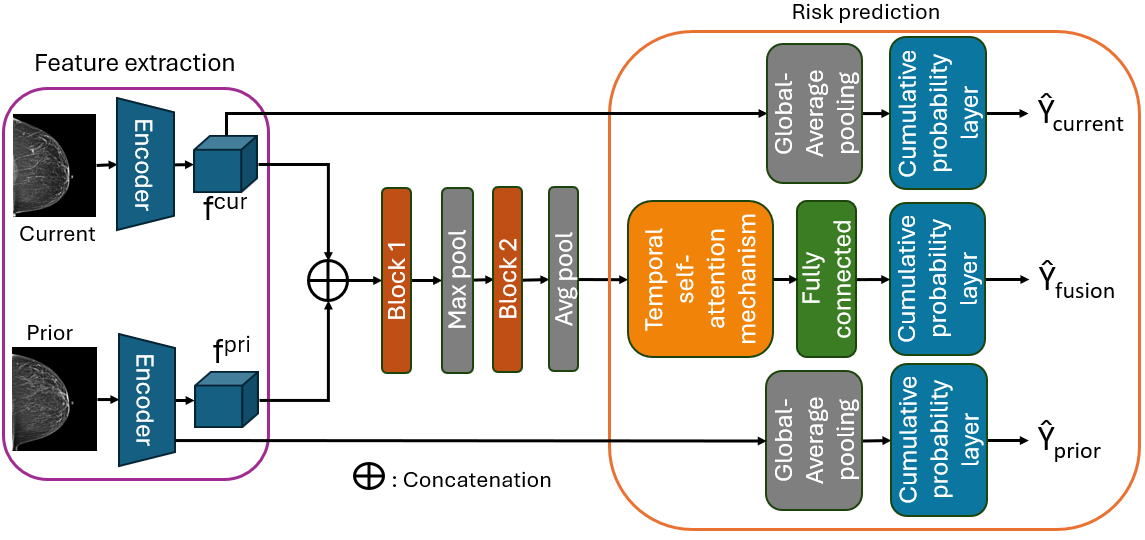}
        \caption{Implicit Alignment}
        \label{fig:impalign}
    \end{subfigure} 
    \\
    \begin{subfigure}{0.23\textwidth}
        \centering
    \includegraphics[height=2.82cm]{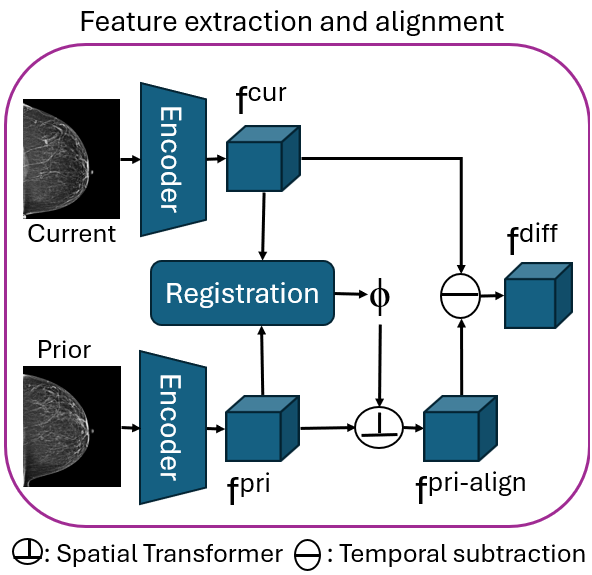}
        \caption{FeatAlign}
                \label{fig:featalign}
    \end{subfigure}
    \begin{subfigure}{0.25\textwidth}
            \centering 
    \includegraphics[height=2.8cm]{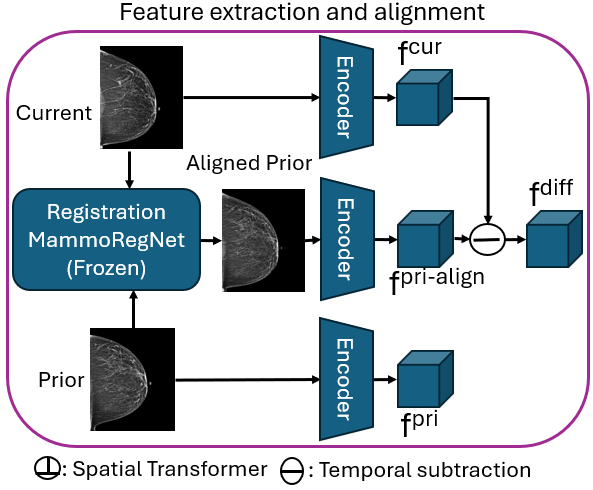}
        \caption{ImgAlign}
                \label{fig:imgalign}
    \end{subfigure}
     \begin{subfigure}{0.23\textwidth}
         \centering   
    \includegraphics[height=2.8cm]{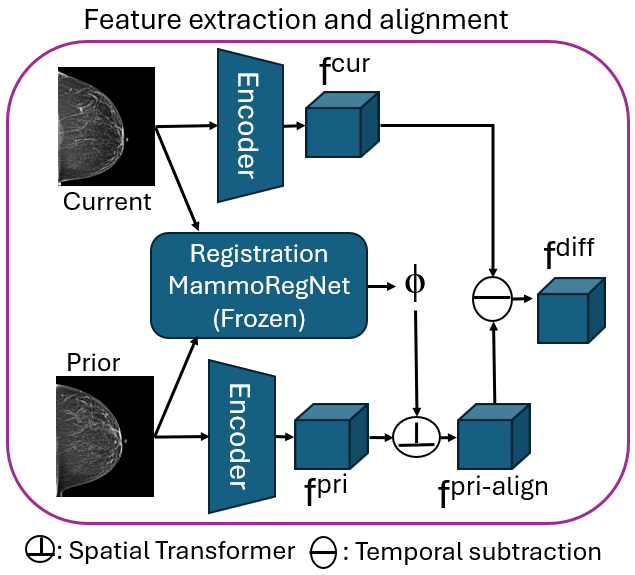}
        \caption{ImgFeatAlign}
                \label{fig:imgfeatalign}
    \end{subfigure}
     \begin{subfigure}{0.26\textwidth}
        \centering
\includegraphics[height=2.8cm]{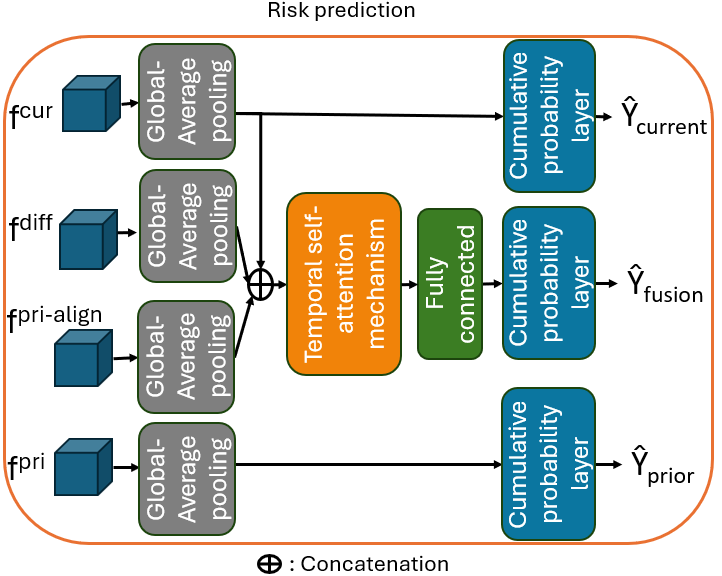}
      \caption{Risk }
    \label{fig:alignrisk}
       \end{subfigure}
    \caption{Overview of longitudinal risk prediction methods: (a) Using only the current mammogram, which serves as our single–timepoint risk model, (b) Direct feature extraction without alignment,  (c) Implicit Alignment, (d) Feature-level alignment, (e)  Image-level alignment with MammoRegNet, (f) Applying MammoRegNet’s deformation field in feature space, and (g) Risk prediction using alignment methods (d), (e), and (f).}
    \label{fig:alignmentmethods}
\end{figure}

To evaluate the impact of alignment on risk prediction, we compare five strategies within a unified framework (Figure~\ref{fig:alignmentmethods}):
\begin{itemize}
    \item Single-timepoint model (Section~\ref{singletimepoint}): Uses only the most recent mammogram to predict risk, providing a baseline to compare longitudinal models against single-timepoint performance.
    \item No alignment (Section~\ref{noalign}): Serves as a longitudinal baseline where no temporal registration is performed between mammograms.
    \item Implicit alignment (Section~\ref{impalign}): Learns spatial correspondences as part of the risk prediction process without an explicit registration step.
    \item Feature-level alignment (Section~\ref{featalign}): Inspired by recent state-of-the-art methods~\citep{Wan_Ordinal_MICCAI2024}, this approach aligns feature maps to enhance semantic consistency but may lose critical spatial information.
    \item Image-level alignment (Section~\ref{imgalign}): Operates directly on pixel space, preserving spatial and textural details important for detecting subtle longitudinal changes.
\end{itemize}

This setup enables a systematic assessment of how different alignment strategies influence the performance of longitudinal breast cancer risk prediction.
 
\subsection{Risk Prediction Head}\label{riskhead}
Before defining the different strategies, we first present the risk prediction head, which is shared across all models and consists of the following key elements: the Temporal Self-Attention Block (Section~\ref{tempatt}), Multilevel Joint Learning (Section~\ref{mljl}), and the Cumulative Probability Layer (Section~\ref{cpl}). Additionally, we introduce the Loss Function and Censoring mechanism in Section~\ref{lossfunc}. These components collectively enable effective modeling of temporal dependencies, feature-level interactions, and time-dependent risk estimation.

\subsubsection{Temporal Self-Attention Block}\label{tempatt}

Building on prior work employing Transformer encoders and attention mechanisms for medical imaging~\citep{2024riskinterpret,miccai2023}, we use a temporal self-attention block~\citep{li2023time} to capture interactions between mammographic representations acquired at different time points. Unlike spatial self-attention, which operates over image patches, this module applies self-attention to a short temporal sequence of feature tokens derived from longitudinal examinations.

For each input sample, convolutional backbones extract feature maps from the current and prior mammograms, each of shape $(C, H, W)$. Global average pooling is applied to reduce these maps into compact feature vectors of dimension $C$, which serve as tokens representing the prior and current images. These tokens are stacked to form a temporal sequence of length two and passed through a multi-head self-attention layer~\citep{attentionpaper}. In this setup, self-attention models interactions between the prior and current representations, enabling the network to emphasize longitudinal changes in the feature space rather than relying solely on absolute image appearance. Following prior work~\citep{Wan_Ordinal_MICCAI2024}, no positional encoding is applied to the prior and current tokens. The attention block is followed by a feedforward network with residual connections and layer normalization, which stabilize training and enhance representational capacity.

For models employing explicit spatial alignment, the temporal self-attention formulation is extended to include a time-aware difference representation. An aligned prior feature map is first computed and subtracted from the current feature map to obtain a difference feature map, which is augmented with a continuous time encoding representing the elapsed interval between examinations~\citep{Wan_Ordinal_MICCAI2024}. Temporal encoding is applied exclusively to the difference features to explicitly model longitudinal changes. After global average pooling, three feature tokens are obtained: the aligned prior token, the time-aware difference token, and the current image token. These are stacked and processed by the same multi-head self-attention layer, without modifying the attention architecture, feedforward layers, or normalization strategy. The inclusion of the difference token enables the attention mechanism to explicitly capture localized longitudinal changes while maintaining a unified framework across both implicit and explicit alignment settings.

\subsubsection{Multilevel Joint Learning}\label{mljl}
The models use multilevel joint learning~\citep{Wan_Ordinal_MICCAI2024} to capture temporal and feature-level dependencies between current and prior mammographic images. In addition to learning a fused representation of current and prior features, the model also uses the individual current and prior feature representations to independently predict the risk associated with each time point. Separate loss terms are computed using the labels corresponding to each image. By jointly optimizing both the fused and individual predictions, the model is encouraged to learn progression-related knowledge that reflects breast cancer development over time, rather than relying solely on the fused features for risk prediction.

\subsubsection{Cumulative Probability Layer}\label{cpl}
The cumulative probability layer~\citep{mirai,sybil,miccai2023} estimates the cumulative probability of an event over time, a formulation commonly used in survival analysis and time-to-event modeling. It consists of two fully connected layers: one that maps feature representations to hazard values at each future time step (e.g., 1 year, 2 years, …), and another that computes a baseline hazard. A ReLU activation~\citep{relu} ensures the hazard values remain non-negative. The final cumulative probability at time $t$ is obtained by summing the hazards from previous time steps, along with the baseline hazard, yielding a continuous cumulative risk representation over time, according to 
\begin{equation}\label{cumprob}
P(t \mid \mathbf{f}) = h_{\text{base}}(\mathbf{f}) + \sum_{j=1}^t h_j(\mathbf{f}) ,  
\end{equation}
where $P(t \mid \mathbf{f})$ is the cumulative probability at time $t$, given the input feature map $\mathbf{f}$, $ h_{base}(\mathbf{f})$ is the baseline hazard, and $h_j(\mathbf{f})$ is the hazard value at each time step $j$. This formulation allows the model to capture the sequential accumulation of risk over time while preserving temporal dependencies.

\subsubsection{Loss function and Censoring}\label{lossfunc}
The model is trained for time-dependent risk estimation using a binary cross-entropy (BCE) loss. The ground-truth $\mathbf{y}$ is a binary vector of length $T_{max}$, where $T_{max}$ represents the maximum observation period (5 years). For each time step $t$, $\mathbf{y}(t)$ is 1 if the patient is diagnosed with cancer within $t$ years and  $0$ otherwise.
However, real-world clinical data often contain censored observations (i.e., patients lost to follow-up before an event occurs). To address this, we follow~\citep{mirai,Wan_Ordinal_MICCAI2024} and use a masking function $\delta(t)$ to ensure the loss computation considers only valid observations. The masking function is defined as follows
\begin{equation}\label{mask}
    \delta(t) =
\begin{cases} 
        1, & \begin{aligned}[t] 
                & \text{if the follow-up period} \geq \min(t, T_{max}) \\  
                & \text{or if the patient is diagnosed within } n \text{ years}.
            \end{aligned} \\ 
        0, & \text{otherwise}.
    \end{cases}
\end{equation}
This masking mechanism ensures that only valid observations contribute to the loss calculation, preventing bias due to incomplete follow-up periods.

Thus, the overall binary cross-entropy loss  $\mathcal{L}_{\text{BCE}}$ is computed as
\begin{equation}
    \mathcal{L}_{\text{BCE}} = \sum_{t=1}^{T_{\text{max}}} \delta(t) \cdot \ell_{\text{BCE}}(t) \; ,
\end{equation}
where $ \ell_{\text{BCE}}(t)$ represents the binary cross-entropy loss at time step $t$ and is given by
\begin{equation}
    \ell_{\text{BCE}}(t) = -\left[  y(t) \cdot \log(P(t)) + (1 -  y(t) \cdot \log(1 -P(t)) \right] \; ,
\end{equation}
with  \( y(t) \) being the ground truth, and \( P(t) \) being the predicted probability of cancer diagnosis at time \( t \).
This formulation ensures that loss contributions come only from valid, observed time points.

\subsection{SingleTimePoint Model}\label{singletimepoint}

We first introduce a single-timepoint risk prediction model, the simplest version of our framework, which estimates breast cancer risk using only the current screening mammogram. While it does not capture temporal changes, it provides a strong baseline and a consistent architectural foundation for longitudinal models. Performance improvements in multi-timepoint models can therefore be attributed to the inclusion of prior exams and alignment mechanisms. 

\textbf{Feature Extraction and Aggregation:}
A convolutional encoder extracts a spatial feature map from the current mammogram, which is aggregated via global average pooling into a compact feature vector.

\textbf{Risk Prediction:}
The pooled feature is processed by the Temporal Self-Attention Block (Section~\ref{tempatt}), adapted here for a single feature token, followed by the risk prediction head (Sections~\ref{mljl}–\ref{lossfunc}). The Cumulative Probability Layer (Section~\ref{cpl}) produces a five-dimensional output representing the cumulative probability of developing breast cancer within 1–5 years.

A schematic overview of the single-timepoint model is shown in Figure~\ref{fig:singletimepoint}.

\subsection{No Alignment}\label{noalign}

The no-alignment model (Figure~\ref{fig:noalign}) extends the single-timepoint baseline by incorporating both current and prior mammograms without applying any alignment. This model serves as a baseline to evaluate the impact of alignment mechanisms on longitudinal risk prediction. It builds on prior work~\citep{mirai,Wan_Ordinal_MICCAI2024} and leverages three key components: the Temporal Self-Attention Block (Section~\ref{tempatt}), Multilevel Joint Learning (Section~\ref{mljl}), and the Cumulative Probability Layer (Section~\ref{cpl}).

\textbf{Feature Extraction and Aggregation:}
A shared encoder extracts feature maps  \(\mathbf{f}^{\text{cur}}\) and  \(\mathbf{f}^{\text{pri}}\)
 from the current and prior images, after which global pooling is applied. The resulting pooled vectors are then \myhl{stacked along the feature dimension} to form a two-token sequence for processing by the temporal self-attention block.

\textbf{Risk Prediction:}
Following attention, multilevel joint learning is applied to predict risk at three levels: fused features, current features, and prior features. Each prediction is passed through a cumulative probability layer to produce a five-dimensional vector representing the estimated probability of developing breast cancer within 1–5 years.

\subsection{Implicit Alignment}\label{impalign}
The implicit alignment model (Figure~\ref{fig:impalign}) incorporates both current and prior mammograms, using two convolutional blocks to implicitly align features.

\textbf{Feature Extraction and Aggregation:}
A shared encoder extracts feature maps from both images, which are concatenated and processed by two convolutional blocks (each with a $3\times 3$ convolution, batch normalization, and ReLU). Max-pooling is applied after the first block to reduce spatial resolution, followed by global average pooling to obtain compact feature vectors.

\textbf{Risk Prediction:}
The pooled features are passed through the temporal self-attention block (Section~\ref{tempatt}) to capture longitudinal dependencies and subtle changes between exams. Multilevel joint learning (Section~\ref{mljl}) is applied, and risk is estimated using the cumulative probability layer (Section~\ref{cpl}) for the fused, current, and prior features.

\subsection{Explicit Alignment}\label{expalign}
In this section, we explore explicit alignment strategies for improving risk prediction, focusing on two distinct approaches: feature-level alignment and image-level alignment. Both methods aim to enhance the model's ability to capture critical temporal changes, but they operate at different levels of representation and offer complementary advantages.

\subsubsection{Feature-Level Alignment}\label{featalign}
The feature-level alignment model (Figures~\ref{fig:featalign},~\ref{fig:alignrisk}) builds on~\citep{Wan_Ordinal_MICCAI2024}, aligning current and prior feature maps via a dense deformation field.  

\textbf{Alignment Block:}  
The Alignment Block, adapted from~\citep{Wan_Ordinal_MICCAI2024}, processes the concatenated current and prior feature maps through two $3\times3$ convolutional layers with batch normalization~\citep{bn} and ReLU, followed by a final $3\times3$ convolution. This block predicts a dense deformation field \(\boldsymbol{\phi}\), which is applied to the prior feature map \(\mathbf{f}^{\text{pri}}\) to produce the aligned map \(\mathbf{f}^{\text{pri-align}}\).  

\textbf{Spatial Transformer Block:}  
The Spatial Transformer Block~\citep{voxelmorph,nicetrans} applies the deformation field using bilinear sampling, ensuring smooth and differentiable alignment between current and prior feature maps.  

\textbf{Loss Function and Regularization:}  
In addition to the feature alignment loss, we introduce regularization via $L_2$ penalties on the displacement field and a Jacobian Determinant (JD) loss to encourage smooth, anatomically plausible deformations. The general feature alignment loss is

\begin{equation}\label{lossfeat} \mathcal{L}_{\text{feat}} = \alpha \left( \left\| \mathbf{f}^{\text{pri-align}} - \mathbf{f}^{\text{cur}} \right\|_2^2 \right) + \beta \left( \sum_{\mathbf{p} \in \Omega} \left\| \nabla \boldsymbol{\phi}(\mathbf{p}) \right\|^2 + \lambda \text{JD} (\boldsymbol{\phi}) \right)\, , \end{equation}
where $\alpha$ and $\beta$ are weighting factors, $\Omega$ represents the set of spatial locations $\mathbf{p}$, $\nabla$ denotes the spatial derivative of $\boldsymbol{\phi}$ at $\mathbf{p}$, and $\lambda$  is the regularization parameter. The regularization term is composed of two components: $L_2$ regularization of the displacement field, $\sum_{\mathbf{p} \in \Omega} \left\| \nabla \boldsymbol{\phi}(\mathbf{p}) \right\|^2$, and Jacobian determinant loss, $\text{JD} (\boldsymbol{\phi})$. This regularization encourages the model to learn a smooth and physically plausible transformation, rather than simply minimizing the differences at the feature level. The regularization parameter $\lambda$  controls the trade-off between alignment accuracy and the physical plausibility of the transformation. Two variants are considered: \textit{FeatAlign} (\(\beta = 0\)) and \textit{FeatAlignReg} (\(\beta = 1\)) to evaluate the effect of regularization.   

\textbf{Risk Prediction with Aligned Features:}  
Aligned, current, prior, and difference features (\(\mathbf{f}^{\text{diff}} = \mathbf{f}^{\text{cur}} - \mathbf{f}^{\text{pri-align}}\)) are integrated using the temporal self-attention mechanism (Section~\ref{tempatt}) and processed via multilevel joint learning (Section~\ref{mljl}) before being passed to the cumulative probability layer (Section~\ref{cpl}) to estimate risk.

\subsubsection{Image-Level Alignment}\label{imgalign}

Traditional registration methods for longitudinal mammograms~\citep{loizidou2022,dadsetan2022deep} are often computationally expensive and require per-image optimization. As an alternative to feature-level alignment, we propose \textbf{MammoRegNet}, a deep learning-based registration network inspired by NICE-Trans~\citep{nicetrans}.  

\textbf{MammoRegNet} performs joint affine and deformable registration in a single encoder-decoder framework. The differences from NICE-Trans are:
\begin{itemize}
    \item Adaptation from 3D to 2D for mammography, replacing 3D convolutions with 2D operations.
    \item The Spatial Transformer blocks were adapted to operate on 2D mammography images.
    \item Decoder performs coarse-to-fine registration with four Swin Transformer refinement steps tailored for 2D images. 
\end{itemize}

\textbf{Loss Function:} MammoRegNet is trained using an unsupervised loss combining image similarity and regularization:
\begin{equation}\label{losssegunc}
       \mathcal{L}_{\text{img}} =  (1 - \text{NCC}_{\text{affine}}) + (1 - \text{NCC}_{\text{final}}) + \gamma \left( \sum_{\mathbf{p} \in \Omega} \left\| \nabla \boldsymbol{\phi}(\mathbf{p}) \right\|^2 + \lambda \text{JD} (\boldsymbol{\phi}) \right)\, ,
\end{equation} 
where NCC measures similarity between the current and transformed prior mammograms, $\Omega$ represents spatial locations, and $\gamma, \lambda$ control the trade-off between alignment accuracy and smoothness. Appendix~\ref{fig:mammoreg} provides full architectural details.

\textbf{Temporal Alignment and Risk Prediction:}  
For both image-level alignment strategies (Figures~\ref{fig:imgalign} and~\ref{fig:imgfeatalign}), current, prior, and aligned prior mammograms are encoded to extract features, and temporal subtraction is applied to generate difference maps. These features are processed through the temporal self-attention mechanism (Section~\ref{tempatt}) and multilevel joint learning (Section~\ref{mljl}), followed by the cumulative probability layer (Section~\ref{cpl}) for risk prediction. In the ImgFeatAlign model, MammoRegNet’s deformation field is applied directly within the feature space.

Detailed visualizations of all strategies are provided in~\ref{fig:alignment_models}

\section{Experimental Setup}\label{expsetup}
This section outlines the experimental setup for training and evaluating our models. We describe the datasets, label generation and preprocessing (Section~\ref{datasetshort}), evaluation metrics (Section~\ref{evalmet}), and model implementation and training details (Section~\ref{impdet}).

\subsection{Datasets, Label Generation, and Pre-processing}\label{datasetshort}

We evaluated our risk prediction models on two large, publicly available mammography datasets: the EMory BrEast Imaging  (EMBED)\footnote{\url{https://aws.amazon.com/marketplace/pp/prodview-unw4li5rkivs2\#overview}}~\citep{embed} dataset and the Cohort of Screen-Aged Women Case Control (CSAW-CC)\footnote{\url{https://snd.se/en/catalogue/dataset/2021-204-1}}~\citep{csawcc} dataset. The publicly released subset of EMBED includes 364,564 2D digital mammograms from 23,256 patients collected between 2013 and 2020, while CSAW-CC comprises mammograms from 873 breast cancer cases and 7,850 controls screened between 2008 and 2015. Both datasets include invasive and non-invasive breast cancer cases.

Negative cases correspond to women without a breast cancer diagnosis during the respective observation periods (8 years for EMBED and 7 years for CSAW-CC). Only patients with at least five years of follow-up were included (following~\citep{Wan_Ordinal_MICCAI2024}), ensuring longitudinal screening data. For patients with more than two screening exams, all consecutive screening pairs were used for training.

Time-to-Cancer labels represent the temporal interval between a screening mammogram and the corresponding cancer diagnosis and were generated following prior work~\citep{Wan_Ordinal_MICCAI2024,Kar_Longitudinal_MICCAI2024}. Dataset-specific procedures were used to identify positive and negative cases and assign cancer timing.

All mammograms were converted from DICOM to 16-bit grayscale PNG format and pre-processed following~\citep{mirai}, including background removal, breast region segmentation, resizing to \(1664 \times 2048\) pixels, and intensity normalization. Datasets were split at the patient level into training, validation, and test sets using a 5:2:3 ratio. 

Additional details on the datasets, including cancer case definitions, follow-up criteria, and breast density information, as well as comprehensive descriptions of the label generation procedures, Time-to-Cancer calculation, image pre-processing pipeline, and dataset splits are provided in~\ref{app:datasets}--\ref{app:preproc}.

\subsection{Evaluation Metrics}\label{evalmet}

We evaluate registration performance and risk prediction accuracy using standard metrics.  

\textbf{Image Registration Metrics:}  
Registration quality is assessed using normalized cross-correlation (NCC) to measure image similarity and the Jacobian determinant to evaluate deformation quality~\citep{reg_njd,nicetrans,DEVOS2019128}. For the Jacobian, we report the Percentage of Negative Determinants (NJD) and the standard deviation of the determinant to quantify smoothness and local foldings.

\textbf{Risk Prediction Metrics:}  
Model performance is evaluated using the concordance index (C-index)~\citep{c_index} and Area Under the ROC Curve (AUC)~\citep{auc} over 1–5 year risk horizons, with 95\% confidence intervals computed via bootstrapping (1,000 resamples)~\citep{mirai, miccai2023, Wan_Ordinal_MICCAI2024, sybil}. Statistical significance of improvements over the second-best model is assessed using a paired non-parametric bootstrap test on performance differences, with a one-sided hypothesis (ImgFeatAlign $>$ ImgAlign, ImgFeatAlign $>$ FeatAlign ) and p $<$ 0.05 considered significant.

\subsection{Implementation Details}\label{impdet}
For the risk prediction task, we adopt the pre-trained Mirai encoder~\citep{mirai} as our backbone. The Mirai encoder is based on a ResNet-18 architecture~\citep{resnet18}, modified to exclude the global average pooling and final fully connected layers. This modification allows us to retain fine-grained spatial feature representations essential for downstream processing. Importantly, the encoder is kept frozen during training to preserve the pre-trained representations. In the feature-level alignment setting, we jointly optimize the risk prediction and alignment tasks using a combined loss function, with the weighting coefficient $\alpha = 0.1 $  (see Equation~\ref{lossfeat}). For FeatAlignReg, we set $\beta=1 $ and $\lambda = 1 \times 10^{-5}$ in Equation~\ref{lossfeat}, following the settings used in MammoRegNet, to include regularization in the alignment loss. For FeatAlign, we set $\beta = 0$ to effectively remove the regularization term. For image-level alignment, the registration model is frozen, and only the risk prediction loss is computed.
 
Risk prediction models are trained using the Adam optimizer~\citep{adam} with a learning rate of $1 \times 10^{-5}$ and weight decay set to $1 \times 10^{-6}$. We use a batch size of 20 and train for 40 epochs. To improve generalization and ensure stable convergence, early stopping and learning rate decay are applied based on the validation C-index. The learning rate is halved if no improvement is observed over 5 consecutive epochs, and training is terminated if the validation metric does not improve within 15 epochs. To prevent overfitting, we applied data augmentations (e.g., RandomAffine, ColorJitter, RandomGamma, and RandomCrop) during training.

For the MammoRegNet architecture, we also use the Adam optimizer~\citep{adam} with a learning rate of $1 \times 10^{-4}$ and a weight decay of $ 1 \times 10^{-6}$. The batch size is set to 6, and training is conducted for 100 epochs. Each epoch consists of 1,500 randomly selected image pairs to ensure robust learning. In the loss function (Equation~\ref{losssegunc}), we use $\lambda = 1 \times 10^{-5}$ and $\gamma = 1$  (following~\citep{nicetrans}) to balance the contributions of different loss components. No learning rate scheduler is applied for MammoRegNet.

All models are implemented in PyTorch 2.0.1~\citep{pytorch} and trained on an AMD Instinct MI210 with 64 GB of memory. Training times ranged from 8 to 31 hours, depending on the complexity of the model and the size of the data set.

\section{Results}\label{result}
This section presents quantitative and qualitative evaluations of our alignment strategies for longitudinal breast cancer risk prediction. Section~\ref{overallriskper} assesses the impact of different alignment methods on overall prediction performance. Section~\ref{infreg} evaluates the impact of regularization on risk prediction performance. Section~\ref{embedana} provides an in-depth analysis on the EMBED dataset using ROC–AUC, precision–recall, short- and long-term performance across breast density categories, and subgroup analyses by cancer type and race. Finally, Section~\ref{deformqual} compares deformation field quality between image- and feature-based registration methods.

\subsection{Effect of Different Alignment Methods on Risk Prediction}\label{overallriskper}

\begin{table}[t]
\centering
\scriptsize
\caption{1–5 year breast cancer risk prediction using different alignment methods. C-index and AUC values with 95\% confidence intervals. $^{*}$ indicates that the increase in C-index and AUC of ImgFeatAlign, compared with ImgAlign (second-best model), is statistically significant (p $<$ 0.05).
$^{\dagger}$ indicates that the increase in C-index and AUC of ImgFeatAlign, compared with FeatAlign (third-best model), is statistically significant (p $<$ 0.05).}
\label{tab:riskpred}
\begin{adjustbox}{center}
\begin{tabular}{llc*{5}{c}}
\toprule
\textbf{Dataset} & \textbf{Method} & \textbf{C-index (\%) $\uparrow$} & \multicolumn{5}{c}{\textbf{Follow-up year AUC (\%) $\uparrow$}} \\
\cmidrule(lr){4-8}
& & & \textbf{1-yr} & \textbf{2-yr} & \textbf{3-yr} & \textbf{4-yr} & \textbf{5-yr} \\
\midrule

\multirow{10}{*}{\textbf{EMBED}} 
 & SingleTimePoint
 & \makecell{67.7 \\ (65.3-70.0)}
 & \makecell{69.6 \\ (66.6-72.4)}
 & \makecell{68.1 \\ (65.5-70.7)}
 & \makecell{66.6 \\ (64.1-69.1)}
 & \makecell{67.0 \\ (64.7-69.5)}
 & \makecell{66.6 \\ (62.7-70.3)} \\
 
& NoAlign        & \makecell{64.0 \\ (61.7-66.7)} & \makecell{64.9 \\ (62.1-67.9)} & \makecell{63.8 \\ (61.1-66.5)} & \makecell{63.7 \\ (61.2-66.3)} & \makecell{62.2 \\ (59.6-64.9)} & \makecell{55.7 \\ (51.4-60.0)} \\
& Implicit       & \makecell{70.9 \\ (68.6-73.3)} & \makecell{72.5 \\ (69.3-75.5)} & \makecell{71.2 \\ (68.6-73.7)} & \makecell{69.3 \\ (66.6-71.8)} & \makecell{70.4 \\ (67.9-72.9)} & \makecell{65.7 \\ (62.0-69.7)} \\
& FeatAlign      & \makecell{72.2 \\ (69.5-75.5)} & \makecell{72.4 \\ (69.5-75.6)} & \makecell{72.2 \\ (69.7-74.8)} & \makecell{72.0 \\ (69.7-74.6)} & \makecell{71.4 \\ (69.1-74.0)} & \makecell{68.5 \\ (64.8-72.0)} \\
& FeatAlignReg   & \makecell{70.6 \\ (67.8-73.2)} & \makecell{71.2 \\ (68.3-74.3)} & \makecell{71.3 \\ (68.6-74.0)} & \makecell{70.7 \\ (68.2-73.5)
} & \makecell{70.7 \\ (68.3-73.3)} & \makecell{65.7 \\ (61.7-69.6)} \\
& ImgAlign       & \makecell{72.3 \\ (69.6-74.8)} & \makecell{73.6 \\ (70.6-76.5)} & \makecell{71.7 \\ (69.0-74.2)} & \makecell{72.3 \\ (69.8-74.5)} & \makecell{72.4 \\ (70.0-74.7)} & \makecell{69.7 \\ (66.2-73.4)} \\
& \textbf{ImgFeatAlign} & \makecell{\textbf{74.7$^{*}$$^{\dagger}$} \\ \textbf{(72.3-77.0)}} & \makecell{\textbf{75.0 $^{\dagger}$  } \\ \textbf{(72.1-77.7)}} & \makecell{\textbf{75.5$^{*}$$^{\dagger}$} \\ \textbf{(73.1-77.9)}} & \makecell{\textbf{75.3$^{*}$$^{\dagger}$} \\ \textbf{(73.1-77.4)}} & \makecell{\textbf{75.9$^{*}$$^{\dagger}$} \\ \textbf{(73.6-78.0)}} & \makecell{\textbf{72.5$^{*}$$^{\dagger}$} \\ \textbf{(68.9-75.7)}} \\
\midrule

\multirow{10}{*}{\textbf{CSAW-CC}} 
 &   SingleTimePoint &  \makecell{  66.7 \\(64.6- 68.6)}&   \makecell{  67.5\\ (64.6-70.1} &   \makecell{ 67.7 \\ (65.5-69.7) } &   \makecell{ 67.9 \\ (65.8-69.8)} &   \makecell{ 66.8 \\ (64.7-68.8)} &  \makecell{68.5  \\(66.3-70.6) } \\
& NoAlign        & \makecell{65.9 \\ (64.0-67.8)} & \makecell{66.1 \\ (63.8-68.3)} & \makecell{66.2 \\ (64.3-68.2)} & \makecell{65.7 \\ (63.8-67.6)} & \makecell{65.1 \\ (63.2-67.0)} & \makecell{66.8 \\ (64.5-68.9)} \\
& Implicit       & \makecell{67.6 \\ (65.8-69.7)
} & \makecell{68.2 \\ (65.7-70.6)} & \makecell{68.5 \\ (66.5-70.5)} & \makecell{68.3 \\ (66.3-70.2)} & \makecell{67.5 \\(65.4-69.5)} & \makecell{68.7 \\ (66.3-71.1)} \\
& FeatAlign      & \makecell{69.1 \\ (67.0-71.1)} & \makecell{70.1 \\ (67.9-72.4)} & \makecell{69.9 \\ (67.9-71.8)} & \makecell{70.0 \\ (68.1-71.9)
} & \makecell{69.5 \\ (67.7-71.4)} & \makecell{71.6 \\ (69.4-73.8)} \\
& FeatAlignReg   & \makecell{68.4 \\ (66.4-70.-4)} & \makecell{68.9 \\ (66.7-71.2)} & \makecell{69.2 \\ (67.4-71.1)} & \makecell{69.8 \\ (68.0-71.6)} & \makecell{69.1 \\ (67.3-71.0)} & \makecell{72.0 \\ (69.9-74.2)} \\
& ImgAlign       & \makecell{70.2 \\ (68.1-72.1)} & \makecell{71.2 \\ (68.9-73.4)} & \makecell{71.0 \\ (69.2-73.1)} & \makecell{71.7 \\ (69.9-73.4)} & \makecell{71.1 \\ (69.1-72.9)} & \makecell{73.9 \\ (71.7-76.0)} \\
& \textbf{ImgFeatAlign} & \makecell{\textbf{70.4} \\ \textbf{(68.2-72.3)}} & \makecell{\textbf{72.0} \\ \textbf{(69.6-74.2)}} & \makecell{\textbf{71.5} \\ \textbf{(69.6-73.4)}} & \makecell{\textbf{72.6 $^{\dagger}$} \\ \textbf{(70.8-74.5)}} & \makecell{\textbf{72.0 $^{\dagger}$} \\ \textbf{(70.0-74.1)}} & \makecell{\textbf{75.2$^{*}$$^{\dagger}$} \\ \textbf{(73.1-77.5)}} \\
\bottomrule
\end{tabular}
\end{adjustbox}
\end{table}

\begin{table}[tbhp]
\centering
\scriptsize
\caption{1–5 year breast cancer risk prediction using different alignment methods. C-index and AUC values with 95\% confidence intervals. Results are also shown for the OA-BreaCR framework~\citep{Wan_Ordinal_MICCAI2024} to assess consistency across different risk prediction models. $^{*}$ indicates that the increase in C-index and AUC of ImgFeatAlign, compared with ImgAlign (second-best model), is statistically significant (p $<$ 0.05).
$^{\dagger}$ indicates that the increase in C-index and AUC of ImgFeatAlign, compared with FeatAlign (third-best model), is statistically significant (p $<$ 0.05). }
\label{tab:riskpred2}
\begin{adjustbox}{center}
\begin{tabular}{llc*{5}{c}}
\toprule
\textbf{Dataset} & \textbf{Method} & \textbf{C-index (\%) $\uparrow$} & \multicolumn{5}{c}{\textbf{Follow-up year AUC (\%) $\uparrow$}} \\
\cmidrule(lr){4-8}
& & & \textbf{1-yr} & \textbf{2-yr} & \textbf{3-yr} & \textbf{4-yr} & \textbf{5-yr} \\
\midrule

\multirow{10}{*}{\textbf{EMBED}} 
& FeatAlign     & \makecell{ 70.4 \\
(68.3-72.5)} & \makecell{70.3 \\ (67.1-73.2)} & \makecell{69.9 \\ (67.2-72.4)} & \makecell{71.0 \\ (68.5-73.4)} & \makecell{71.1 \\ (68.9-73.4)} & \makecell{70.9 \\ (68.6-73.3)} \\
& FeatAlignReg   & \makecell{ 67.9 \\ (65.5-70.3)} & \makecell{67.1 \\ (63.9-70.4)} & \makecell{67.6 \\ (64.8-70.4)} & \makecell{67.9 \\ (65.0-70.5)} & \makecell{67.1 \\ (64.2-69.8)} & \makecell{67.8 \\ (65.0-70.5)} \\
& ImgAlign       & \makecell{ 71.0\\ (68.8-73.2)} & \makecell{ 72.1 \\ (69.8-74.8)} & \makecell{71.4 \\ (68.6-74.2)} &\makecell{71.6\\ (68.9-74.2)} & \makecell{ 71.9\\ (69.4-74.2)} &\makecell{ 72.3\\ (70.0-74.6)} \\
& \textbf{ImgFeatAlign} & \makecell{\textbf{72.3$^{\dagger}$} \\ \textbf{(70.3-74.2)}} &  \makecell{\textbf{72.9$^{\dagger}$} \\ \textbf{(70.0-75.7)}}  &  \makecell{\textbf{72.3$^{\dagger}$} \\ \textbf{(67.4-72.7)}}  &  \makecell{\textbf{72.4$^{\dagger}$} \\ \textbf{(70.1-74.9)}}  &  \makecell{\textbf{72.3} \\ \textbf{(70.0-74.7)}} &  \makecell{\textbf{73.2$^{\dagger}$} \\ \textbf{(70.8-75.6)}}  \\
\midrule

\multirow{10}{*}{\textbf{CSAW-CC}} 

& FeatAlign     & \makecell{ 59.1\\ (57.3-60.9)} & \makecell{61.4\\ (59.0-63.7)} & \makecell{59.5\\ (57.4-61.6)} & \makecell{60.9\\ (58.9-62.9)} & \makecell{63.1\\ (61.1-65.2)} & \makecell{63.4\\ (61.4-65.5)} \\
& FeatAlignReg    & \makecell{58.1\\ (56.4-59.8)} & \makecell{58.0\\ (55.7-60.2)} & \makecell{57.6\\ (55.6-59.5)} & \makecell{58.8\\ (57.0-60.7)} & \makecell{61.0\\ (59.0-62.9)} & \makecell{60.7\\ (58.7-62.7)} \\
& ImgAlign   & \makecell{60.8\\ (59.1-62.6)} & \makecell{63.3\\ (60.9-65.6)} & \makecell{61.2\\ (59.2-63.2)} & \makecell{63.8\\ (61.8-65.8)} & \makecell{65.7\\ (63.6-67.6)} & \makecell{66.3\\ (64.2-68.2)} \\
& \textbf{ImgFeatAlign}  & \makecell{\textbf{61.6$^{\dagger}$}\\ \textbf{(61.2-65.8)}} & \makecell{\textbf{63.5$^{\dagger}$}\\ \textbf{(59.1-63.7)}} & \makecell{\textbf{61.7$^{\dagger}$}\\ \textbf{(59.6-63.8)}} & \makecell{\textbf{65.0$^{\dagger}$}\\ \textbf{(63.0-66.9)}} & \makecell{\textbf{67.0$^{\dagger}$}\\ \textbf{(65.0-68.9)}} & \makecell{\textbf{67.7$^{\dagger}$}\\ \textbf{(65.6-69.9)}} \\
\bottomrule
\end{tabular}
\end{adjustbox}
\end{table}

In this study, we evaluated the impact of a single–timepoint model and various alignment strategies: no alignment, implicit alignment, and explicit alignment at the feature and image levels, on breast cancer risk prediction. \footnote{Note, for image-level alignment, we introduce MammoRegNet, a deep learning–based registration model for longitudinal mammography. While this section focuses on risk prediction, detailed MammoRegNet registration results are provided in~\ref{app1}.}

Table~\ref{tab:riskpred} summarizes 1–5 year breast cancer risk prediction performance in terms of C-index and AUC (95\% confidence intervals) across alignment strategies. \myhl{\footnote{\myhl{See}~\ref{ablationstud} \myhl{for ablation studies demonstrating the contribution of the multi-head self-attention layer within the temporal self-attention mechanism.}}} Overall, methods incorporating explicit spatial–temporal alignment outperform NoAlign across datasets and time horizons. ImgFeatAlign achieves the strongest and most consistent performance, with the highest C-index and stable AUC over time, followed by ImgAlign. Implicit alignment yields moderate gains, while NoAlign shows the poorest performance, with pronounced AUC degradation at longer horizons, particularly on EMBED. These improvements are statistically significant. On EMBED, ImgFeatAlign achieves higher C-index and AUC than ImgAlign and FeatAlign across multiple follow-up horizons (p $<$ 0.05). On CSAW-CC, significant gains are mainly observed at longer follow-up periods, indicating greater benefits for long-term risk prediction. The SingleTimePoint model outperforms NoAlign, indicating that ignoring spatial misalignment degrades performance, underscoring the critical role of alignment. \myhl{The NoAlign model consistently achieves lower performance than the SingleTimePoint baseline, suggesting that longitudinal information alone is insufficient to improve prediction accuracy. Without explicit alignment, spatial inconsistencies between examinations can hinder the model's ability to learn meaningful temporal changes, reducing the benefit of incorporating prior mammograms.} However, it is consistently inferior to all alignment-based methods. Among feature-level approaches, FeatAlign outperforms FeatAlignReg, suggesting a trade-off between deformation smoothness and predictive accuracy. Overall, image-based alignment methods outperform feature-based approaches, with ImgFeatAlign providing the most robust and accurate risk prediction across datasets and time horizons.

To assess generalizability, we compare alignment strategies within the OA-BreaCR framework~\citep{Wan_Ordinal_MICCAI2024}, as summarized in Table~\ref{tab:riskpred2}. OA-BreaCR applies attention-based alignment to feature maps using an L2 loss to model temporal changes and adopts an ordinal learning formulation for joint time-to-event prediction and risk stratification. In contrast, our method aligns feature maps first, applies attention afterward, and optimizes a single risk prediction objective using binary cross-entropy. Despite these architectural differences, consistent trends are observed: ImgAlign and ImgFeatAlign outperform methods relying solely on L2-based feature alignment. Within OA-BreaCR, ImgAlign feeds aligned images directly into the encoder, while ImgFeatAlign uses unaligned images and leverages deformation fields from MammoRegNet for feature-level alignment. FeatAlign corresponds to OA-BreaCR’s framework, and FeatAlignReg adds a regularization loss. Both ImgFeatAlign and ImgAlign achieve superior predictive performance.

These results indicate that effective spatial–temporal alignment, is a key driver of improved risk prediction, independent of the downstream learning formulation.

\subsubsection{Performance across density categories}

 \begin{figure}[tb]
    \centering
    % First subfigure
    \begin{subfigure}[b]{0.48\textwidth}
        \centering
        \includegraphics[width=0.9\textwidth]{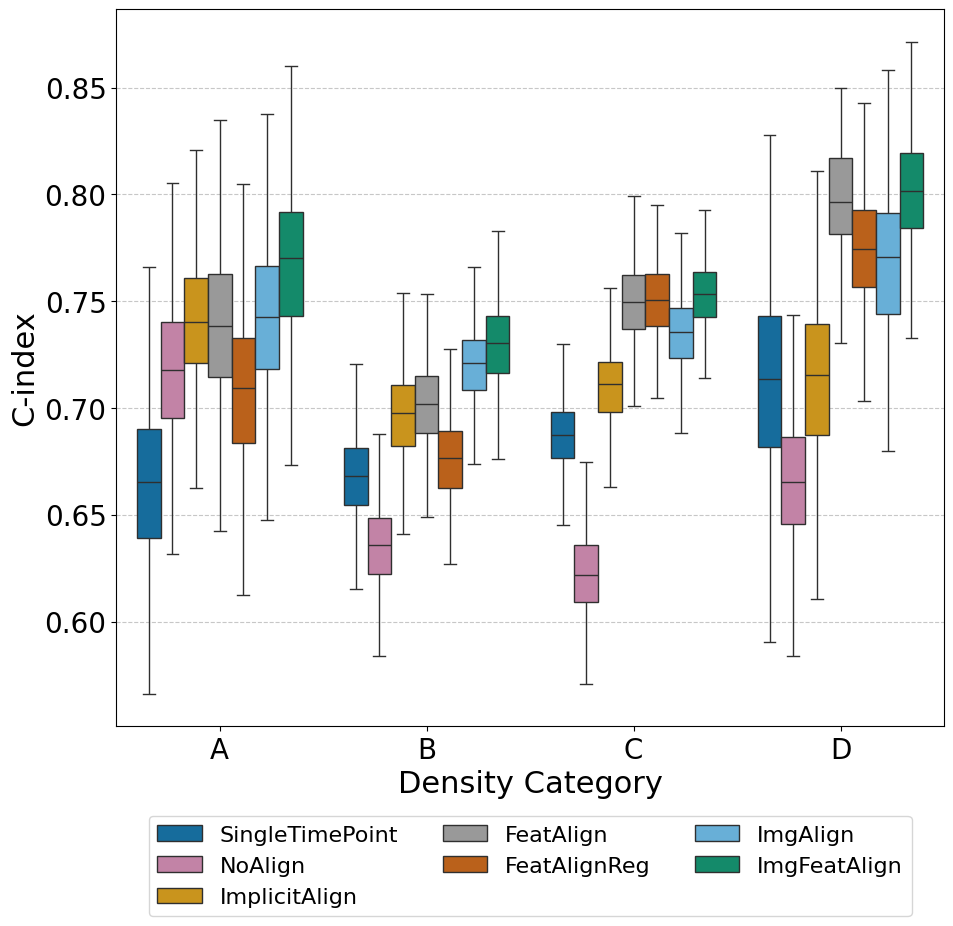}
            \caption{EMBED dataset}
        \label{fig:cindex_embed}
    \end{subfigure}
    \hfill
    % Second subfigure
    \begin{subfigure}[b]{0.48\textwidth}
        \centering
        \includegraphics[width=0.9\textwidth]{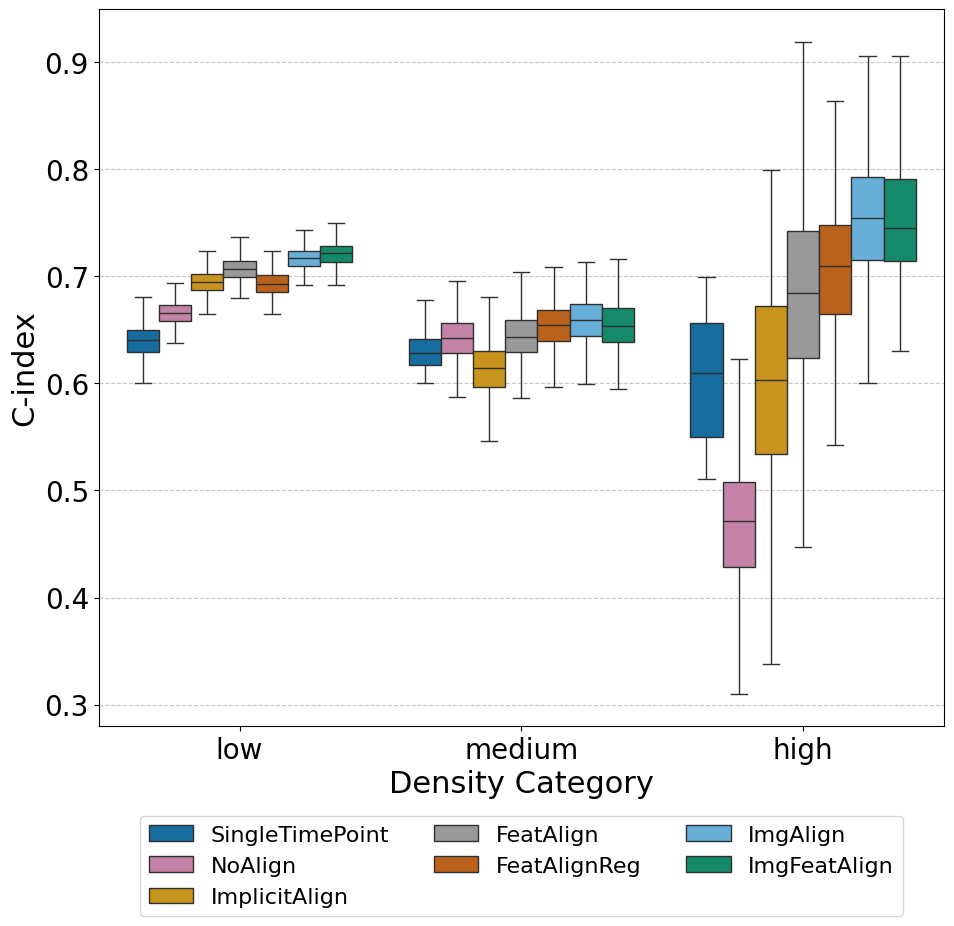}
            \caption{CSAW-CC dataset}
        \label{fig:cindex_csaw}
    \end{subfigure}
    \caption{Analysis of C-index by density category for the different alignment methods across both datasets. }
    \label{fig:cindexdensity}
\end{figure}

 Figure~\ref{fig:cindexdensity} shows C-index across breast density categories for both datasets, providing insight into model robustness under varying imaging conditions. ImgFeatAlign consistently achieves the highest C-index across nearly all density levels, demonstrating strong generalization. Among feature-based approaches, FeatAlign outperforms FeatAlignReg, suggesting that regularization, while smoothing transformations, can reduce predictive accuracy. Implicit methods perform reasonably in lower-density breasts but struggle in denser tissue. Image-based methods, particularly ImgFeatAlign, show improved performance with increasing density, indicating that spatial alignment at the image level effectively captures complex structural variations for more accurate risk prediction.
 
\subsection{Influence of the regularization on the Risk prediction in FeatAlignReg}\label{infreg}

\begin{table}[tbhp]
\centering
\scriptsize
\caption{1–5 year breast cancer risk prediction using different different $\beta$ values in Equation~\ref{lossfeat} in FeatAlignReg ($\lambda$ is set to $1 \times 10^{-5}$) We keep $\alpha$ fixed with 0.1 similar to FeatAlign to evaluate the influence of the regularization on the risk prediction performance. C-index and AUC values with 95\% confidence intervals. }
\label{tab:riskpredlambda}
\begin{adjustbox}{center,  max width=0.9\textwidth}
\begin{tabular}{lcc*{5}{c}c}
\toprule
\textbf{Dataset} & \textbf{$\beta$} & \textbf{C-index (\%) $\uparrow$} & \multicolumn{5}{c}{\textbf{Follow-up year AUC (\%) $\uparrow$}} &\textbf{ NJD (\%) $\downarrow$}\\
\cmidrule(lr){4-8}
& & & \textbf{1-yr} & \textbf{2-yr} & \textbf{3-yr} & \textbf{4-yr} & \textbf{5-yr} \\
\midrule

\multirow{10}{*}{\textbf{EMBED}} 

& \makecell{0.0}   & \makecell{72.2 \\ (69.5-75.5)} & \makecell{72.4 \\ (69.5-75.6)} & \makecell{72.2 \\ (69.7-74.8)} & \makecell{72.0 \\ (69.7-74.6)} & \makecell{71.4 \\ (69.1-74.0)} & \makecell{68.5 \\ (64.8-72.0)}  & \makecell{2.7096\\(2.6808-1.7383)}  \\

&  \makecell{ 0.1 } &
\makecell{ 71.5\\ (68.3-72.9)} & 
\makecell{ 71.3\\ (68.6-73.9)} & 
\makecell{71.8 \\(69.5-74.1) }&
\makecell{ 72.1\\ (70.0-74.2} & 
\makecell{ 71.4\\ (69.1-73.5) }&
\makecell{ 69.2\\(65.5-72.5) }&
\makecell{ 0.0911 \\ (0.0869-0.0949 } \\

& \makecell{ 0.5 } &
\makecell{ 70.8\\ (68.7-73.2) } & 
\makecell{ 70.7\\(67.9-73.4) } & 
\makecell{ 71.0\\ (68.7-73.4) }&
\makecell{ 71.3\\ (69.2-73.6)}&
\makecell{ 70.3 \\ (68.0-72.5)} & 
\makecell{ 68.2\\ (64.7-71.6)} &
\makecell{ 0.0357 \\(0.0332-0.0383) } \\

& \makecell{ 1 } &
 \makecell{70.6 \\ (67.8-73.2)} & \makecell{71.2 \\ (68.3-74.3)} & \makecell{71.3 \\ (68.6-74.0)} & \makecell{70.7 \\ (68.2-73.5)
} & \makecell{70.7 \\ (68.3-73.3)} &  \makecell{65.7 \\ (61.7-69.6)} &  \makecell{0.0385 \\(0.0171-0.0791)}  \\

  &  \makecell{ 5  } &
\makecell{ 69.5\\ (67.2-71.7) } &
\makecell{ 69.4\\(66.5-72.1) } & 
\makecell{ 70.6\\ (68.3-73.0)} & 
\makecell{ 70.2\\ (67.9-72.4)}&
\makecell{ 69.3\\ (66.9-71.4) } &
\makecell{ 68.0\\(64.4-71.4) } & 
\makecell{0.0070 \\(0.0062-0.0079) } \\

\midrule

\multirow{10}{*}{\textbf{CSAW-CC}} 
& 0.0      & \makecell{69.1 \\ (67.0-71.1)} & \makecell{70.1 \\ (67.9-72.4)} & \makecell{69.9 \\ (67.9-71.8)} & \makecell{70.0 \\ (68.1-71.9)
} & \makecell{69.5 \\ (67.7-71.4)} & \makecell{71.6 \\ (69.4-73.8)} & \makecell{3.6970\\(3.6667-3.7270)}\\

 &  \makecell{  0.1 } &
\makecell{ 69.5\\ (67.4-71.6)} &
\makecell{ 70.1\\(67.5-72.8) } & 
\makecell{ 69.4\\ (67.4-71.5)} & 
\makecell{ 69.9\\(67.9-71.8) }&
\makecell{69.4 \\ (67.5-71.4} &
\makecell{ 71.8\\ (69.7-73.9)} & 
\makecell{ 0.0136\\(0.0119-0.0152) } \\

  &  \makecell{ 0.5  } &
\makecell{ 68.3\\ (66.3-70.3)} &
\makecell{ 69.1\\(66.7-71.7) } & 
\makecell{69.3 \\(67.1-71.0) } & 
\makecell{ 70.1\\ (68.3-71.9)}&
\makecell{ 69.0\\(67.1-71.9) } &
\makecell{ 71.9\\(69.8-74.0) } & 
\makecell{ 0.0079\\ (0.0075-0.0084} \\

 & \makecell{ 1 }  & \makecell{68.4 \\ (66.4-70.-4)} & \makecell{68.9 \\ (66.7-71.2)} & \makecell{69.2 \\ (67.4-71.1)} & \makecell{69.8 \\ (68.0-71.6)} & \makecell{69.1 \\ (67.3-71.0)} & \makecell{72.0 \\ (69.9-74.2)} & \makecell{0.0059\\(0.0049-0.0071)} \\

  &  \makecell{ 5  } &
\makecell{ 67.1\\ (65.2-69.3)} &
\makecell{ 67.9\\ (65.5-70.4) } & 
\makecell{ 67.8\\(66.9-70.8) } & 
\makecell{ 69.3\\(67.6-71.2) }&
\makecell{ 68.1\\(66.2-70.1) } &
\makecell{ 70.8\\(68.7-72.9) } & 
\makecell{ 0.0 \\ (0.0-0.0)} \\

\bottomrule
\end{tabular}
\end{adjustbox}
\end{table}

To evaluate the impact of the regularization term on risk prediction performance, we conducted experiments varying the weighting parameter $\beta$ in Equation~\ref{lossfeat}. The FeatAlign model corresponds to $\beta=0$, effectively removing regularization, while FeatAlignReg corresponds to 
$\beta=1$, which was used in our main reported results.

Table~\ref{tab:riskpredlambda} reports the 1–5 year breast cancer risk prediction performance in terms of C-index and AUC, along with the NJD as a measure of deformation field regularity. For both datasets, the FeatAlign model ($\beta=0$) achieves slightly higher predictive performance compared to some intermediate $\beta$ values, indicating that small regularization does not strongly degrade risk prediction. Increasing $\beta$ leads to a gradual reduction in predictive performance, particularly for longer follow-up years, while simultaneously improving the smoothness of the deformation fields as indicated by lower NJD values. Our main FeatAlignReg results ($\beta=1$) strike a balance between maintaining competitive risk prediction performance and enforcing smooth, physically plausible deformation fields.

These findings highlight the trade-off between regularization strength and predictive accuracy in feature-level alignment: while regularization stabilizes the deformation fields, overly strong regularization can reduce risk prediction performance.

\subsection{In-Depth Analysis on the EMBED dataset}\label{embedana}
While both datasets are used to evaluate overall model performance, the following analyses, including ROC curves, precision trends, and recall trends, focus specifically on the EMBED dataset. Compared to the CSAW-CC dataset, EMBED offers greater diversity, incorporating images from multiple imaging systems and a broader patient population. This diversity enhances the robustness and generalizability of the findings, making it a more suitable benchmark for evaluating alignment methods in real-world scenarios.

\subsubsection{ROC-AUC, Precision, and Recall Across Alignment Strategies}\label{rocaucemb}

\begin{figure}[tbh]
    \centering
    \includegraphics[width=1.0\textwidth]{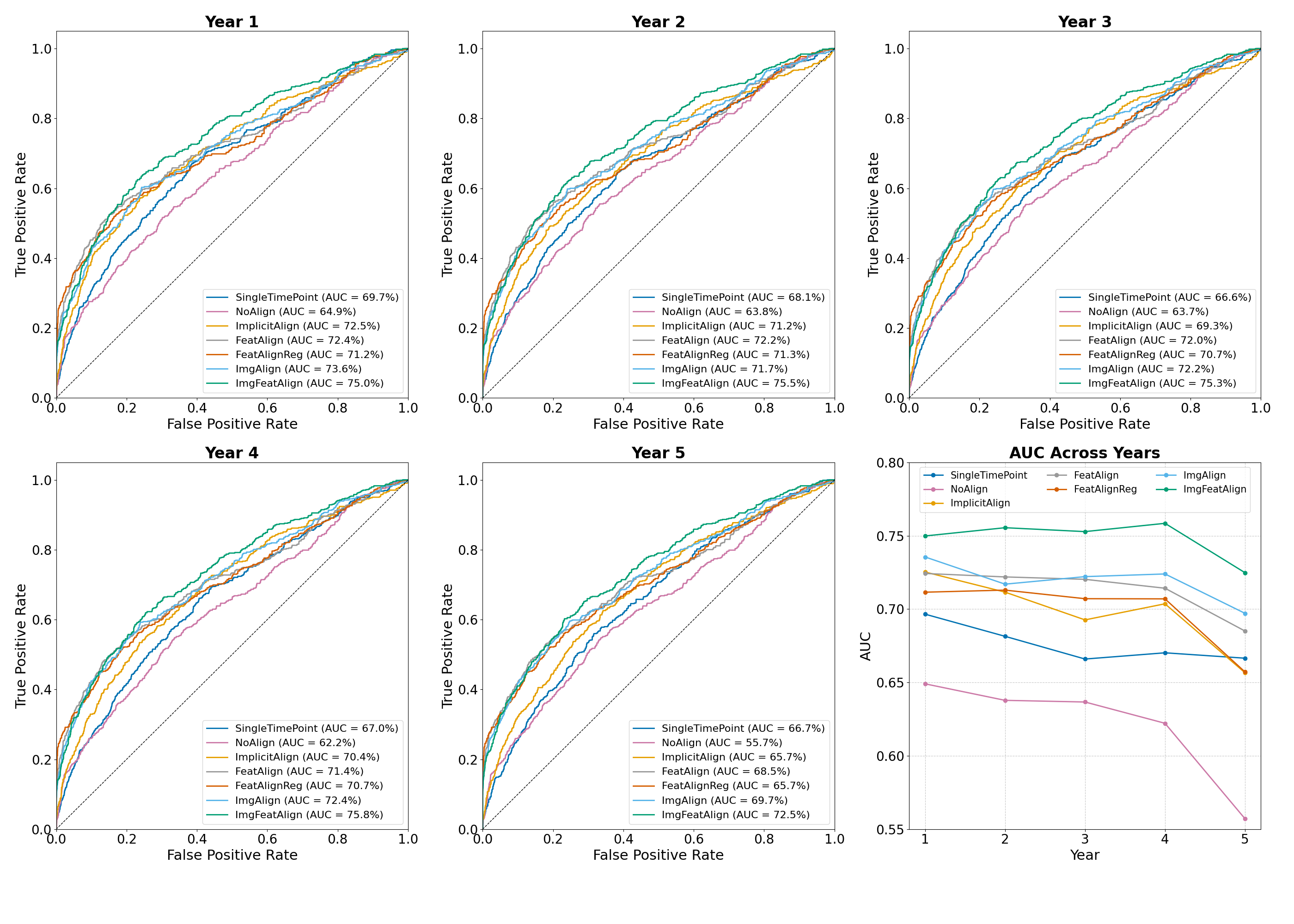}
    \caption{AUC-ROC curves for risk predictions on the EMBED dataset across 1-5 year outcomes: comparison of different alignment methods. }
    \label{fig:rocaucembed}
\end{figure}

\begin{figure}[tbh]
    \centering
    \begin{subfigure}[b]{0.48\textwidth}
        \centering
        \includegraphics[width=0.8\textwidth]{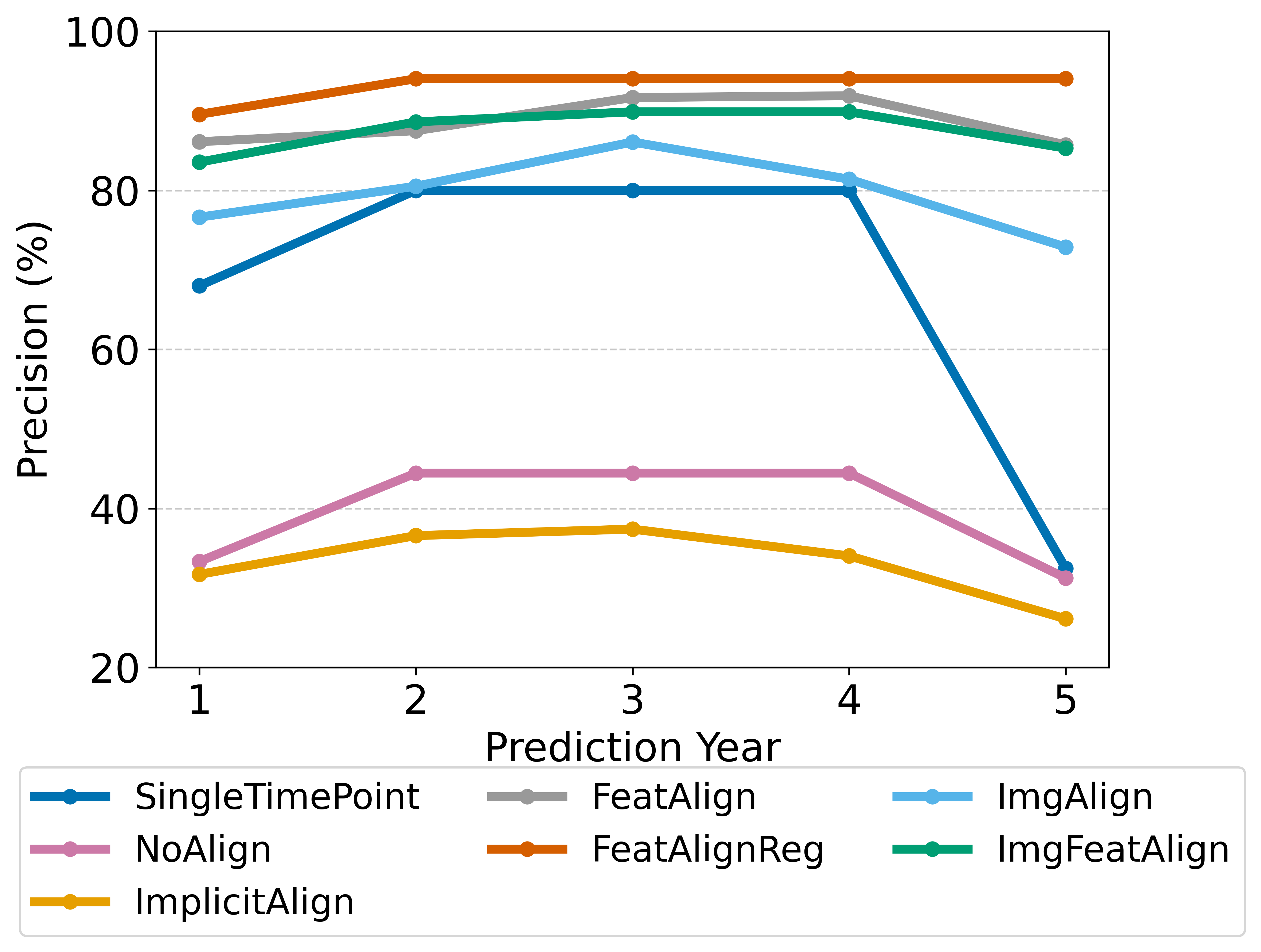}
        \caption{Precision of alignment methods across years.}
        \label{fig:precision}
    \end{subfigure}
    \hfill
    \begin{subfigure}[b]{0.48\textwidth}
        \centering
        \includegraphics[width=0.8\textwidth]{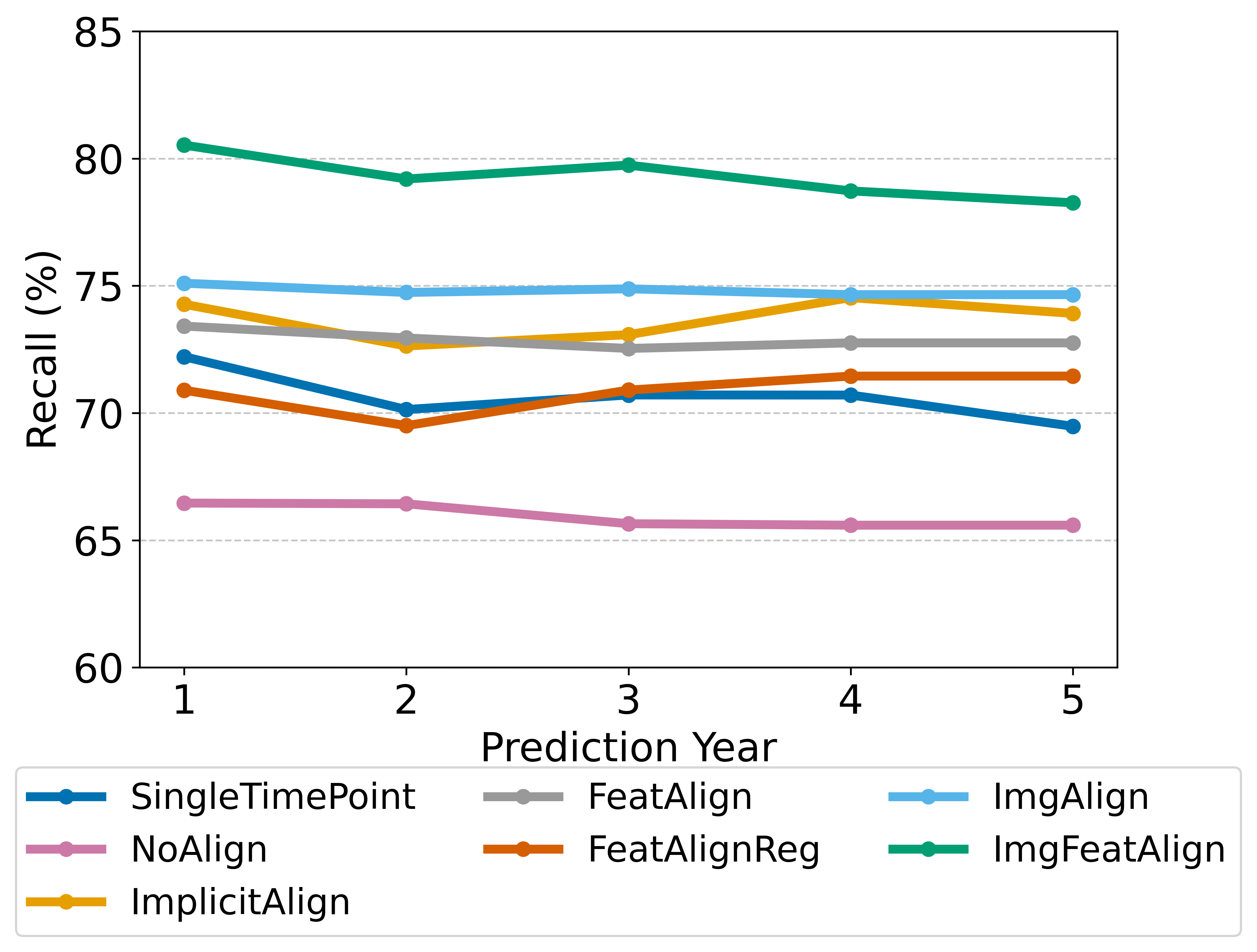}
        \caption{Recall of alignment methods across years.}
        \label{fig:recall}
    \end{subfigure}
    \caption{Comparison of different alignment methods in predicting time to cancer development. (a) Precision, (b) Recall.}
    \label{fig:precision_recall}
\end{figure}

Figure~\ref{fig:rocaucembed} shows ROC-AUC performance on the EMBED dataset over a five-year follow-up for different alignment strategies. ImgFeatAlign consistently achieves the highest AUC across all years, demonstrating strong predictive performance and temporal stability. ImgAlign ranks second, highlighting the benefits of image-based alignment.

Figures~\ref{fig:precision} and~\ref{fig:recall} compare precision and recall for time-to-cancer prediction. Explicit alignment methods (FeatAlign, FeatAlignReg, ImgAlign, ImgFeatAlign) consistently outperform implicit alignment, NoAlign, and SingleTimePoint. FeatAlignReg achieves the highest precision, reflecting the advantage of anatomically constrained deformations, while ImgFeatAlign attains the highest recall, emphasizing its strength in identifying high-risk patients. ImgAlign performs slightly below ImgFeatAlign, and feature-level methods show lower recall. NoAlign exhibits the poorest performance across all metrics.

Overall, explicit alignment methods, particularly ImgFeatAlign, provide the strongest predictive performance across ROC-AUC, precision, and recall, combining temporal stability with accurate risk stratification to support effective early detection and clinical decision-making.

 \subsubsection{Density-Stratified Performance Across Time Horizons}\label{shortlongterm}
\begin{figure}[tbph]
    \centering
    \begin{subfigure}[b]{0.46\textwidth}
        \centering
        \includegraphics[width=0.95\textwidth]{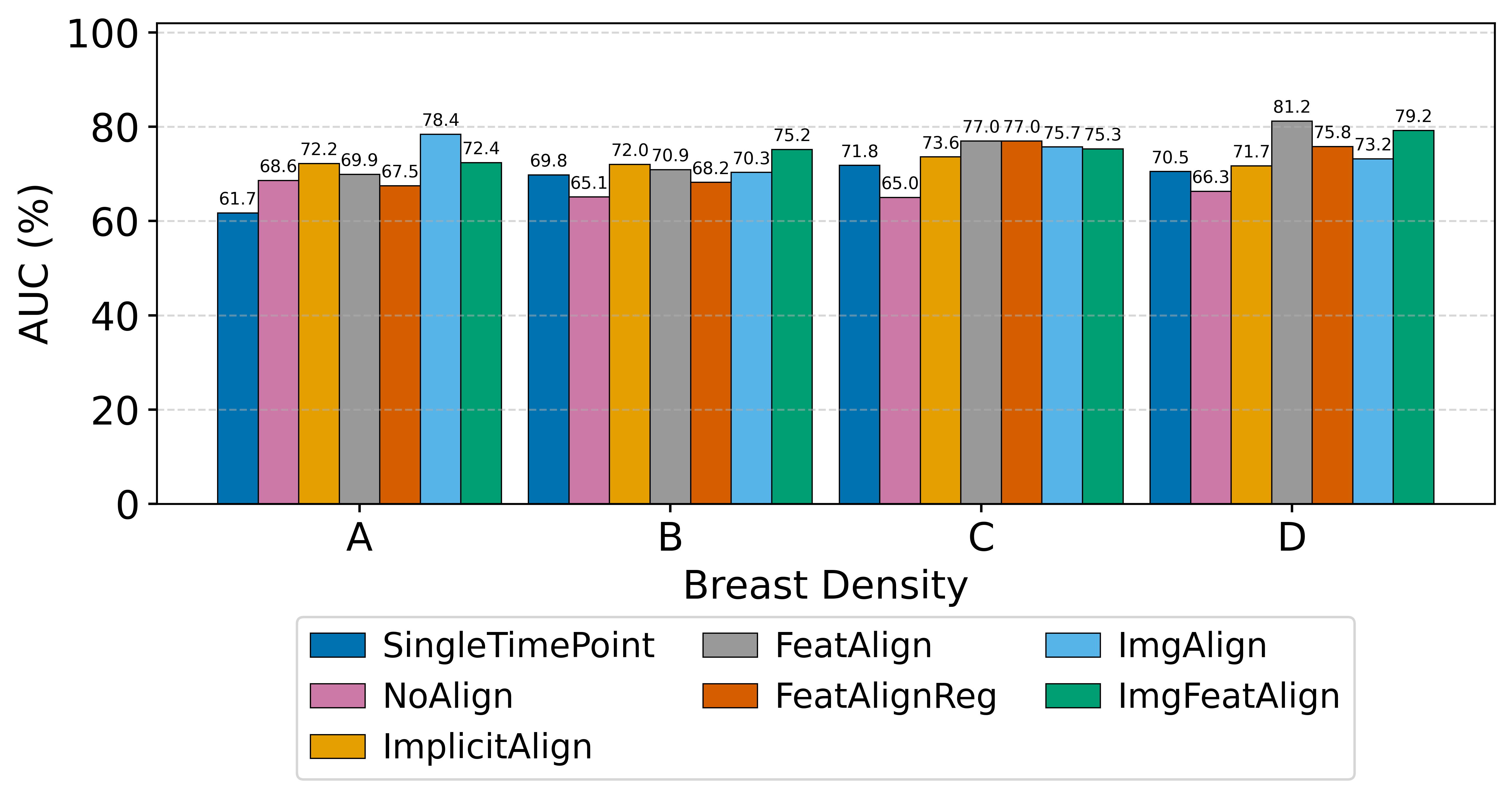}
        \caption{Short-term risk (1-year).}
        \label{fig:shorttermrisk}
    \end{subfigure}
    \hfill
    \begin{subfigure}[b]{0.46\textwidth}
        \centering
        \includegraphics[width=0.95\textwidth]{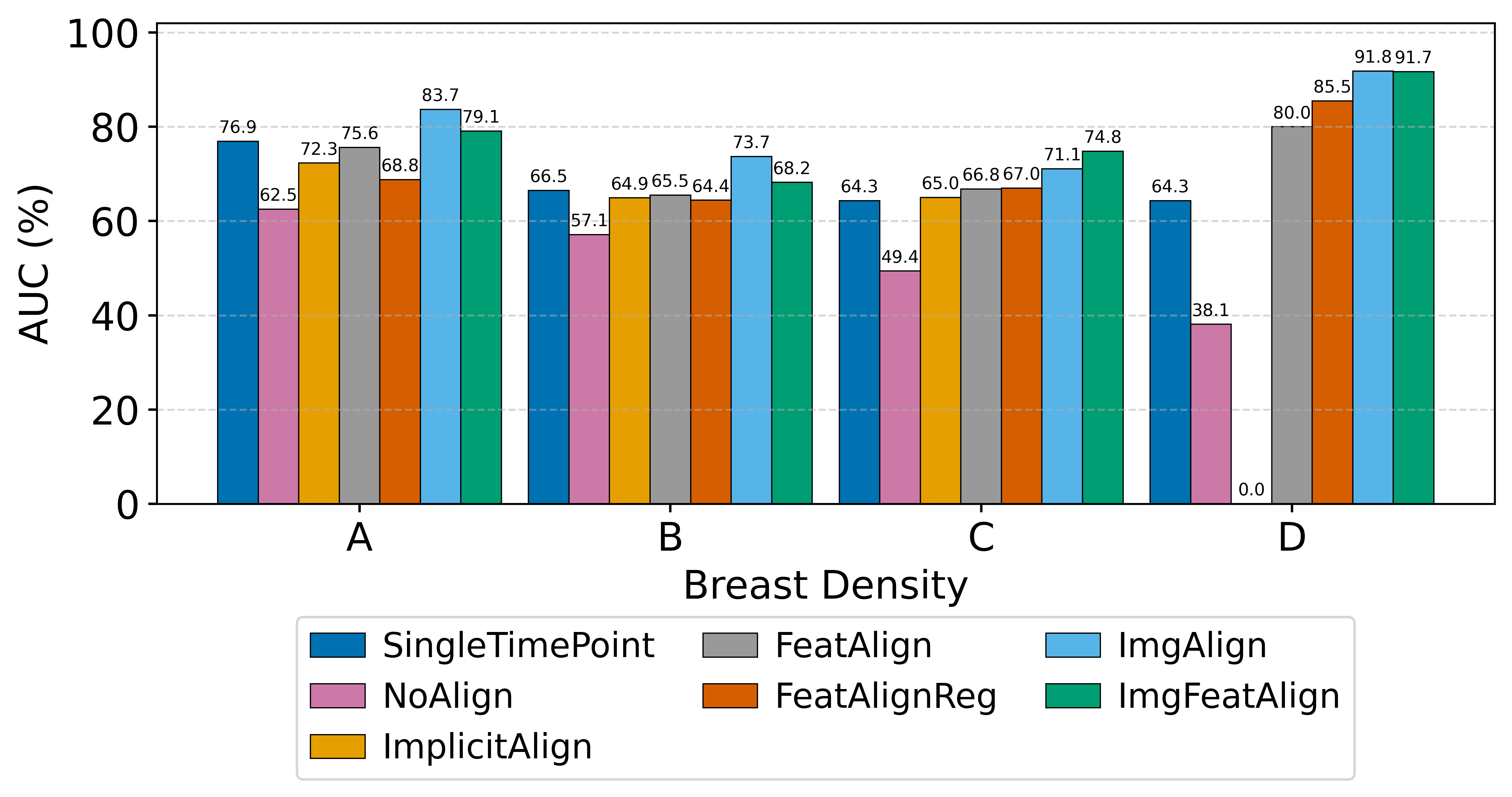}
        \caption{Long-term risk (5-year).}
        \label{fig:longtermrisk}
    \end{subfigure}

    \caption{Density-stratified AUC performance of different alignment methods for predicting time to cancer development. (a) Short-term (1-year) risk prediction. (b) Long-term (5-year) risk prediction.}
    \label{fig:density_auc}
\end{figure}
 
Figure~\ref{fig:shorttermrisk} and~\ref{fig:longtermrisk} summarize AUC performance for short- and long-term risk prediction across breast density categories. For short-term prediction, image-level alignment methods, particularly ImgFeatAlign, consistently achieve the highest performance across most densities, with ImgAlign performing competitively in lower-density cases (Density A). Feature-level methods (FeatAlign and FeatAlignReg) show moderate performance, while NoAlign exhibits the lowest AUCs and ImplicitAlign remains slightly better but inferior to explicit methods.

For long-term prediction, alignment is even more critical. Image-based methods maintain high AUCs across all categories, including Density D, where the dataset is particularly limited (12 images). ImplicitAlign fails in this category, while feature-level methods perform moderately in lower-density categories but are less robust in dense tissue. Notably, FeatAlignReg can outperform FeatAlign in higher-density cases due to its anatomically constrained deformations. These results emphasize the importance of robust spatial alignment, especially image-level strategies, for reliable long-term risk prediction in complex or data-scarce settings.

 \subsubsection{Stratified Performance Across Clinical and Demographic Subgroups}\label{subgroupeval}
 
\begin{figure}[tb]
    \centering
        \centering
        \includegraphics[width=0.30\textwidth]{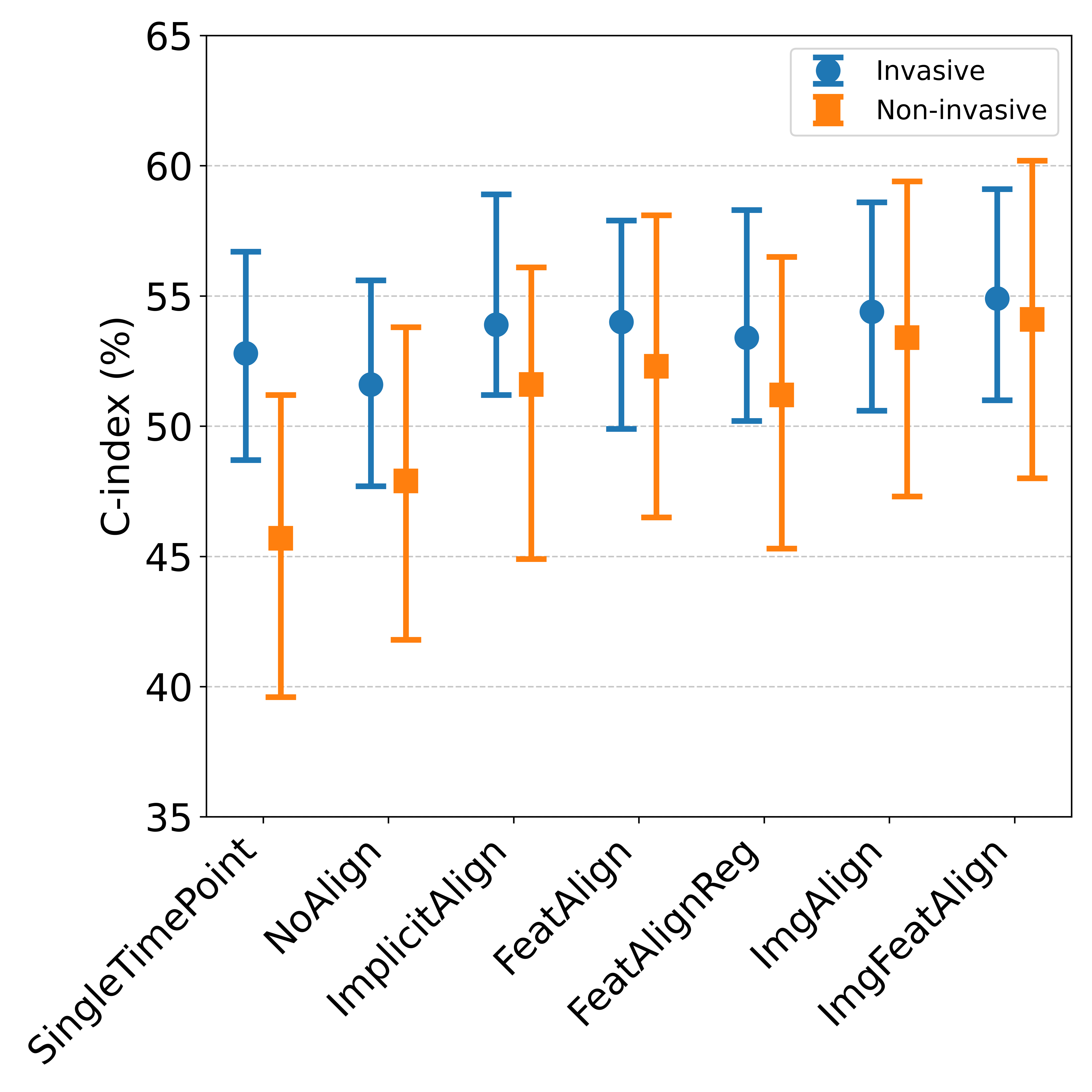}
    \caption{Comparison of model performance for non-invasive and invasive cancer. C-index values with 95\% confidence intervals are shown. }
    \label{fig:EMBED_C_indexcancertypes}
\end{figure}

\begin{figure}[tb]
    \centering
    \includegraphics[width=0.50\textwidth]{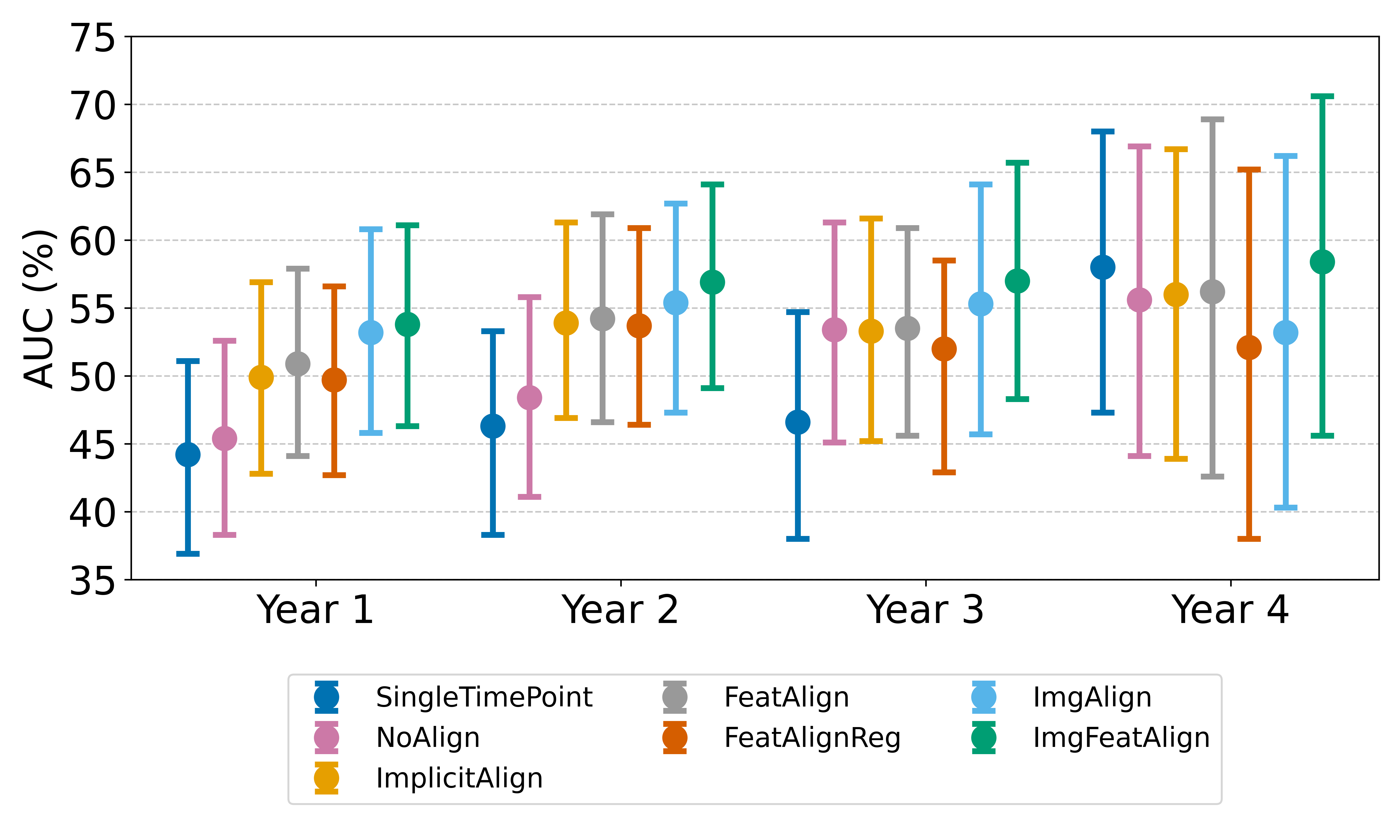}
    \caption{AUC for non-invasive cancer for each year with 95\% confidence intervals. }
    \label{fig:EMBED_AUC_noninvasive}
\end{figure}

\begin{figure}[tb]
    \centering
    \includegraphics[width=0.50\textwidth]{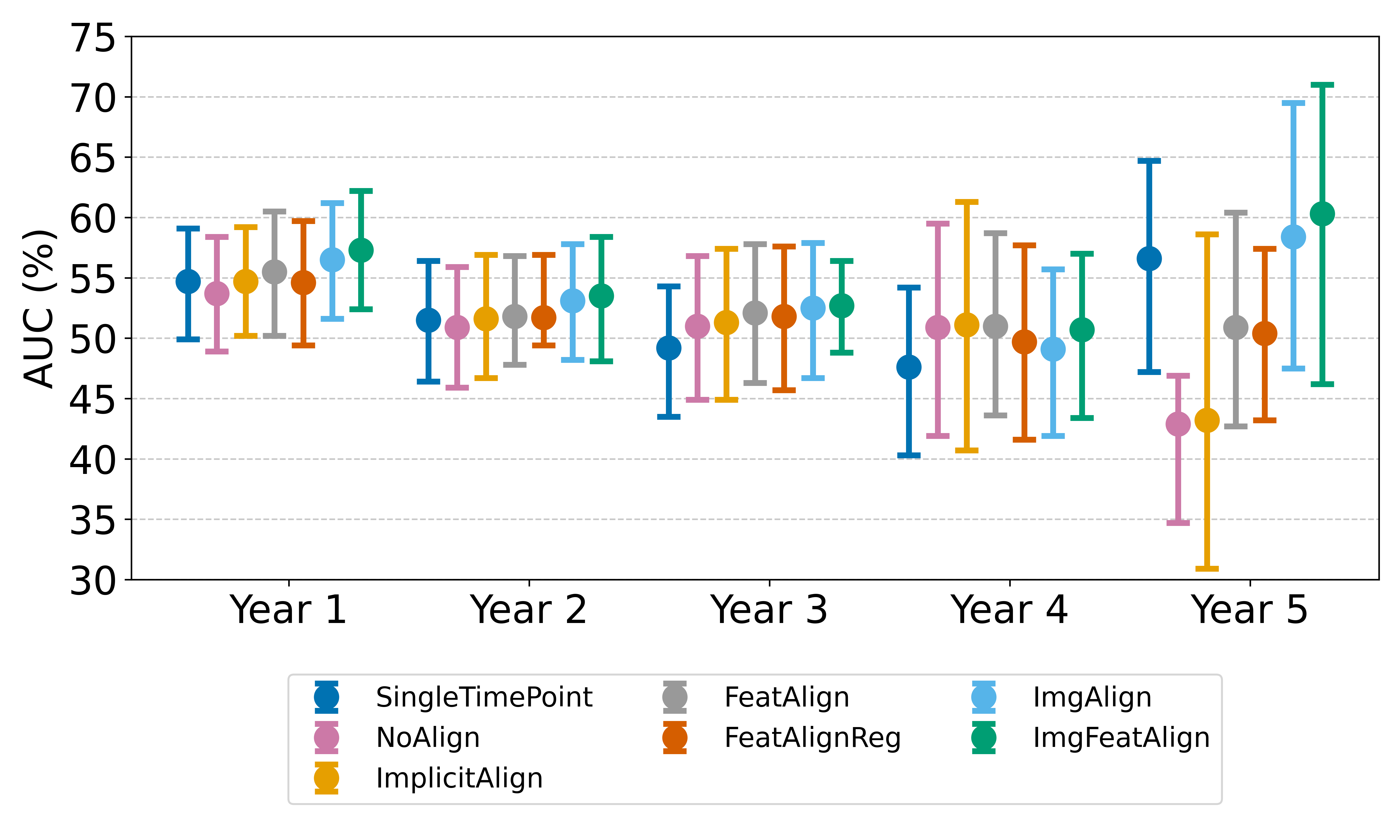 }
    \caption{AUC for invasive cancer for each year with 95\% confidence intervals. }
    \label{fig:EMBED_AUC_invasive}
\end{figure}

\begin{figure}[tb]
    \centering
    \includegraphics[width=0.55\textwidth]{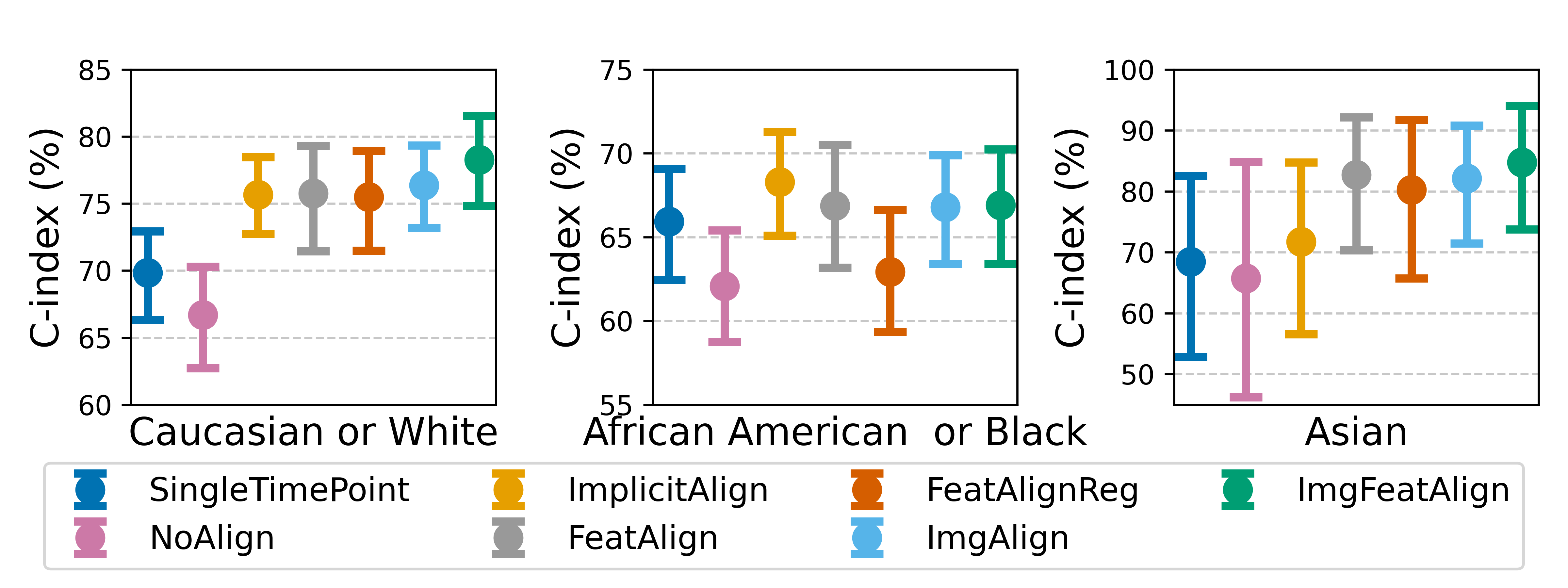}
    \caption{C-index values with 95\% confidence intervals for different race categories.}
    \label{fig:EMBED_C_index_race}
\end{figure}

\begin{figure}[tb]
    \centering
    \includegraphics[width=0.55\textwidth]{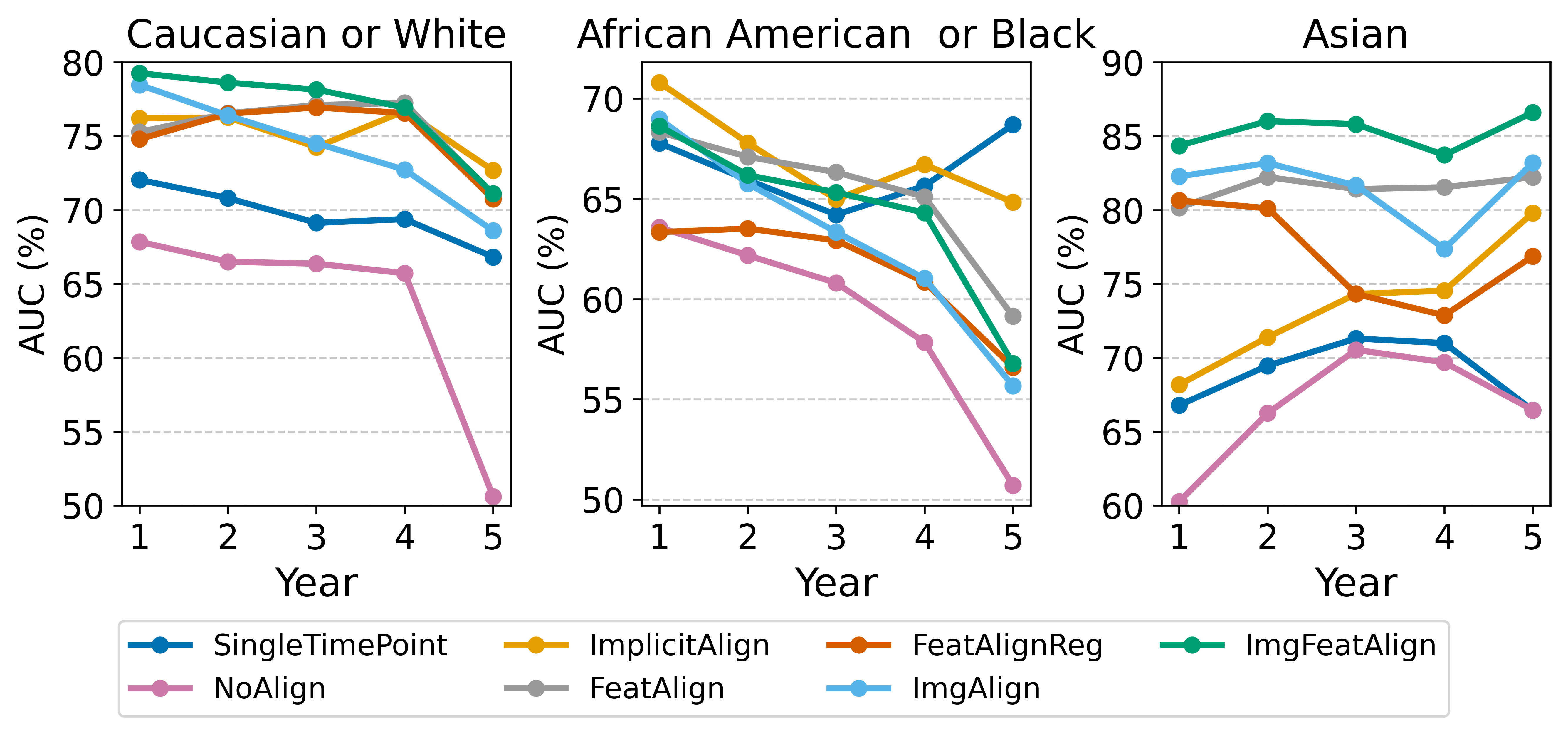}
    \caption{AUC for the different races. }
    \label{fig:EMBED_AUC_race}
\end{figure}

We further evaluated the robustness of alignment methods across cancer subtypes and racial groups. For cancer type (Figures~\ref{fig:EMBED_C_indexcancertypes}–\ref{fig:EMBED_AUC_invasive}), ImgFeatAlign consistently achieves the highest C-index and AUC for both non-invasive and invasive cancers, followed by ImgAlign and FeatAlign. Trends are consistent across follow-up years, with explicit alignment yielding more stable and robust predictions than single-timepoint or non-aligned approaches.

For race (Figures~\ref{fig:EMBED_C_index_race}–\ref{fig:EMBED_AUC_race}), alignment-based models improve performance across all groups. SingleTimePoint outperforms NoAlign, while ImgFeatAlign achieves the highest C-index and AUC for Caucasian/White and Asian cohorts. For African American/Black patients, performance differences are smaller, particularly in early follow-up years, with ImplicitAlign occasionally performing best in the short term. Feature-level regularization (FeatAlignReg) does not consistently improve results over FeatAlign. Overall, explicit alignment methods, especially ImgFeatAlign, demonstrate robust and stable discrimination across cancer types and racial groups, highlighting both their generalizability and the importance of evaluating model performance in underrepresented populations.

\subsection{Comparison of Deformation Field Quality}\label{deformqual}
\begin{table}[t]
\centering
\scriptsize
\caption{Quantitative comparison of deformation field quality for CC and MLO views separately and combined. NJD: percentage of negative Jacobian determinants. Std. dev. Jacobian: standard deviation of the Jacobian determinant.}
\label{tab:deformqual}
\begin{adjustbox}{center, max width=0.9\textwidth}
\begin{tabular}{llcccccc}
\toprule
\textbf{Dataset} & \textbf{Method} 
& \multicolumn{3}{c}{\textbf{NJD (\%) $\downarrow$}}
& \multicolumn{3}{c}{\textbf{Std. dev. Jacobian $\downarrow$}} \\
\cmidrule(lr){3-5}
\cmidrule(lr){6-8}
& & \textbf{CC} & \textbf{MLO} & \textbf{Combined}
& \textbf{CC} & \textbf{MLO} & \textbf{Combined} \\
\midrule

\multirow{3}{*}{\textbf{EMBED}}
& \makecell{FeatAlign}
& \makecell{2.9135 \\ (2.8737-2.9540)} & \makecell{2.5218 \\ (2.4835-2.5562) } & \makecell{2.7096\\(2.6808-1.7383)}
& \makecell{0.4561 \\ (0.4588-0.4545)} & \makecell{0.46535 \\ (0.4615-0.4696)}  & \makecell{0.4599\\(0.4562-0.4640)} \\

& \makecell{FeatAlignReg}
& \makecell{0.0415 \\ (0.0383-0.0451)} & \makecell{0.0346 \\ (0.0323-0.0372)} & \makecell{0.0385\\(0.0171-0.0791)}
&\makecell{\textbf{0.0778} \\ \textbf{(0.0754-0.0805)}} &  \makecell{\textbf{0.0914} \\ \textbf{(0.0884-0.0946)}} &  \makecell{\textbf{0.0889}\\\textbf{(0.0855-0.0931)}} \\

& \makecell{ImgAlign \\ ImgFeatAlign}
& \makecell{\textbf{0.0014} \\ \textbf{(0.0013-0.0015)}} & \makecell{\textbf{0.0012} \\ \textbf{(0.0009-0.0016)}} & \makecell{\textbf{0.0013}\\\textbf{(0.0012-0.0015)}}
& \makecell{0.1215 \\ (0.1209-0.1222} & \makecell{0.1401 \\ (0.1395-0.1408)} & \makecell{0.1408\\(0.1403-0.1413)} \\
\midrule

\multirow{3}{*}{\textbf{CSAW-CC}}
& \makecell{FeatAlign}
& \makecell{3.9002 \\ (3.8582-3.9407) } & \makecell{3.4685 \\ (3.4293-3.5094)} & \makecell{3.6970\\(3.6667-3.7270)}
& \makecell{0.5759 \\ (0.5639-0.5881)} & \makecell{0.6121 \\ (0.5954-0.6309)} & \makecell{0.5965\\(0.5897-0.6030)} \\

& \makecell{FeatAlignReg}
& \makecell{0.0061 \\ (0.0046-0.0079)} & \makecell{0.0053 \\ (0.0041-0.0067)} & \makecell{0.0059\\(0.0049-0.0071)}
& \makecell{\textbf{0.1115} \\ \textbf{(0.1105-0.1117)}} & \makecell{0.1517 \\ (0.1511-0.1524)} & \makecell{\textbf{0.1329}\\\textbf{(0.1323-0.1334)}} \\

& \makecell{ImgAlign \\  ImgFeatAlign}
& \makecell{\textbf{0.0011} \\ \textbf{(0.0009-0.0014)}} & \makecell{ \textbf{0.0007} \\ \textbf{(0.0005-0.0008) }}  & \makecell{\textbf{0.0008}\\\textbf{(0.0006-0.0009)}}
& \makecell{0.1282 \\ (0.1276-0.1287) } & \makecell{\textbf{0.1437} \\ \textbf{(0.1431-0.1444)} } & \makecell{0.1402\\(0.1397-0.1406)} \\
\bottomrule
\end{tabular}
\end{adjustbox}
\end{table}

Table~\ref{tab:deformqual} reports deformation field quality across methods, measured by the Percentage of Negative Jacobian Determinants and the standard deviation of the Jacobian determinant. Image-based methods, ImgAlign and ImgFeatAlign, achieve the highest quality, with the lowest NJD and minimal variability, producing smooth, anatomically plausible transformations. FeatAlign exhibits poor deformation quality, with elevated NJD and high variability, while introducing regularization in FeatAlignReg improves both metrics, though it remains slightly below image-based performance.  Importantly, deformation quality varies by view: across both datasets, NJD is generally lower for MLO, indicating fewer regions with non-physical deformations, while the standard deviation is lower for CC, reflecting more uniform and stable transformations. This view-specific pattern highlights the distinct challenges posed by different mammographic views.

Overall, these results underscore the superiority of image-based registration and the importance of regularization in improving feature-based alignment.

\begin{figure}[t]
    \centering
    % Row 1
    \begin{subfigure}{0.22\textwidth}
        \centering
        \includegraphics[height=4cm]{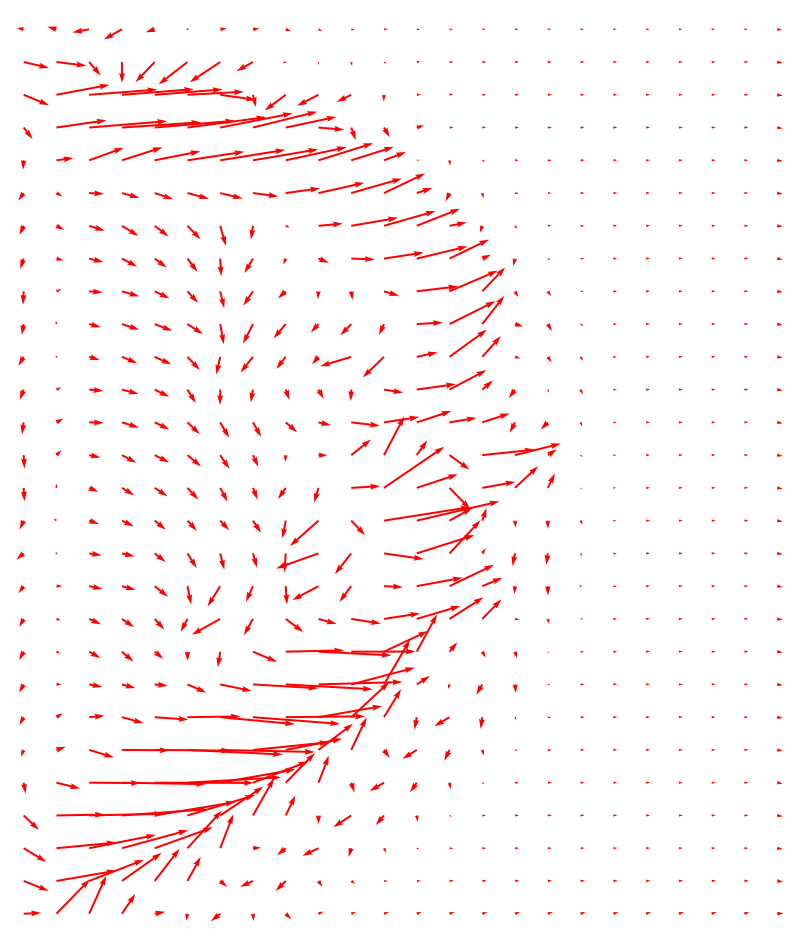}
        \caption{FeatAlign}
        \label{fig:featalign1}
    \end{subfigure}
    \hfill
    \begin{subfigure}{0.22\textwidth}
        \centering
        \includegraphics[height=4cm]{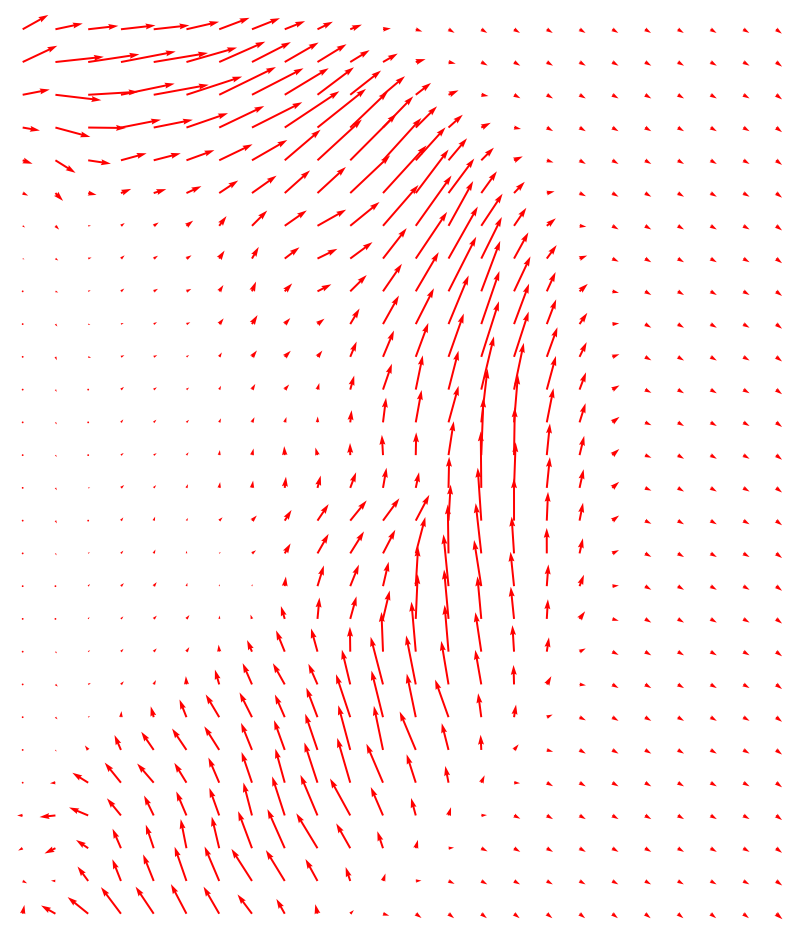}
        \caption{FeatAlignReg}
        \label{fig:featalignreg1}
    \end{subfigure}
    \hfill
    \begin{subfigure}{0.27\textwidth}
        \centering
        \includegraphics[height=4cm]{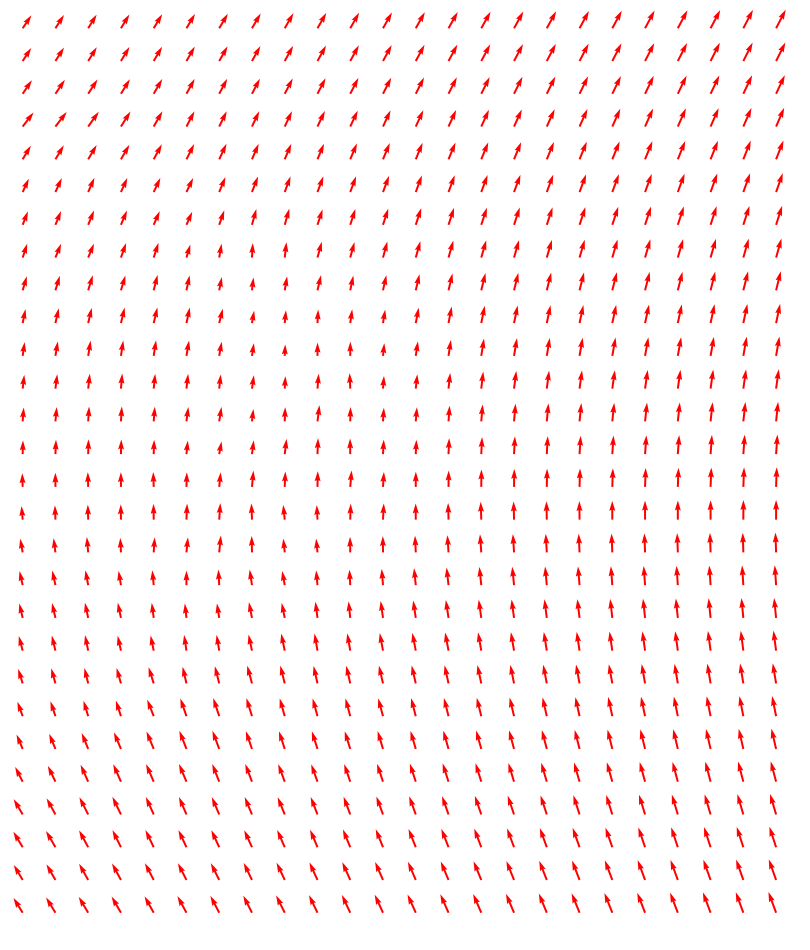}
        \caption{ImgFeatAlign/ImgAlign}
        \label{fig:imgalign1}
    \end{subfigure}

    % Row 2
    \begin{subfigure}{0.2\textwidth}
        \centering
        \includegraphics[height=4cm]{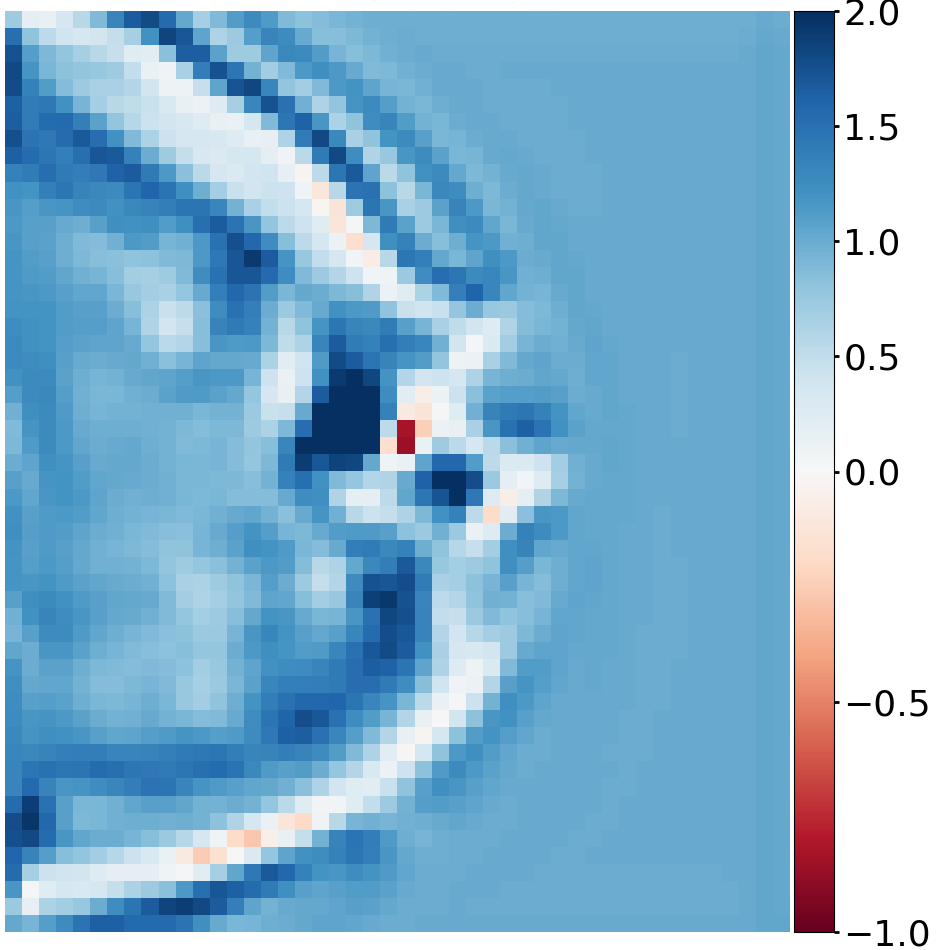}
        \caption{FeatAlign}
        \label{fig:featalign2}
    \end{subfigure}
    \hfill
    \begin{subfigure}{0.2\textwidth}
        \centering
        \includegraphics[height=4cm]{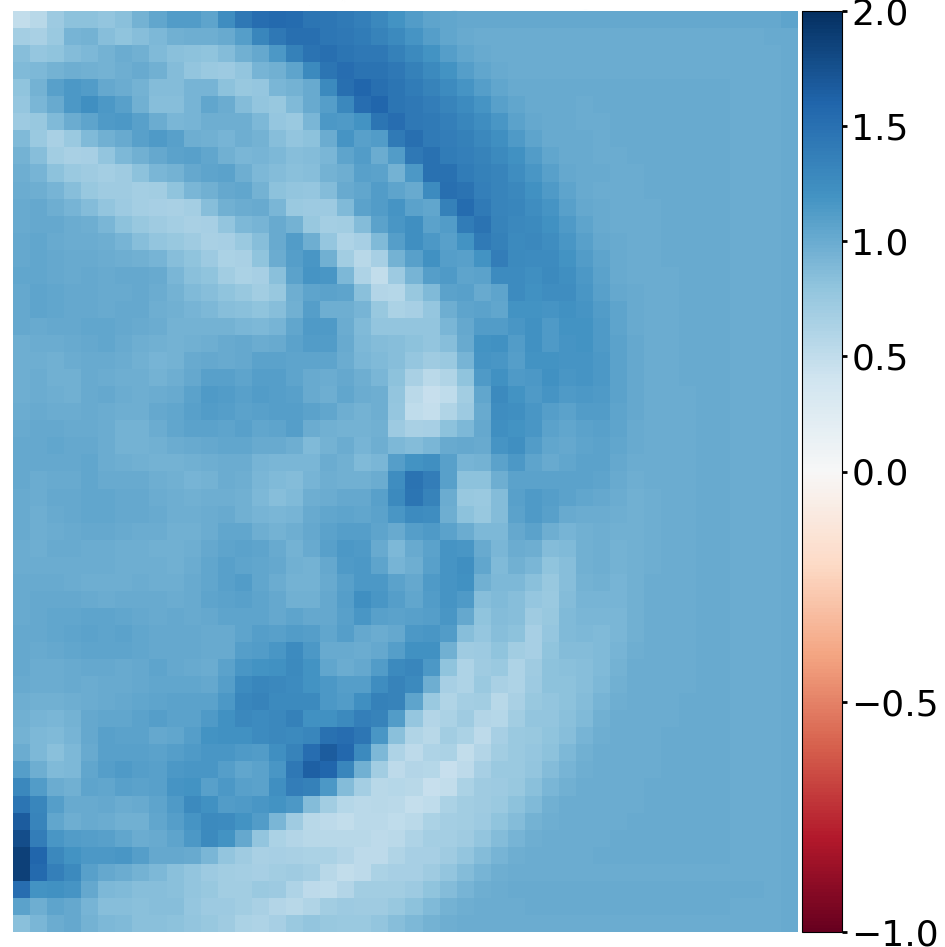}
        \caption{FeatAlignReg}
        \label{fig:featalignreg2}
    \end{subfigure}
    \hfill
    \begin{subfigure}{0.26\textwidth}
        \centering
        \includegraphics[height=4cm]{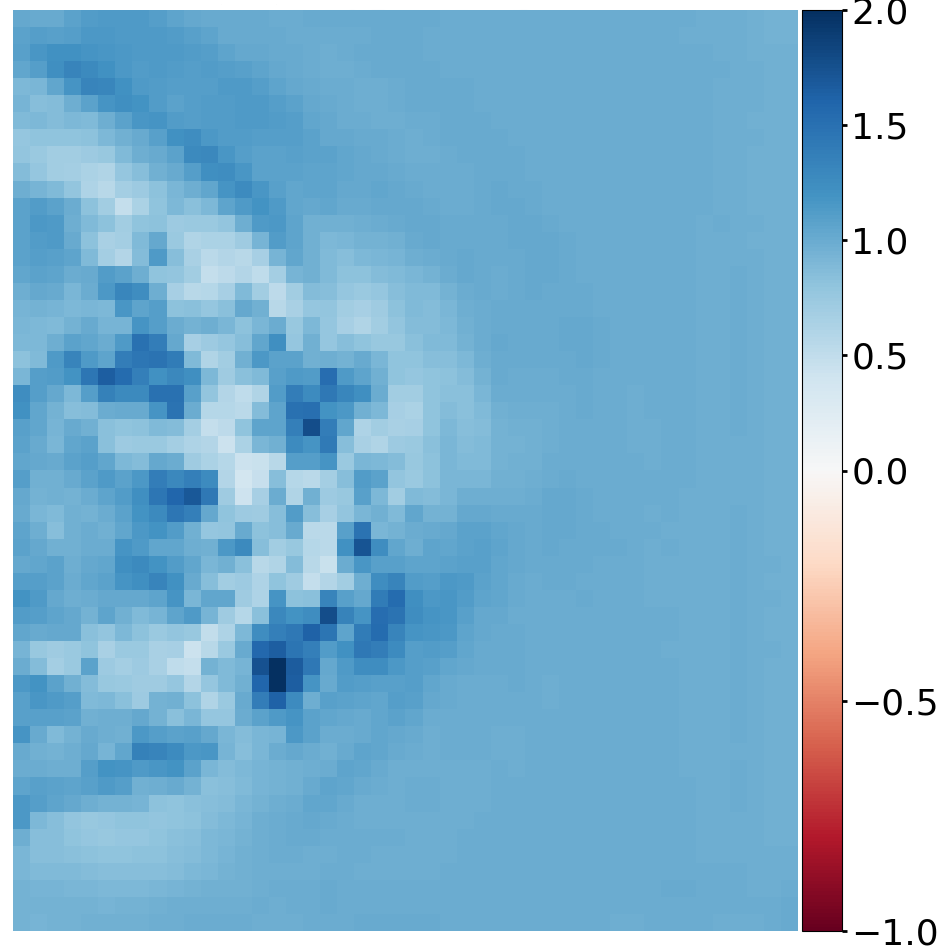}
        \caption{ImgFeatAlign/ImgAlign}
        \label{fig:imgalign2}
    \end{subfigure}
    \caption{Comparison of deformation field quality. The top row shows displacement vectors, and the bottom row displays Jacobian determinant maps (white/blue: valid; orange/red: invalid non-invertible regions).}
    \label{fig:deformationfield}
\end{figure}

\begin{figure}[t]
    \centering
    % Row 1
    \begin{subfigure}{0.22\textwidth}
        \centering
        \includegraphics[height=4cm]{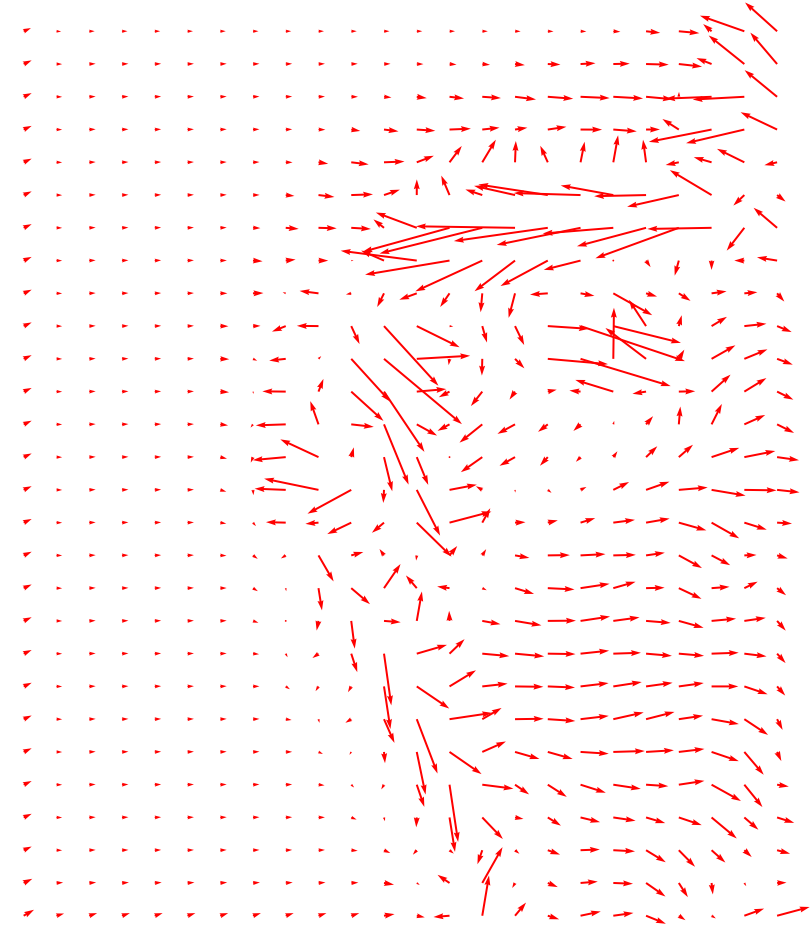}
        \caption{FeatAlign}
        \label{fig:featalign3}
    \end{subfigure}
    \hfill
    \begin{subfigure}{0.22\textwidth}
        \centering
        \includegraphics[height=4cm]{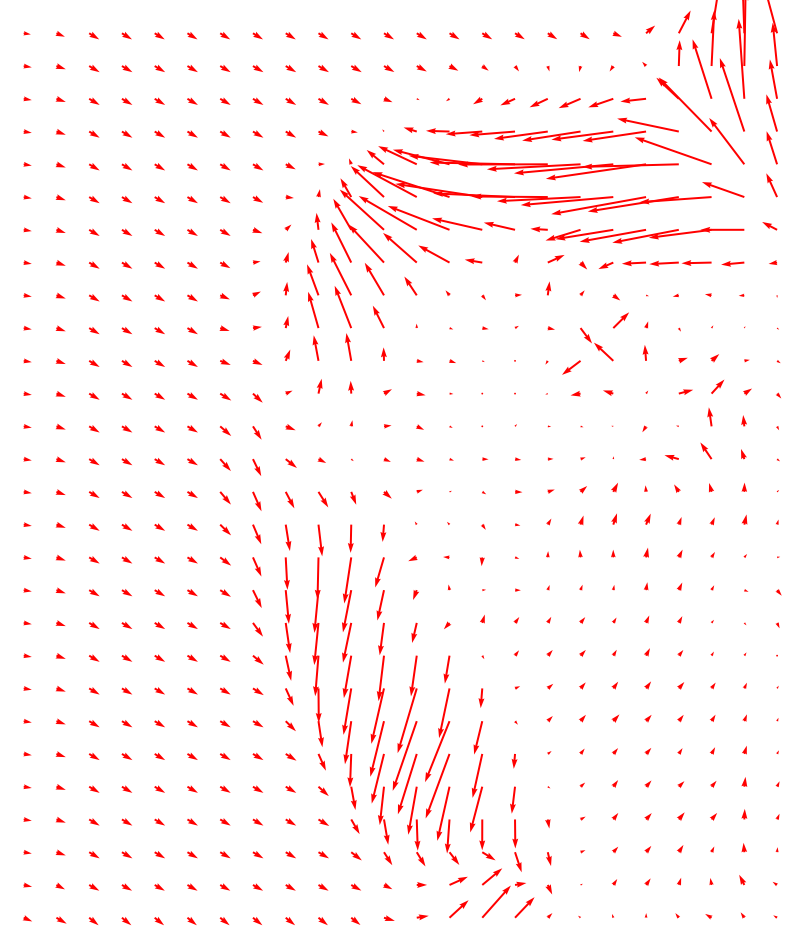}
        \caption{FeatAlignReg}
        \label{fig:featalignreg3}
    \end{subfigure}
    \hfill
    \begin{subfigure}{0.27\textwidth}
        \centering
        \includegraphics[height=4cm]{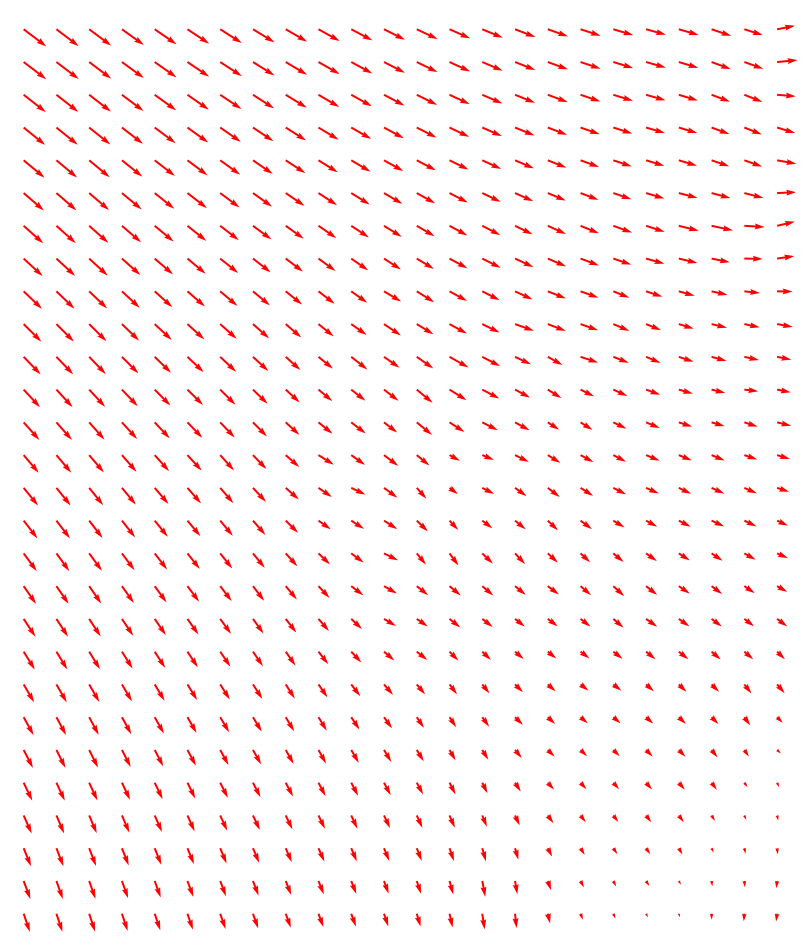}
        \caption{ImgFeatAlign/ImgAlign}
        \label{fig:imgalign3}
    \end{subfigure}
    
    % Row 2
    \begin{subfigure}{0.2\textwidth}
        \centering
        \includegraphics[height=4cm]{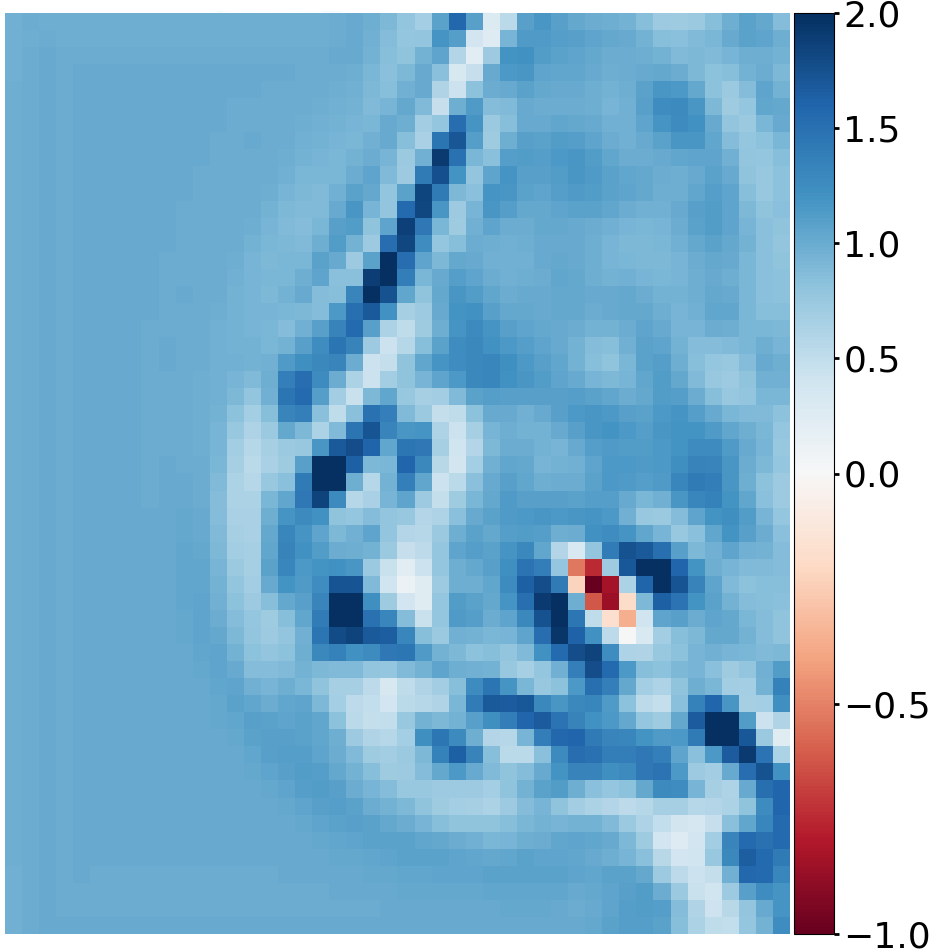}
        \caption{FeatAlign}
        \label{fig:featalign4}
    \end{subfigure}
    \hfill
    \begin{subfigure}{0.2\textwidth}
        \centering
        \includegraphics[height=4cm]{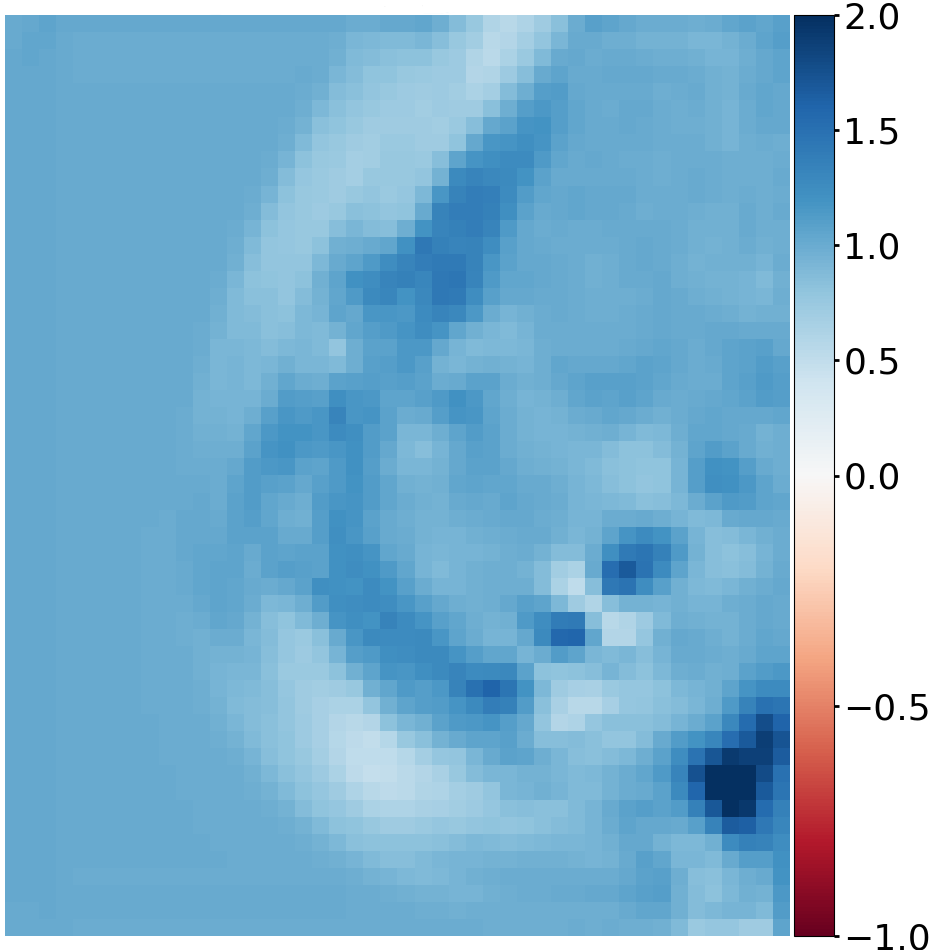}
        \caption{FeatAlignReg}
        \label{fig:featalignreg4}
    \end{subfigure}
    \hfill
    \begin{subfigure}{0.26\textwidth}
        \centering
        \includegraphics[height=4cm]{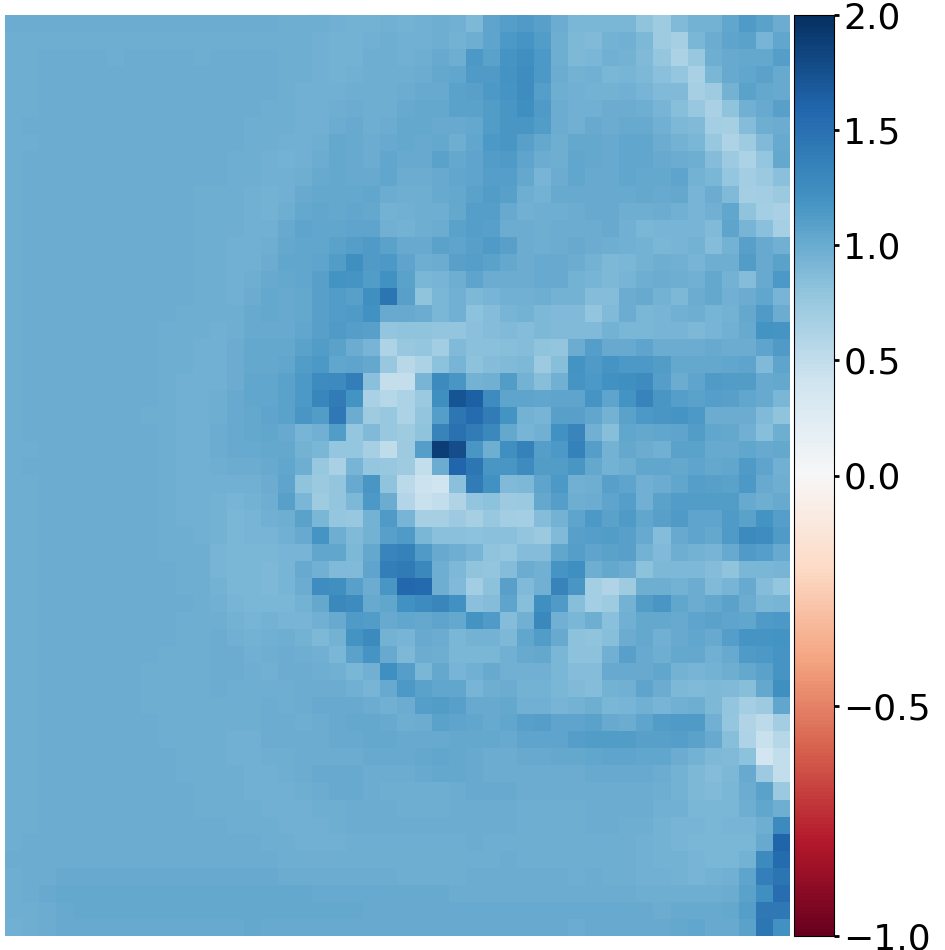}
        \caption{ImgFeatAlign/ImgAlign}
        \label{fig:imgalign4}
    \end{subfigure}
    \caption{Comparison of deformation field quality. The top row shows displacement vectors, and the bottom row displays Jacobian determinant maps (white/blue: valid; orange/red: invalid non-invertible regions).}
    \label{fig:deformationfield2}
\end{figure}

Figures~\ref{fig:deformationfield} and~\ref{fig:deformationfield2} illustrate the deformation fields produced by the registration methods, showing displacement vectors (top) and Jacobian determinant maps (bottom).  FeatAlign generates irregular, noisy, and unstructured fields, with regions of negative Jacobian determinants indicating invalid, non-invertible transformations. Regularization in FeatAlignReg improves smoothness and invertibility, though the transformations remain localized and lack the global coherence of image-based methods. In contrast, ImgAlign and ImgFeatAlign, based on MammoRegNet, produce smooth, spatially consistent, and anatomically meaningful fields with no invalid regions. These globally coherent transformations, supported by MammoRegNet’s joint affine and deformable components, yield robust and interpretable longitudinal alignment, as further illustrated in Figures~\ref{fig:regmodel1} and~\ref{fig:regmodel2}.

\section{Discussion}\label{discussion}
This study demonstrates the clear superiority of image-based alignment methods, particularly ImgFeatAlign, for longitudinal breast cancer risk prediction. Across multiple evaluation criteria, including predictive accuracy, precision, recall, and deformation field quality, ImgFeatAlign consistently outperforms feature-based approaches. These results underscore the critical role of precise spatial alignment in modeling temporal changes in mammography.

The strength of image-based methods lies in their ability to preserve fine-grained anatomical structures, enabling the generation of smooth and spatially consistent deformation fields. In contrast, feature-based approaches, which operate at a more abstract level, often compromise spatial fidelity. Although regularization techniques, as implemented in FeatAlignReg, enhance the plausibility of deformation fields, they also reduce risk prediction performance. This highlights a trade-off between the quality of the deformation field and the effectiveness of risk prediction.

A notable contribution of this work is the introduction of a hybrid strategy, ImgFeatAlign, which applies image-derived deformation fields within the feature space. This approach leverages the anatomical consistency of image-based registration while benefiting from the representational efficiency of feature-level modeling. By combining spatial precision and feature-level abstraction, the approach achieves robust and reliable performance across different patient populations and imaging conditions.

From a translational perspective, the ability to maintain accurate predictions at multiple time points is vital to identifying high-risk patients. This is particularly critical in challenging imaging conditions, where spatial misalignment can compromise model reliability. ImgFeatAlign exhibits strong generalization across datasets and breast density categories, supporting its potential for real-world clinical deployment.

In summary, these findings highlight that spatial alignment is not merely a preprocessing step, but a foundational component in building robust, trustworthy models for longitudinal risk prediction.

\section{Limitations and Outlook}\label{limitationsoutlook}

While this study demonstrates the effectiveness of image-based alignment strategies for longitudinal breast cancer risk prediction, several promising directions for future work remain beyond the scope of the current study. First, the proposed longitudinal risk prediction models are restricted to using only a single prior mammogram for each patient. In clinical practice, however, radiologists often review multiple historical examinations to assess temporal changes more comprehensively. Leveraging multiple prior images could enable richer temporal modeling and potentially improve risk estimation by capturing longer-term progression patterns.

Second, the proposed framework performs single-view risk prediction, relying on either the CC or MLO view independently. In contrast, clinical decision-making typically involves the joint assessment of both views, as they provide complementary anatomical information. The inability to integrate multiview data may limit the model’s capacity to fully characterize breast tissue and subtle pathological changes.

Additional limitations include the reliance on image registration quality, as inaccuracies in deformation estimation may propagate into the aligned feature representations. \myhl{Furthermore, this work focuses exclusively on mammography and does not consider digital breast tomosynthesis (DBT). While mammography remains the primary imaging modality in many breast cancer screening programs, DBT provides complementary three-dimensional information that may further improve longitudinal risk prediction.}

Looking ahead, future work will focus on developing multiview longitudinal risk prediction models that jointly analyze CC and MLO views and can incorporate multiple prior mammograms per patient, \myhl{and investigate the integration of DBT into longitudinal risk prediction frameworks}. Such models would more closely reflect clinical workflows and enable more robust temporal reasoning. Extending the framework to support longer imaging histories, improving alignment robustness, and validating performance across diverse populations and imaging platforms are additional important next steps toward clinically deployable longitudinal risk assessment systems.

\section{Conclusion}\label{conclusion}

In conclusion, this study highlights the importance of accurate spatial alignment for longitudinal breast cancer risk prediction. Among the evaluated approaches, image-based alignment methods, particularly ImgFeatAlign, consistently outperform alternative strategies by preserving anatomical correspondence and enabling temporally coherent feature modeling. By integrating deformation fields from image-based registration into the feature space, ImgFeatAlign effectively balances spatial precision with high-level semantic representation, leading to improved predictive accuracy. \myhl{Our findings demonstrate that the value of longitudinal mammograms depends not only on their availability, but also on accurate spatial alignment. Unaligned prior examinations may introduce noise that outweighs the potential benefits of additional temporal information.}

These results underscore the clinical potential of robust longitudinal modeling for personalized breast cancer risk assessment. Accurate alignment of temporal imaging data can enhance the detection of subtle changes over time, supporting earlier identification of individuals at elevated risk and enabling more informed screening and prevention strategies. Overall, this work provides a foundation for future advances in longitudinal, imaging-based prognostic models and contributes to the broader goal of precision medicine in breast cancer care.

\section*{CRediT authorship contribution statement}
\textbf{Solveig Thrun:} Conceptualization, Software, Methodology, Formal analysis, Investigation, Visualization, Writing - original draft, Writing - review \& editing. \textbf{Stine Hansen:} Conceptualization, Methodology, Writing - review \& editing, Supervision.  \textbf{ Zijun Sun:} Methodology, Writing - review \& editing. \textbf{Nele Blum:} Methodology, Writing - review \& editing, Supervision. \textbf{Suaiba A. Salahuddin:} Methodology, Writing - review \& editing. \textbf{Xin Wang:} Methodology, Writing - review \& editing. \textbf{Kristoffer Wickstrøm:}  Conceptualization, Methodology, Writing - review \& editing, Supervision.  \textbf{Elisabeth Wetzer:}  Conceptualization, Methodology, Writing - review \& editing, Supervision.  \textbf{Robert Jenssen:}  Conceptualization, Methodology, Writing - review \& editing, Supervision, Funding acquisition.   \textbf{Maik Stille:}  Methodology, Writing - review \& editing, Supervision. \textbf{Michael Kampffmeyer:} Conceptualization, Methodology, Writing - review \& editing, Supervision,  Project administration, Funding acquisition.

\section*{Declaration of Competing Interest}
The authors declare that they have no known competing financial interests or personal relationships that could have appeared to influence the work reported in this paper.

 \section*{Data availability}
The source code used in this study is publicly available on GitHub, with the link provided in the abstract. All datasets used in this work are publicly accessible and links to request access are provided in the Experimental Setup section.
 
\section*{Acknowledgements}
This work was supported by The Research Council of Norway (Visual Intelligence, grant no. 309439 as well as FRIPRO grant no. 315029 and 360068 and IKTPLUSS grant no. 303514). We acknowledge Sigma2 – the National Infrastructure for High Performance Computing and Data Storage in Norway, for awarding this project access to the LUMI supercomputer, owned by the EuroHPC Joint Undertaking, hosted by CSC (Finland) and the LUMI consortium through Sigma2 [grant no. NN8106K].

\appendix

\section{\texorpdfstring{\myhl{Ablation Study of the Temporal Self-Attention Mechanism}}{Ablation Study of the Temporal Self-Attention Mechanism}}\label{ablationstud}

\myhl{To disentangle the effect of the multi-head self-attention layer within the temporal self-attention mechanism, we conducted two ablation studies. Both variants retain the same overall network architecture and training protocol, differing only in the temporal self-attention mechanism.

In the first ablation, the multi-head self-attention layer was removed while the remaining feedforward block was retained. This variant evaluates whether explicit temporal attention is necessary or whether comparable performance can be achieved using a feedforward-only architecture (see  Table}~\ref{tab:riskpredFFN1}).

\myhl{In the second ablation, the multi-head self-attention layer was replaced with a feedforward network with a hidden dimension of 1024, resulting in approximately the same number of trainable parameters as the original multi-head self-attention layer (approximately 1.05 million parameters) (see Table}~\ref{tab:riskpredffn3}).\myhl{ This comparison controls for model capacity and assesses whether any performance gains can be attributed to the attention mechanism itself rather than to an increased number of parameters.

Together, these ablation studies isolate the contribution of the multi-head self-attention mechanism from that of model capacity, demonstrating the benefit of the mechanism in the temporal self-attention. }

\begin{table}[thbp]
\centering
\scriptsize
\caption{1–5 year breast cancer risk prediction using different alignment methods. C-index and AUC values with 95\% confidence intervals.  In this ablation, the multi-head self-attention layer and the first Add + Norm layer were removed from the temporal self-attention mechanism (orange box in Figure~\ref{fig:alignment_models}). The remaining module consists of a Feedforward (with a hidden dimension of 2048) $\rightarrow$ Add + Norm block. }
\label{tab:riskpredFFN1}
\begin{adjustbox}{center}
\myblue{
\begin{tabular}{llc*{5}{c}}
\toprule
\textbf{Dataset} & \textbf{Method} & \textbf{C-index (\%) $\uparrow$} & \multicolumn{5}{c}{\textbf{Follow-up year AUC (\%) $\uparrow$}} \\
\cmidrule(lr){4-8}
& & & \textbf{1-yr} & \textbf{2-yr} & \textbf{3-yr} & \textbf{4-yr} & \textbf{5-yr} \\
\midrule
\multirow{7}{*}{\textbf{EMBED}}
& SingleTimePoint
& \makecell{62.2 \\ (59.9-64.4)}
& \makecell{62.6 \\ (59.9-65.2)}
& \makecell{62.8 \\ (60.4-65.3)}
& \makecell{61.1 \\ (58.7-63.4)}
& \makecell{60.1 \\ (57.8-62.4)}
& \makecell{60.9 \\ (57.1-64.7)} \\

& NoAlign
& \makecell{60.3 \\ (58.1-62.5)}
& \makecell{61.3 \\ (58.5-64.2)}
& \makecell{60.3 \\ (57.8-62.7)}
& \makecell{59.3 \\ (56.7-61.6)}
& \makecell{58.7 \\ (56.3-61.2)}
& \makecell{58.8 \\ (54.3-63.5)} \\

& ImplicitAlign
& \makecell{66.6 \\ (63.8-69.4)}
& \makecell{66.4 \\ (62.9-69.7)}
& \makecell{67.2 \\ (64.2-70.2)}
& \makecell{67.9 \\ (65.1-70.7)}
& \makecell{67.9 \\ (65.1-70.6)}
& \makecell{67.1 \\ (63.4-70.7)} \\

& FeatAlign
& \makecell{68.1 \\ (65.4-72.7)}
& \makecell{68.2 \\ (65.1-71.5)}
& \makecell{68.4 \\ (65.5-71.1)}
& \makecell{69.2 \\ (66.5-71.7)}
& \makecell{69.1 \\ (66.5-71.7)}
& \makecell{68.5 \\ (64.6-72.0)} \\

& FeatAlignReg
& \makecell{67.3 \\ (64.7-69.9)}
& \makecell{67.2 \\ (64.4-70.2)}
& \makecell{67.2 \\ (64.5-69.9)}
& \makecell{68.4 \\ (65.9-71.0)}
& \makecell{68.3 \\ (65.8-70.9)}
& \makecell{67.7 \\ (64.2-71.5)} \\

& ImgAlign
& \makecell{68.4 \\ (65.7-71.0)}
& \makecell{68.6 \\ (65.3-71.8)}
& \makecell{69.1 \\ (66.1-71.8)}
& \makecell{69.5 \\ (66.8-72.1)}
& \makecell{69.3 \\ (66.8-71.9)}
& \makecell{69.4 \\ (65.9-72.9)} \\

& \textbf{ImgFeatAlign}
& \makecell{\textbf{69.7} \\ \textbf{(67.4-71.7)}}
& \makecell{\textbf{69.8} \\ \textbf{(67.0-72.4)}}
& \makecell{\textbf{70.1} \\ \textbf{(67.6-72.5)}}
& \makecell{\textbf{71.2} \\ \textbf{(69.0-73.4)}}
& \makecell{\textbf{71.5} \\ \textbf{(69.3-73.6)}}
& \makecell{\textbf{70.3} \\ \textbf{(66.5-74.1)}} \\

\midrule
\multirow{7}{*}{\textbf{CSAW-CC}}
& SingleTimePoint
& \makecell{61.8 \\ (59.7-63.7)}
& \makecell{62.5 \\ (59.7-65.1)}
& \makecell{62.4 \\ (60.2-64.4)}
& \makecell{62.1 \\ (59.9-64.1)}
& \makecell{61.5 \\ (59.4-63.6)}
& \makecell{62.9 \\ (60.4-65.5)} \\

& NoAlign
& \makecell{60.5 \\ (58.5-62.6)}
& \makecell{61.1 \\ (58.3-63.8)}
& \makecell{61.0 \\ (58.8-63.0)}
& \makecell{61.1 \\ (59.0-63.2)}
& \makecell{60.0 \\ (57.8-62.2)}
& \makecell{61.4 \\ (58.8-64.0)} \\

& ImplicitAlign
& \makecell{62.8 \\ (61.0-64.7)}
& \makecell{64.4 \\ (61.9-66.9)}
& \makecell{63.1 \\ (61.1-65.1)}
& \makecell{63.0 \\ (61.0-64.9)}
& \makecell{62.5 \\ (60.5-64.6)}
& \makecell{66.3 \\ (64.0-68.6)} \\

& FeatAlign
& \makecell{64.8 \\ (62.8-65.9)}
& \makecell{65.4 \\ (62.9-67.7)}
& \makecell{64.9 \\ (62.8-66.9)}
& \makecell{64.2 \\ (62.2-66.2)}
& \makecell{65.1 \\ (63.0-67.2)}
& \makecell{65.7 \\ (63.5-68.1)} \\

& FeatAlignReg
& \makecell{62.7 \\ (61.7-65.6)}
& \makecell{64.9 \\ (62.5-67.1)}
& \makecell{64.6 \\ (62.6-66.7)}
& \makecell{62.2 \\ (62.3-62.2)}
& \makecell{64.2 \\ (62.3-66.4)}
& \makecell{67.2 \\ (64.8-69.6)} \\

& ImgAlign
& \makecell{65.7 \\ (63.8-67.5)}
& \makecell{66.2 \\ (63.5-68.8)}
& \makecell{65.1 \\ (63.0-67.2)}
& \makecell{65.7 \\ (63.7-67.7)}
& \makecell{66.1 \\ (64.2-68.2)}
& \makecell{67.6 \\ (65.3-69.9)} \\

& \textbf{ImgFeatAlign}
& \makecell{\textbf{66.3} \\ \textbf{(64.4-68.3)}}
& \makecell{\textbf{68.2} \\ \textbf{(65.6-70.9)}}
& \makecell{\textbf{66.9} \\ \textbf{(64.9-69.0)}}
& \makecell{\textbf{68.1} \\ \textbf{(66.0-70.1)}}
& \makecell{\textbf{67.5} \\ \textbf{(65.4-69.6)}}
& \makecell{\textbf{68.3} \\ \textbf{(65.8-70.9)}} \\

\bottomrule
\end{tabular}
}
\end{adjustbox}
\end{table}

\begin{table}[thbp]
\centering
\scriptsize
\caption{1–5 year breast cancer risk prediction using different alignment methods. C-index and AUC values with 95\% confidence intervals. In this ablation, the multi-head self-attention layer in the temporal fusion module (orange box in Figure~\ref{fig:alignment_models}) was replaced with a feedforward layer (with a hidden dimension of 1024), resulting in approximately the same number of trainable parameters as the original self-attention layer. The modified temporal self-attention mechanism consists of Feedforward (hidden dimension = 1024) $\rightarrow$ Add + Norm $\rightarrow$ Feedforward (hidden dimension = 2048) $\rightarrow$ Add + Norm. }
\label{tab:riskpredffn3}
\begin{adjustbox}{center}
\myblue{
\begin{tabular}{llc*{5}{c}}
\toprule
\textbf{Dataset} & \textbf{Method} & \textbf{C-index (\%) $\uparrow$} & \multicolumn{5}{c}{\textbf{Follow-up year AUC (\%) $\uparrow$}} \\
\cmidrule(lr){4-8}
& & & \textbf{1-yr} & \textbf{2-yr} & \textbf{3-yr} & \textbf{4-yr} & \textbf{5-yr} \\
\midrule
% EMBED
\multirow{7}{*}{\textbf{EMBED}}
& SingleTimePoint
& \makecell{59.2 \\ (56.6-61.7)}
& \makecell{58.8 \\ (56.0-61.5)}
& \makecell{58.3 \\ (55.9-60.8)}
& \makecell{59.4 \\ (57.0-61.8)}
& \makecell{61.5 \\ (59.0-64.0)}
& \makecell{58.9 \\ (54.6-63.2)} \\

& NoAlign
& \makecell{58.2 \\ (55.8-60.8)}
& \makecell{58.6 \\ (55.7-61.6)}
& \makecell{57.9 \\ (55.4-60.3)}
& \makecell{57.4 \\ (54.9-59.8)}
& \makecell{57.4 \\ (54.9-59.9)}
& \makecell{52.6 \\ (48.1-57.1)} \\

& ImplicitAlign
& \makecell{68.8 \\ (66.5-70.9)}
& \makecell{69.7 \\ (67.0-72.4)}
& \makecell{68.9 \\ (66.4-71.4)}
& \makecell{69.6 \\ (67.4-71.9)}
& \makecell{70.4 \\ (68.0-72.8)}
& \makecell{68.7 \\ (64.5-72.9)} \\

& FeatAlign
& \makecell{69.8 \\ (67.3-72.4)}
& \makecell{70.2 \\ (67.2-73.4)}
& \makecell{69.9 \\ (67.1-72.6)}
& \makecell{71.1 \\ (68.6-73.7)}
& \makecell{70.6 \\ (68.1-73.1)}
& \makecell{70.6 \\ (67.1-74.0)} \\

& FeatAlignReg
& \makecell{69.2 \\ (66.7-71.9)}
& \makecell{69.4 \\ (66.4-72.5)}
& \makecell{69.5 \\ (66.8-72.1)}
& \makecell{70.1 \\ (67.6-72.7)}
& \makecell{69.5 \\ (67.0-72.1)}
& \makecell{69.1 \\ (65.5-72.6)} \\

& ImgAlign
& \makecell{70.6 \\ (68.2-73.2)}
& \makecell{70.3 \\ (67.2-73.4)}
& \makecell{70.5 \\ (67.9-73.1)}
& \makecell{70.8 \\ (68.3-73.4)}
& \makecell{70.2 \\ (67.9-72.6)}
& \makecell{71.1 \\ (67.6-74.6)} \\

& \textbf{ImgFeatAlign}
& \makecell{\textbf{71.9} \\ \textbf{(69.5-74.4)}}
& \makecell{\textbf{71.7} \\ \textbf{(68.7-74.8)}}
& \makecell{\textbf{71.2} \\ \textbf{(68.6-73.8)}}
& \makecell{\textbf{71.2} \\ \textbf{(68.8-73.6)}}
& \makecell{\textbf{71.5} \\ \textbf{(69.1-73.9)}}
& \makecell{\textbf{71.7} \\ \textbf{(68.3-75.2)}} \\

% CSAW-CC
\midrule
\multirow{7}{*}{\textbf{CSAW-CC}}
& SingleTimePoint
& \makecell{60.7 \\ (58.6-62.7)}
& \makecell{61.2 \\ (58.4-63.9)}
& \makecell{61.5 \\ (59.4-63.5)}
& \makecell{61.8 \\ (59.7-63.8)}
& \makecell{60.8 \\ (58.7-62.9)}
& \makecell{60.2 \\ (57.7-62.5)} \\

& NoAlign
& \makecell{59.1 \\ (57.2-61.1)}
& \makecell{59.5 \\ (56.9-62.2)}
& \makecell{59.5 \\ (57.4-61.5)}
& \makecell{59.1 \\ (57.0-61.1)}
& \makecell{58.1 \\ (56.0-60.1)}
& \makecell{59.0 \\ (56.4-61.5)} \\

& ImplicitAlign
& \makecell{63.4 \\ (61.6-65.3)}
& \makecell{65.8 \\ (63.1-68.4)}
& \makecell{64.7 \\ (62.6-66.8)}
& \makecell{66.3 \\ (64.2-68.3)}
& \makecell{64.8 \\ (62.7-66.9)}
& \makecell{67.9 \\ (65.4-70.2)} \\

& FeatAlign
& \makecell{65.3 \\ (63.4-67.1)}
& \makecell{67.1 \\ (64.4-69.6)}
& \makecell{66.6 \\ (64.5-68.5)}
& \makecell{66.0 \\ (64.1-67.9)}
& \makecell{65.1 \\ (63.1-67.1)}
& \makecell{67.4 \\ (65.0-69.6)} \\

& FeatAlignReg
& \makecell{64.2 \\ (62.3-66.1)}
& \makecell{66.8 \\ (64.2-69.2)}
& \makecell{65.4 \\ (63.2-67.4)}
& \makecell{65.1 \\ (63.0-67.0)}
& \makecell{64.8 \\ (62.5-66.8)}
& \makecell{66.4 \\ (63.9-68.8)} \\

& ImgAlign
& \makecell{66.6 \\ (64.7-68.4)}
& \makecell{67.2 \\ (64.5-69.8)}
& \makecell{67.1 \\ (65.1-69.1)}
& \makecell{66.6 \\ (64.4-68.6)}
& \makecell{66.4 \\ (64.2-68.4)}
& \makecell{67.1 \\ (64.7-69.4)} \\

& \textbf{ImgFeatAlign}
& \makecell{\textbf{67.5} \\ \textbf{(65.6-69.4)}}
& \makecell{\textbf{67.6} \\ \textbf{(65.1-70.1)}}
& \makecell{\textbf{67.8} \\ \textbf{(65.7-69.8)}}
& \makecell{\textbf{66.9} \\ \textbf{(64.9-68.8)}}
& \makecell{\textbf{66.8} \\ \textbf{(64.7-68.7)}}
& \makecell{\textbf{67.4} \\ \textbf{(65.1-69.7)}} \\ 
\bottomrule
\end{tabular}
}
\end{adjustbox}
\end{table}

\section{Dataset Details, Label Generation, and Pre-processing}

\subsection{Dataset Descriptions}\label{app:datasets}

\textbf{EMory BrEast Imaging Dataset (EMBED).}
The EMory BrEast Imaging Dataset (EMBED)~\citep{embed} is a large, racially diverse mammography dataset comprising screening and diagnostic exams collected between 2013 and 2020 using Hologic, GE, and Fujifilm systems. The publicly released subset used in this study includes 364,564 mammograms from 23,256 patients, corresponding to approximately 20\% of the full 2D and synthesized C-view dataset. In this study, we included only standard 2D digital mammograms from EMBED. Synthesized C-view images were not used. In addition to mammography images, EMBED provides clinical metadata, including BI-RADS tissue density categories~\citep{birads}. Both invasive and screen-detected non-invasive (in situ) breast cancer cases are included in the EMBED dataset.

\textbf{Cohort of Screen-Aged Women Case Control (CSAW-CC).}
The Cohort of Screen-Aged Women Case Control (CSAW-CC) dataset~\citep{csawcc} consists of mammography exams acquired at Karolinska University Hospital in Stockholm, Sweden, between November 2008 and December 2015. All images were collected using Hologic systems. The dataset includes 873 first-time breast cancer cases among women aged 40--74 years, along with 7,850 randomly selected healthy controls from the same screening period. Clinical metadata are provided, including breast percent density estimates generated using the LIBRA software~\citep{libra}. The CSAW-CC dataset also includes both invasive and non-invasive breast cancer cases.

\textbf{Follow-up and inclusion criteria.}
For both datasets, negative cases correspond to women who were not diagnosed with breast cancer during the respective study observation periods (8 years for EMBED and 7 years for CSAW-CC).
Time-to-last-follow-up for negative cases is defined as the time from the index screening mammogram to the last available screening exam within the study period, during which no cancer diagnosis was recorded.
We included only patients with at least five years of follow-up data in both datasets following~\citep{Wan_Ordinal_MICCAI2024}. All included patients therefore have at least two timepoints. For patients with more than two timepoints, we generate all consecutive pairs for training. For example, a patient with four timepoints \((t_1, t_2, t_3, t_4)\) contributes three consecutive pairs: \((t_1, t_2)\), \((t_2, t_3)\), and \((t_3, t_4)\). Most patients in both datasets have screenings approximately 1–2 years apart, although some intervals can be longer, up to 3–4 years.

\subsection{Label Generation}\label{app:labelgen}

The Time-to-Cancer label represents the temporal interval between the date of a screening mammogram and the corresponding breast cancer diagnosis.

\textbf{EMBED labeling.}
Following the approach in OA-BreaCR~\citep{Wan_Ordinal_MICCAI2024}, positive cases in the EMBED dataset were identified based on the BI-RADS score and the pathological severity score. Specifically, positive cases included mammograms with a BI-RADS score of 6 (indicating known biopsy-proven cancer) or a pathological severity score of 0 or 1 (corresponding to invasive and non-invasive cancers). Negative cases were selected from mammograms with BI-RADS scores of 1 or 2 (indicating negative or benign findings), as well as images of BI-RADS 0 that were later reclassified as BI-RADS 1 or 2 in subsequent diagnostic studies. For EMBED, the cancer diagnosis date reported in this study corresponds to the date of the pathology result (i.e., biopsy confirmation). 

\textbf{CSAW-CC labeling.}
Label generation for CSAW-CC followed the approach in LoMaR~\citep{Kar_Longitudinal_MICCAI2024}, positive cases in the CSAW-CC dataset were identified based on radiological timing (rad\_timing), where images with a rad\_timing of 1 correspond to screen-detected cancers, and those with a rad\_timing of 2 correspond to interval cancers. 
The negative group consisted of images without any recorded rad\_timing, corresponding to women with no cancer diagnosis during the study period, as indicated by a dedicated variable in the dataset metadata.
Biopsy dates are not available in the CSAW-CC dataset. Therefore, the rad\_timing variable provided by the dataset was used.

To determine the cancer year for each patient, the dataset was grouped by patient ID, and the record with the latest exam year was selected. The cancer year was then assigned based on the rad\_timing of the cancer diagnosis for this exam:
\begin{itemize}
    \item For screen-detected cancers (rad\_timing = 1), the cancer year was set to the year of the mammogram exam.
    \item For interval cancers (rad\_timing = 2), the cancer year was set to one year after the exam year. However, if the most recent exam year was 2016, the cancer year was adjusted to 2016 to accommodate specific follow-up protocols, as 2016 marked the final year of the CSAW-CC study.
\end{itemize} 

\textbf{Time-to-Cancer computation.}
For both datasets, the Time-to-Cancer label was computed as the difference between the assigned cancer year and the mammogram exam year.  
For both datasets, only negative prior screening mammograms were used in the experiments, as routine screening follow-up ceases once a patient is diagnosed with breast cancer.

\subsection{Pre-processing and Dataset Splits}\label{app:preproc}

Preprocessing follows the procedure described in the main paper, including DICOM-to-PNG conversion, background removal, breast segmentation, and resizing. Table~\ref{tab:timecancerdataset} presents the distribution of the Time-to-Cancer label and the dataset split for the EMBED and CSAW-CC datasets. While follow-up criteria and the train/validation/test split ratio are described in the main paper, these tables provide the detailed counts for both datasets.

Breast density information differs between the two datasets. EMBED provides BI-RADS density categories, whereas CSAW-CC dataset provides breast percent density estimates generated using the LIBRA software. These estimates are categorized into three groups with equal bin widths, corresponding to low (0–33\%), medium (33–66\%), and high (66–100\%) density levels. Table~\ref{tab:combined_densitydistr} shows the density distribution for both the EMBED and CSAW-CC datasets.

 \begin{table}[t]
\centering
\scriptsize
\caption{Time-to-cancer label distribution and dataset splits. Percentages are reported relative to the total number of cases per split.}
\label{tab:timecancerdataset}
\begin{adjustbox}{center}
\begin{tabular}{llccccccccc}
\toprule
& & \multicolumn{6}{c}{\textbf{Positive}} & \textbf{Negative} & \\
\cmidrule(lr){3-8} \cmidrule(lr){9-9}
\textbf{Dataset} & \textbf{Split} & \textbf{1Y} & \textbf{2Y} & \textbf{3Y} & \textbf{4Y} & \textbf{5Y} & \textbf{All} & \textbf{$>$5Y} & \textbf{Total} \\
\midrule

\multirow{4}{*}{\textbf{EMBED}} 
& Train &
\makecell{539\\(2.1\%)} &
\makecell{190\\(0.7\%)} &
\makecell{163\\(0.6\%)} &
\makecell{90\\(0.3\%)} &
\makecell{67\\(0.3\%)} &
\makecell{1049\\(4.0\%)} &
\makecell{25076\\(96.0\%)} &
\makecell{26125}\\

& Val &
\makecell{216\\(2.1\%)} &
\makecell{60\\(0.6\%)} &
\makecell{57\\(0.5\%)} &
\makecell{39\\(0.4\%)} &
\makecell{26\\(0.3\%)}&
\makecell{398\\(3.8\%)} &
\makecell{9998\\(96.2\%)} &
\makecell{ 10396}\\

& Test &
\makecell{310\\(2.0\%)} &
\makecell{141\\(0.9\%)} &
\makecell{97\\(0.6\%)} &
\makecell{62\\(0.4\%)} &
\makecell{46\\(0.3\%)} &
\makecell{656\\(4.2\%)} &
\makecell{15021\\(95.8\%)} &
\makecell{15677 }\\

& \textbf{Total} &
\makecell{\textbf{1065}\\\textbf{(2.0\%)}} &
\makecell{\textbf{391}\\\textbf{(0.7\%)}} &
\makecell{\textbf{317}\\\textbf{(0.6\%)}} &
\makecell{\textbf{191}\\\textbf{(0.4\%)}} &
\makecell{\textbf{139}\\\textbf{(0.3\%)}} &
\makecell{\textbf{2103}\\\textbf{(4.0\%)}} &
\makecell{\textbf{50095}\\\textbf{(96.0\%)}} &
\makecell{\textbf{52198}} \\

\midrule

\multirow{4}{*}{\textbf{CSAW-CC}} 
& Train &
\makecell{960\\(2.9\%)} &
\makecell{480\\(1.4\%)} &
\makecell{336\\(1.0\%)} &
\makecell{324\\(1.0\%)} &
\makecell{184\\(0.5\%)} &
\makecell{2284\\(6.8\%)} &
\makecell{31376\\(93.2\%)} &
\makecell{33660 }\\

& Val &
\makecell{396\\(3.0\%)} &
\makecell{188\\(1.4\%)} &
\makecell{160\\(1.2\%)}&
\makecell{132\\(1.0\%)} &
\makecell{72\\(0.5\%)} &
\makecell{948\\(7.1\%)} &
\makecell{12376\\(92.9\%)} &
\makecell{13324 }\\

& Test &
\makecell{568\\(2.8\%)} &
\makecell{312\\(1.5\%)} &
\makecell{200\\(1.0\%)}&
\makecell{204\\(1.0\%)} &
\makecell{112\\(0.6\%)} &
\makecell{1396\\(6.9\%) }&
\makecell{18804\\(93.1\%)} &
\makecell{20200 } \\

& \textbf{Total} &
\makecell{\textbf{1924}\\\textbf{(2.9\%)}} &
\makecell{\textbf{980}\\\textbf{(1.5\%)}} &
\makecell{\textbf{696}\\\textbf{(1.0\%)}} &
\makecell{\textbf{660}\\\textbf{(1.0\%)}} &
\makecell{\textbf{368}\\\textbf{(0.5\%)}} &
\makecell{\textbf{4628}\\\textbf{(6.9\%)}} &
\makecell{\textbf{62556}\\\textbf{(93.1\%)}} &
\makecell{ \textbf{67184} } \\

\bottomrule
\end{tabular}
\end{adjustbox}
\end{table}

\begin{table}[t]
\centering
\scriptsize
\caption{Density distribution across the EMBED and CSAW-CC datasets. Percentages are reported relative to the total number of cases per density category.}
\label{tab:combined_densitydistr}
\begin{adjustbox}{center}
\begin{tabular}{lccccccc}
\toprule
& \multicolumn{4}{c}{\textbf{EMBED (BI-RADS)}} & \multicolumn{3}{c}{\textbf{CSAW-CC}} \\
\cmidrule(lr){2-5} \cmidrule(lr){6-8}
\textbf{Split} & \textbf{A} & \textbf{B} & \textbf{C} & \textbf{D} & \textbf{Low} & \textbf{Medium} & \textbf{High} \\
\midrule
Train &
\makecell{2822\\(47.6\%)} &
\makecell{11225\\(50.9\%)} &
\makecell{10608\\(49.7\%)} &
\makecell{1110\\(49.2\%)} &
\makecell{24932\\(50.0\%)} &
\makecell{8404\\(49.9\%)} &
\makecell{324\\(61.7\%)} \\

Val &
\makecell{1216\\(20.5\%)} &
\makecell{4288\\(19.4\%)} &
\makecell{4256\\(19.9\%)} &
\makecell{540\\(23.9\%)} &
\makecell{9755\\(19.6\%)} &
\makecell{3460\\(20.6\%)} &
\makecell{109\\(20.8\%)} \\

Test &
\makecell{1885\\(31.8\%)} &
\makecell{6547\\(29.7\%)} &
\makecell{6492\\(30.4\%)} &
\makecell{608\\(26.9\%)} &
\makecell{15145\\(30.4\%)} &
\makecell{4963\\(29.5\%)} &
\makecell{92\\(17.5\%)} \\

\midrule
\textbf{Total} &
\makecell{\textbf{5923}} &
\makecell{\textbf{22060}} &
\makecell{\textbf{21356}} &
\makecell{\textbf{2258}} &
\makecell{\textbf{49832}} &
\makecell{\textbf{16827}} &
\makecell{\textbf{525}} \\
\bottomrule
\end{tabular}
\end{adjustbox}
\end{table}

\textbf{Registration model training data.}
For training the registration models (MammoRegNet), we randomly selected 1,000 patients who were excluded from the risk prediction dataset in order to prevent data leakage and ensure no patient overlap between the registration and risk prediction datasets, using two images per laterality and view combination. This selection method helps minimize training time and accommodate memory limitations. The dataset is then randomly split at the patient level into training, validation, and test sets, following the same 5:2:3 ratio.

\section{Schematic Illustrations of Longitudinal Risk Prediction Architectures}
\begin{figure}[htbp]
    \centering
 \begin{subfigure}[b]{0.47\textwidth}
        \centering
        \includegraphics[width=1.0\textwidth]{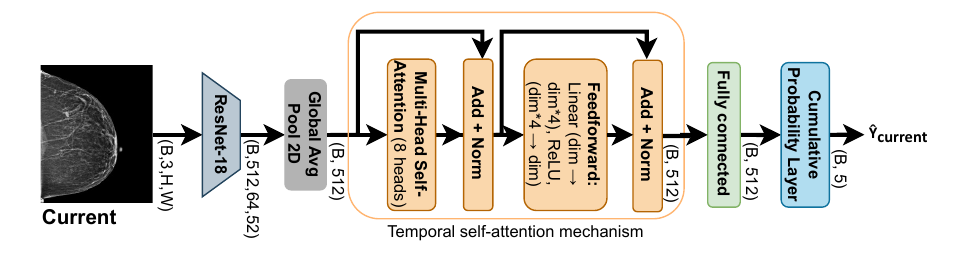}
        \caption{SingleTimePoint}
        \label{fig:Asingletimepoint}
    \end{subfigure}  \quad
    \begin{subfigure}[b]{0.47\textwidth}
        \centering
        \includegraphics[width=1.0\textwidth]{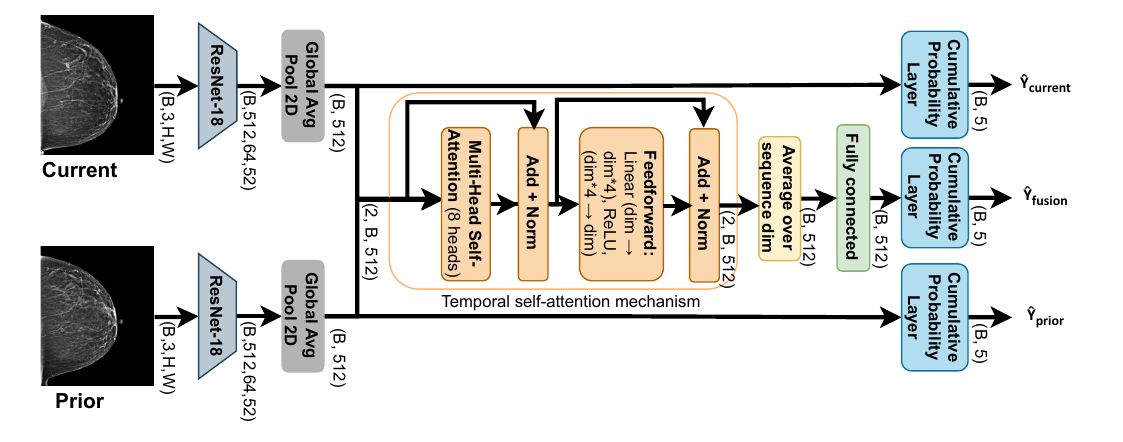}
        \caption{NoAlign}
        \label{fig:Anoalign}
    \end{subfigure}
    \\
    \begin{subfigure}[b]{\textwidth}
        \centering
        \includegraphics[width=0.7\textwidth]{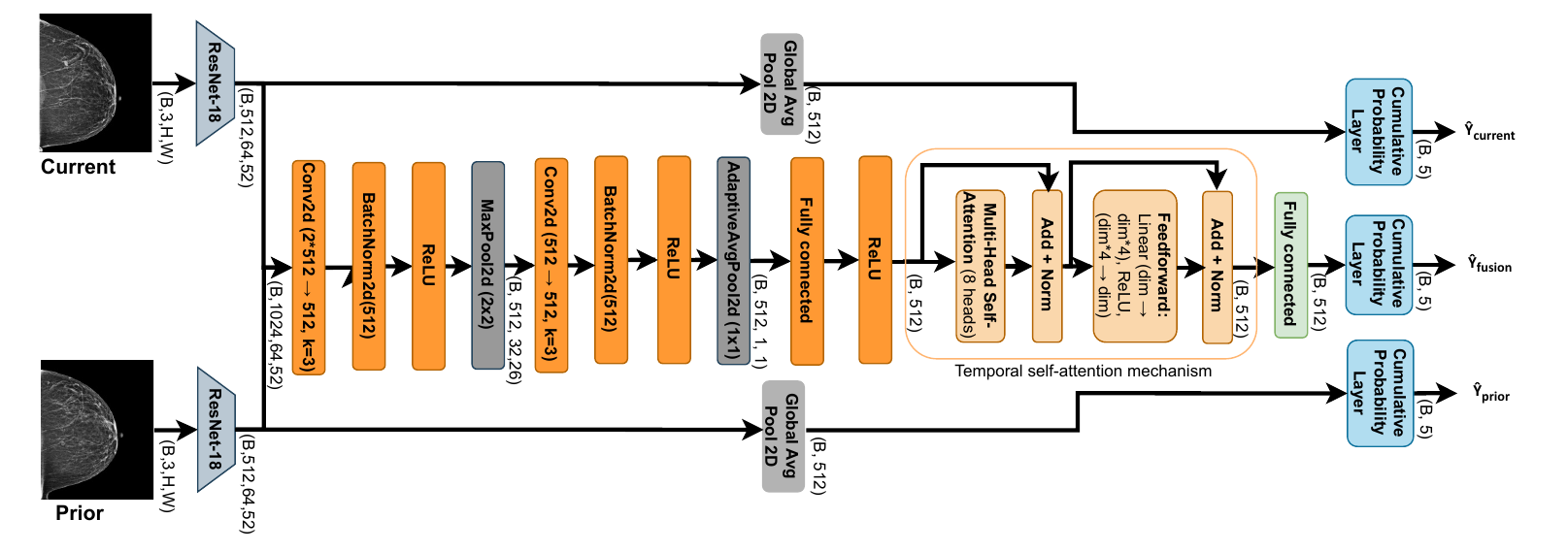}
        \caption{ImplicitAlign}
        \label{fig:Aimplicit}
    \end{subfigure}
\\
    \begin{subfigure}[b]{\textwidth}
        \centering
        \includegraphics[width=0.7\textwidth]{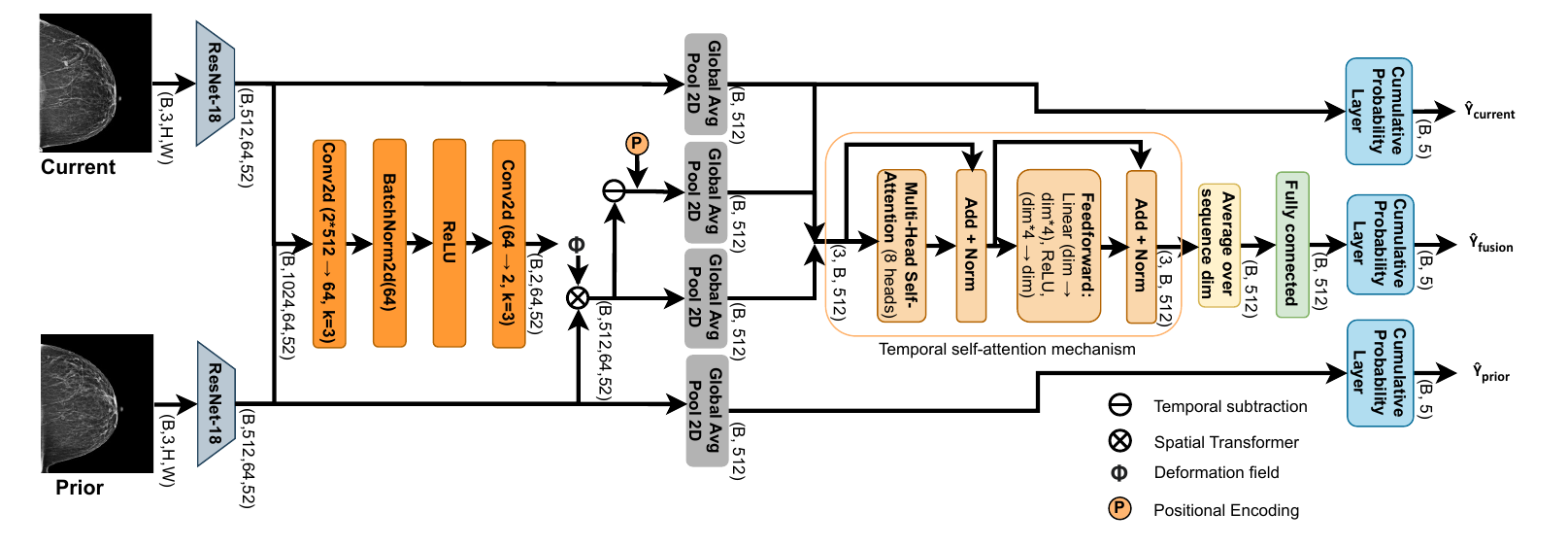}
        \caption{FeatAlign, FeatAlignReg}
        \label{fig:Afeatalign}
    \end{subfigure}
\\        \begin{subfigure}[b]{\textwidth}
        \centering
        \includegraphics[width=0.7\textwidth]{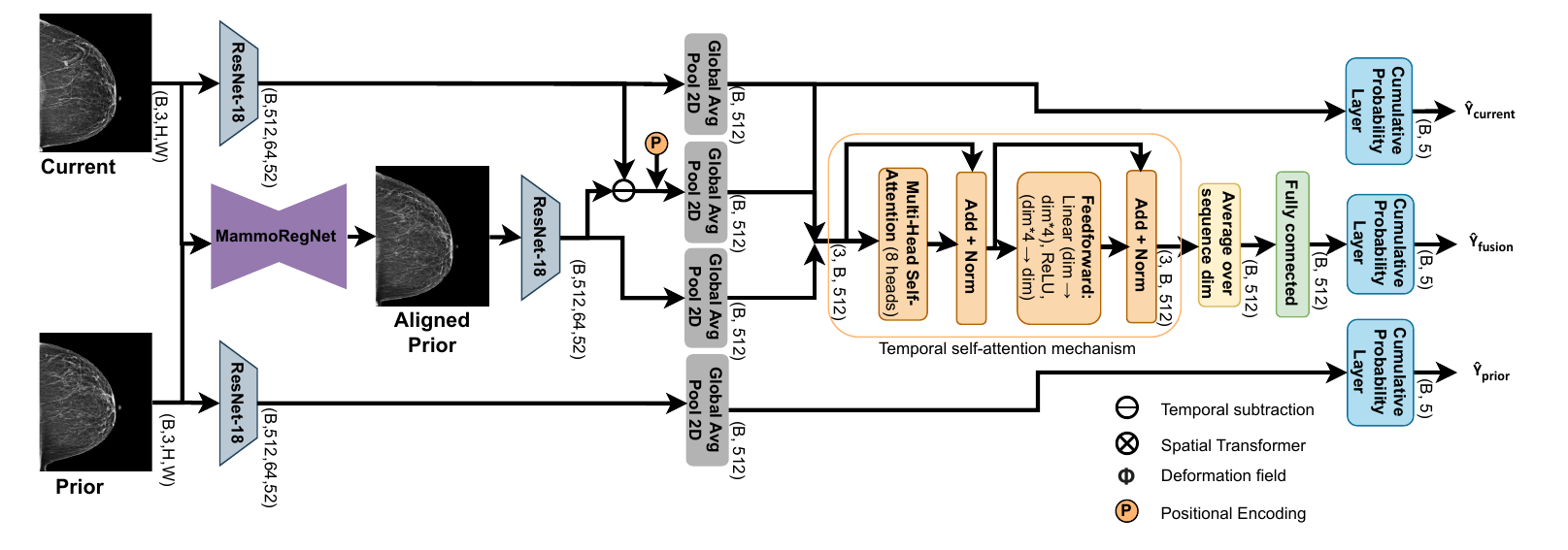}
        \caption{ImgAlign}
        \label{fig:Aimgalign}
    \end{subfigure}

    \vspace{1em}

    \begin{subfigure}[b]{\textwidth}
        \centering
        \includegraphics[width=0.65\textwidth]{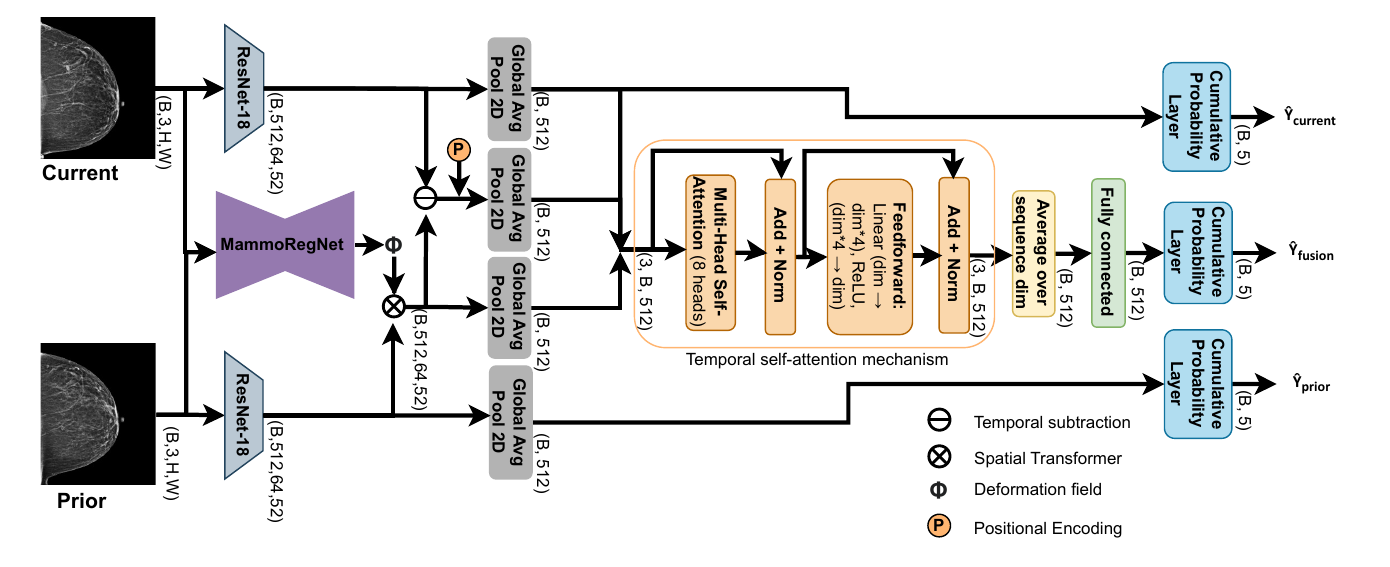}
        \caption{ImgFeatAlign}
        \label{fig:Aimgfeatalign}
    \end{subfigure}

    \caption{Visualization of the architectures for the longitudinal risk prediction. }
    \label{fig:alignment_models}
\end{figure}

Figure~\ref{fig:alignment_models} provides a detailed visualization of the various alignment strategies evaluated within the unified risk prediction framework. Each subfigure illustrates a distinct architectural design, including sinlgetimepoint, no alignment, implicit alignment, feature-level alignment (FeatAlign, FeatAlignReg), image-level alignment (ImgAlign) and a combined Image-Feature Alignment (ImgFeatAlign). These visualizations highlight how temporal and modality-specific information is integrated or aligned between different model variants to improve temporal consistency and predictive performance.

\section{MammoRegNet}
\label{app1}

\subsection{Architectural Details}
\begin{figure}[t]
    \centering
\includegraphics[width=1.0\textwidth]{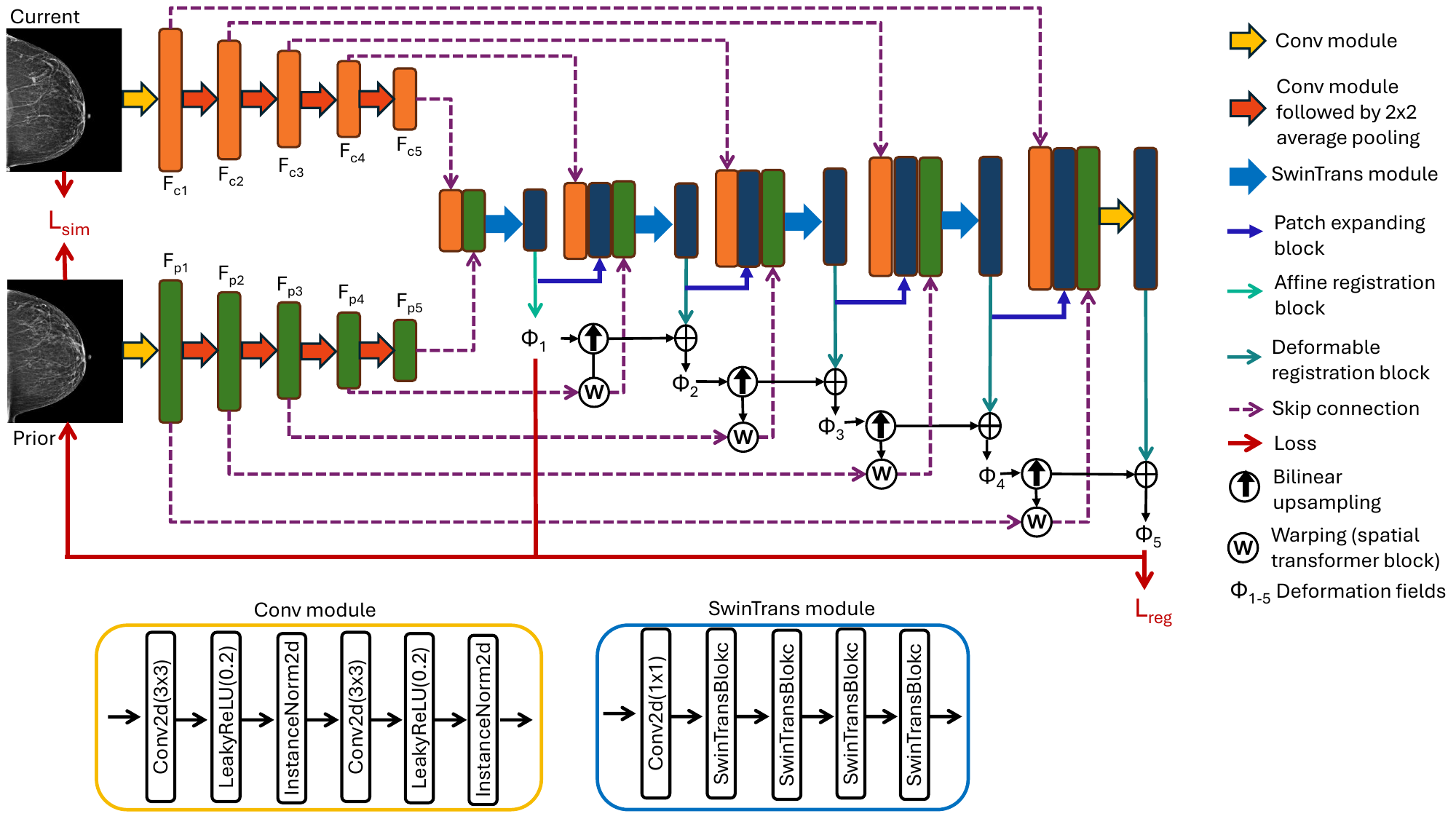}
    \caption{Architecture of the proposed MammoRegNet.}
    \label{fig:mammoreg}
\end{figure}

MammoRegNet performs both affine and deformable registration within a single unified network, leveraging an encoder-decoder structure. 

The encoder consists of two parallel paths, each designed to independently learn features from the current (fixed) and prior (moving) mammograms. Both paths consist of five convolutional modules with shared weights. Each convolutional module follows a consistent structure: a convolutional layer with a kernel size of $3\times 3$, followed by a Leaky ReLU activation function with a negative slope of 0.2, an instance normalization layer, another convolutional layer with a kernel size of $3\times 3$, a second Leaky ReLU activation with a slope of 0.2, and a final instance normalization layer. To progressively reduce spatial resolution while preserving essential features, average pooling with a $2 \times 2$ kernel is applied after the second, third, fourth, and fifth convolutional modules.

The features of the current and prior mammograms are then concatenated and passed to the decoder, which comprises four SwinTrans modules and one convolutional module. Each SwinTrans module includes a convolutional layer and four SwinTrans blocks~\citep{swintransblock}, with patch-expanding blocks between adjacent modules to enhance spatial resolution. The decoder performs the registration in a coarse-to-fine manner consisting of five steps, starting with an affine registration followed by four steps of deformable refinements. At each step, the output of the decoder module is input to either an affine or deformable registration block, which maps the features to a displacement field. This iterative process results in progressively finer registration with each decoder step, effectively enabling joint affine and deformable registration.

\subsection{Results}
Table~\ref{tab:mammoregnet2} summarizes the performance of the image-based registration model, MammoRegNet, on the EMBED and CSAW-CC datasets. The results demonstrate a marked improvement in image similarity, measured by Normalized Cross-Correlation, across all registration stages, from the initial input through affine and final deformable registration. This progression highlights the model’s ability to achieve accurate spatial alignment of mammography images. Moreover, the Percentage of Negative Jacobian Determinants remains exceptionally low across both datasets, indicating that MammoRegNet produces smooth and anatomically plausible deformation fields. Together, these findings underscore the model's effectiveness in delivering high-quality, reliable mammography image registration.

\begin{figure}[H]
     \centering
    \begin{subfigure}{0.15\textwidth} 
     \includegraphics[width=\textwidth]{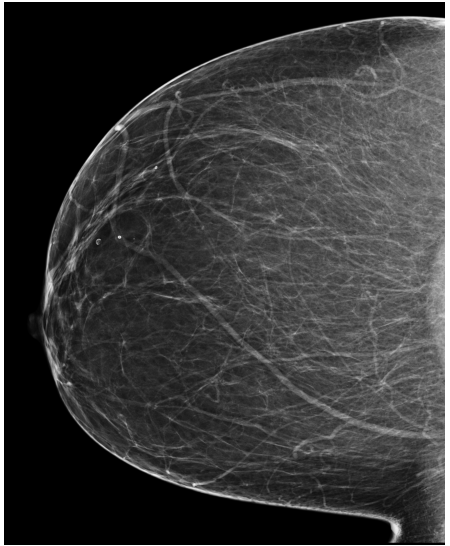}
    \end{subfigure}
    \begin{subfigure}{0.15\textwidth}
        \includegraphics[width=\textwidth]{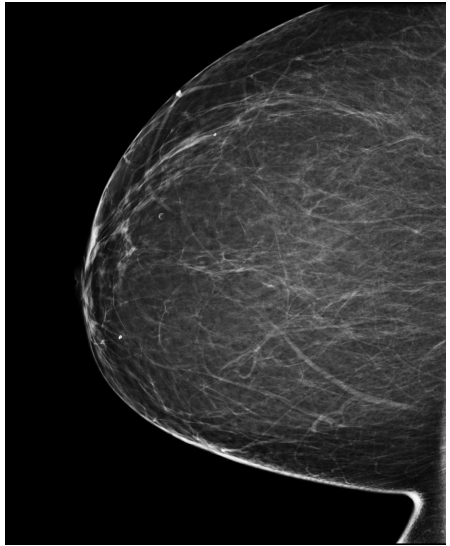}
    \end{subfigure}
    \begin{subfigure}{0.15\textwidth}
        \includegraphics[width=\textwidth]{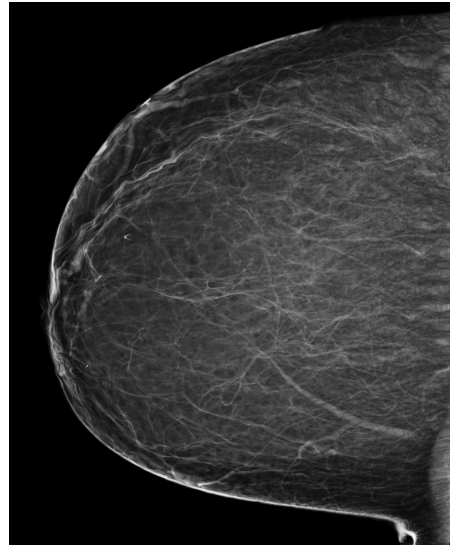}
    \end{subfigure}
    \begin{subfigure}{0.15\textwidth}
        \includegraphics[width=\textwidth]{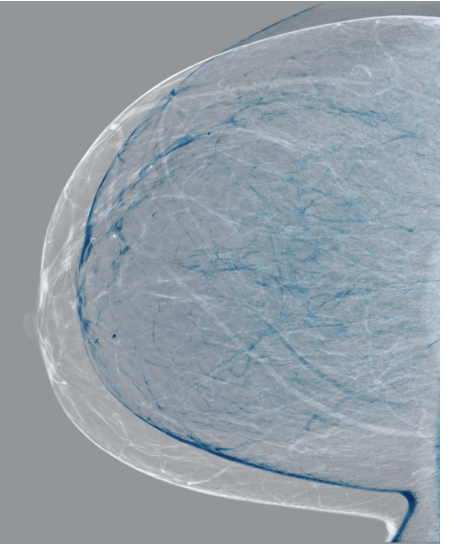}
    \end{subfigure}
    \begin{subfigure}{0.15\textwidth}
        \includegraphics[width=\textwidth]{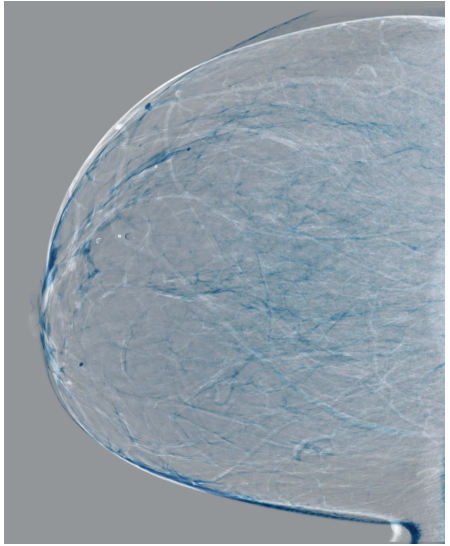}
    \end{subfigure}
    \begin{subfigure}{0.15\textwidth}
        \includegraphics[width=\textwidth]{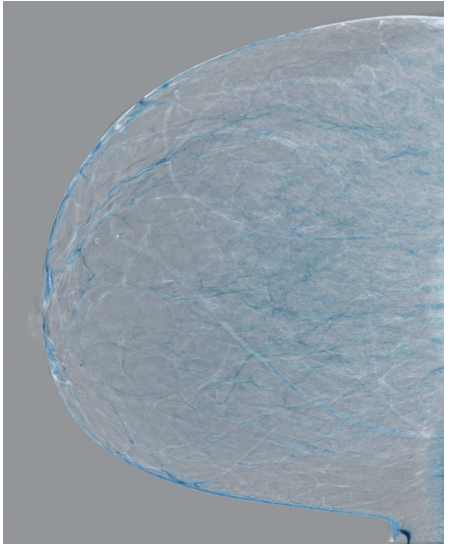}
    \end{subfigure} \\
     \begin{subfigure}{0.15\textwidth}
        \includegraphics[width=\textwidth]{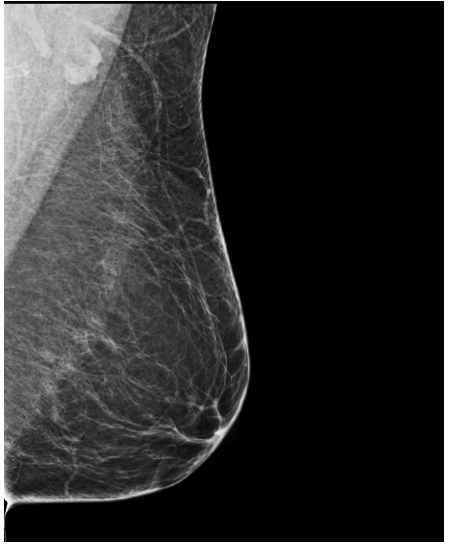}
    \end{subfigure}
      \begin{subfigure}{0.15\textwidth}
        \includegraphics[width=\textwidth]{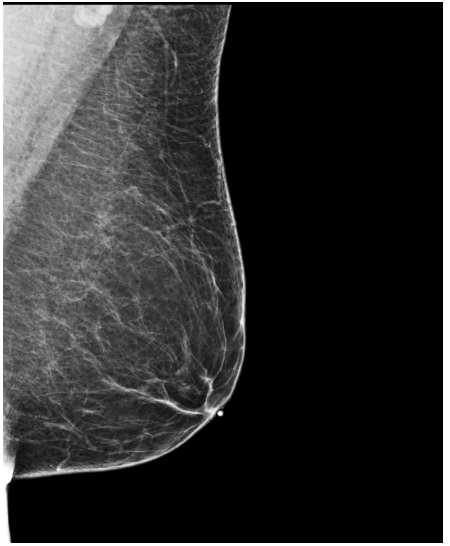}
    \end{subfigure}
      \begin{subfigure}{0.15\textwidth}
        \includegraphics[width=\textwidth]{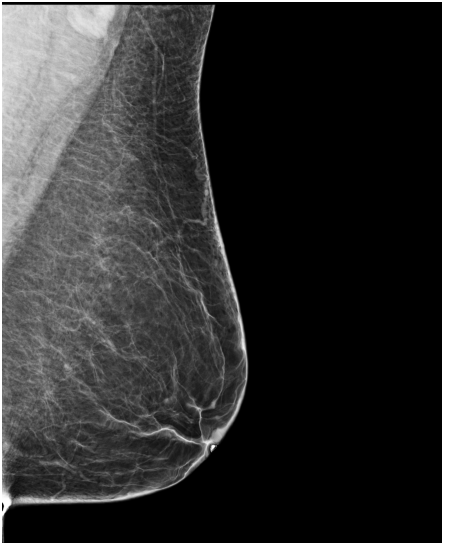}
    \end{subfigure}
      \begin{subfigure}{0.15\textwidth}
        \includegraphics[width=\textwidth]{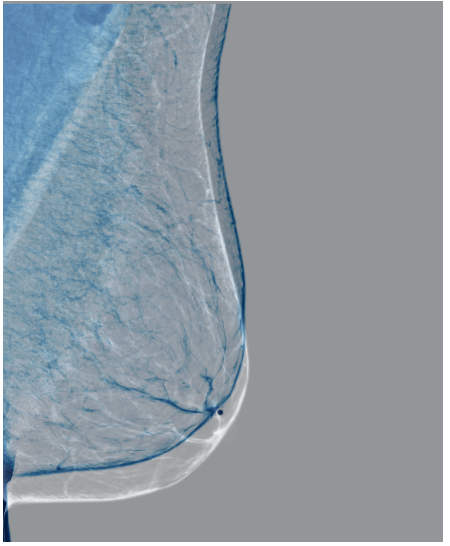}
    \end{subfigure}
      \begin{subfigure}{0.15\textwidth}
        \includegraphics[width=\textwidth]{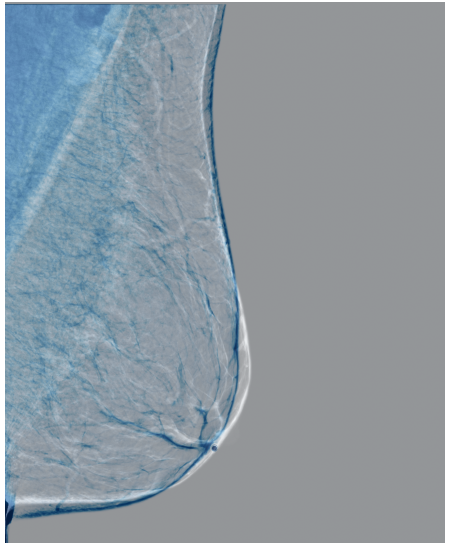}
    \end{subfigure}
      \begin{subfigure}{0.15\textwidth}
        \includegraphics[width=\textwidth]{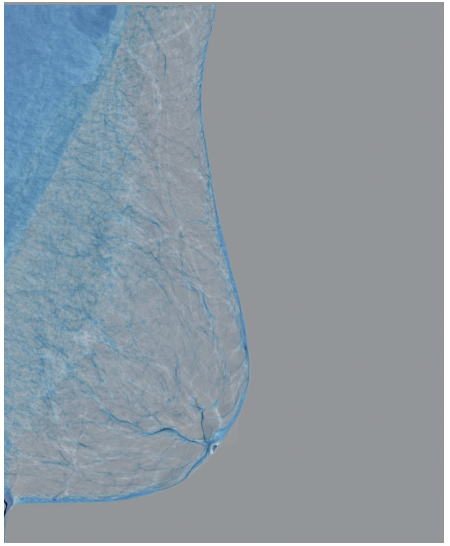}
    \end{subfigure} \\
    \begin{subfigure}{0.15\textwidth}
        \includegraphics[width=\textwidth]{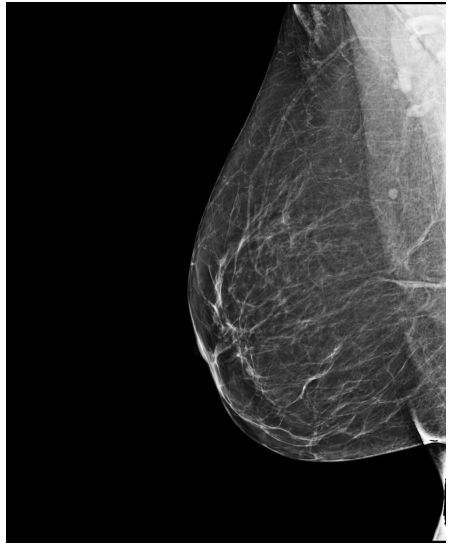}
    \end{subfigure}
      \begin{subfigure}{0.15\textwidth}
        \includegraphics[width=\textwidth]{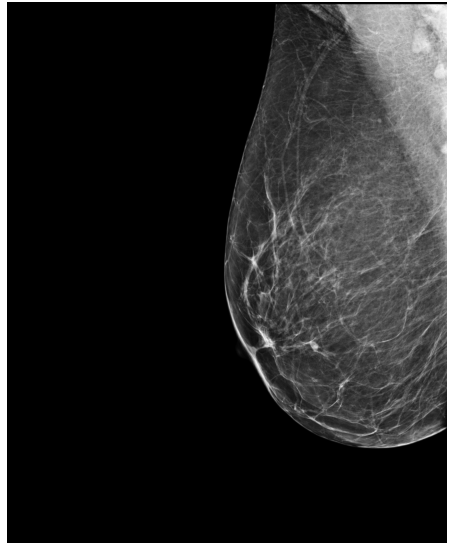}
    \end{subfigure}
      \begin{subfigure}{0.15\textwidth}
        \includegraphics[width=\textwidth]{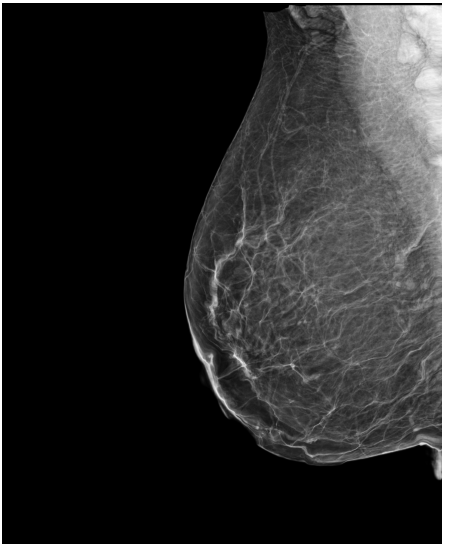}
    \end{subfigure}
      \begin{subfigure}{0.15\textwidth}
        \includegraphics[width=\textwidth]{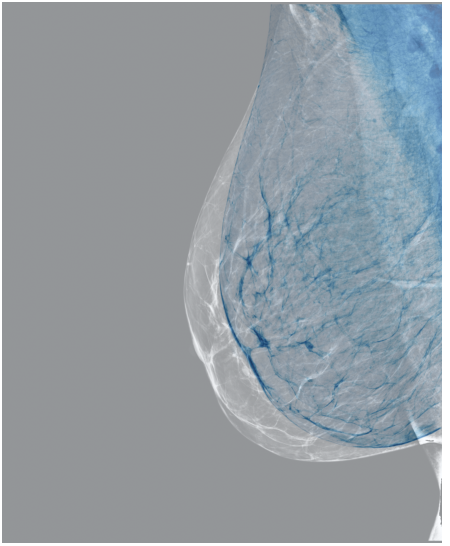}
    \end{subfigure}
      \begin{subfigure}{0.15\textwidth}
        \includegraphics[width=\textwidth]{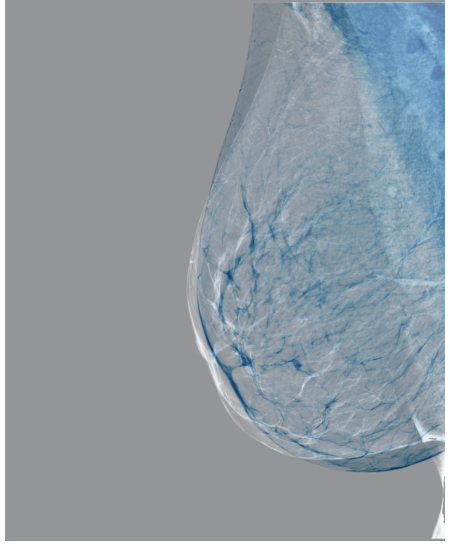}
    \end{subfigure}
      \begin{subfigure}{0.15\textwidth}
        \includegraphics[width=\textwidth]{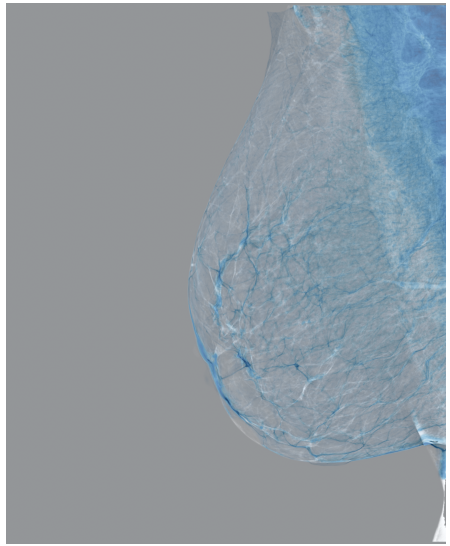}
    \end{subfigure} \\
    \begin{subfigure}{0.15\textwidth}
        \includegraphics[width=\textwidth]{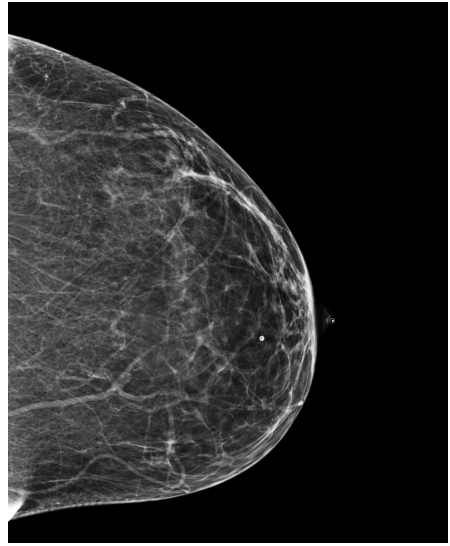}
           \caption{}
    \end{subfigure}
      \begin{subfigure}{0.15\textwidth}
        \includegraphics[width=\textwidth]{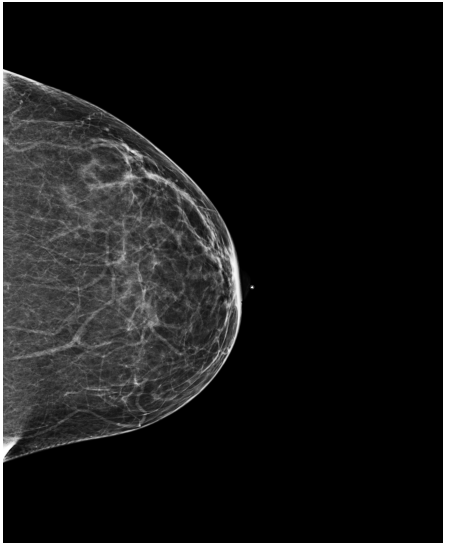}
                \caption{}
    \end{subfigure}
      \begin{subfigure}{0.15\textwidth}
        \includegraphics[width=\textwidth]{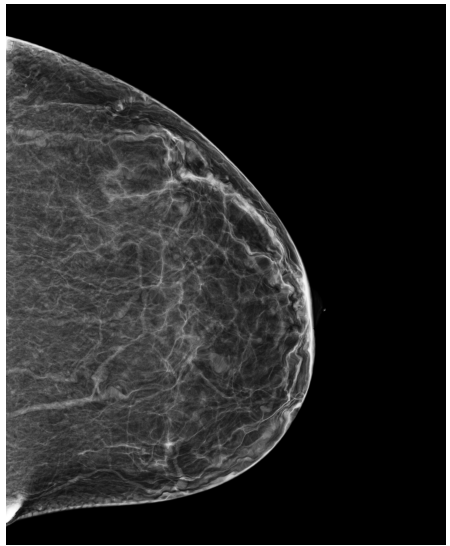}
            \caption{}    
    \end{subfigure} 
      \begin{subfigure}{0.15\textwidth}
        \includegraphics[width=\textwidth]{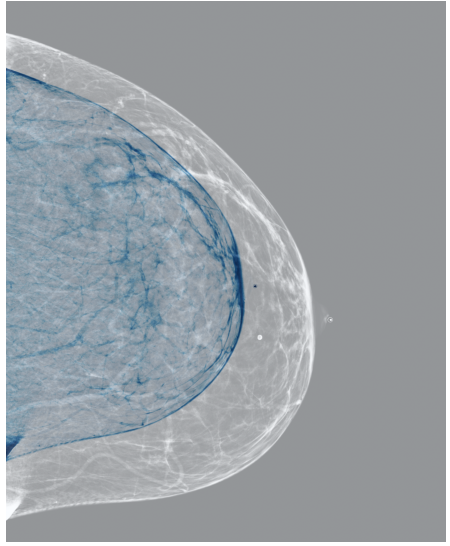}
            \caption{}    
    \end{subfigure}
      \begin{subfigure}{0.15\textwidth}
        \includegraphics[width=\textwidth]{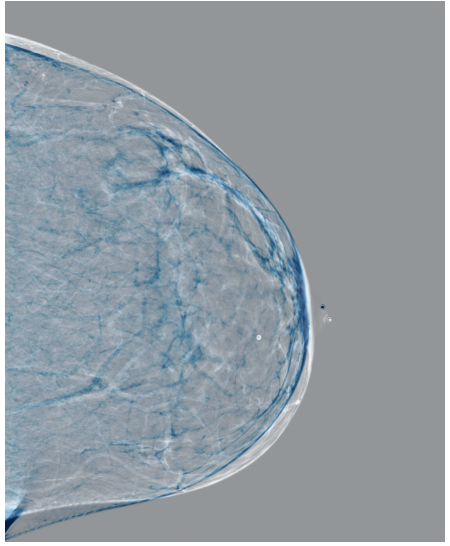}
            \caption{}    
    \end{subfigure}
      \begin{subfigure}{0.15\textwidth}
        \includegraphics[width=\textwidth]{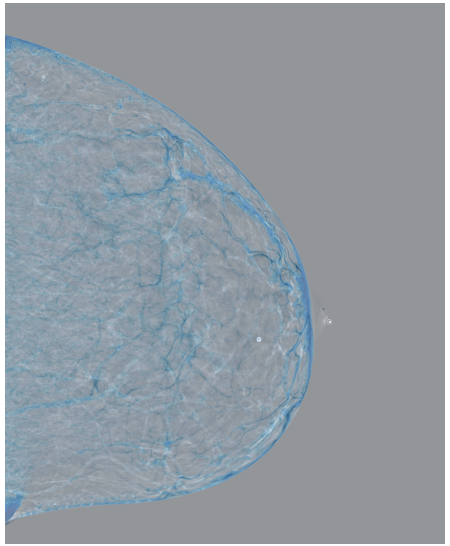}
            \caption{}    
    \end{subfigure}
     \caption{Registration result of MammoRegNet. Visual results of the MammoRegNet registration performance for patients without cancer.  (a) Current mammogram, (b) prior mammogram, (c) aligned prior mammogram, (d) current mammogram overlaid with the prior mammogram, (e) current mammogram overlaid with the affine-transformed prior mammogram, and (f) current mammogram overlaid with the final transformed prior mammogram.}
     \label{fig:regmodel1}
 \end{figure}

\begin{figure}[H]
     \centering
    \begin{subfigure}{0.15\textwidth}
        \includegraphics[width=\textwidth]{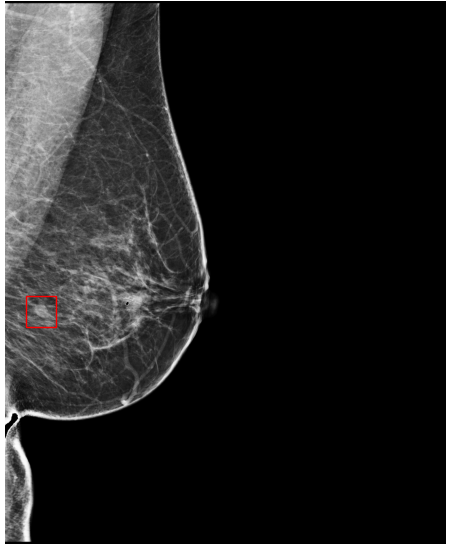}
    \end{subfigure}
      \begin{subfigure}{0.15\textwidth}
        \includegraphics[width=\textwidth]{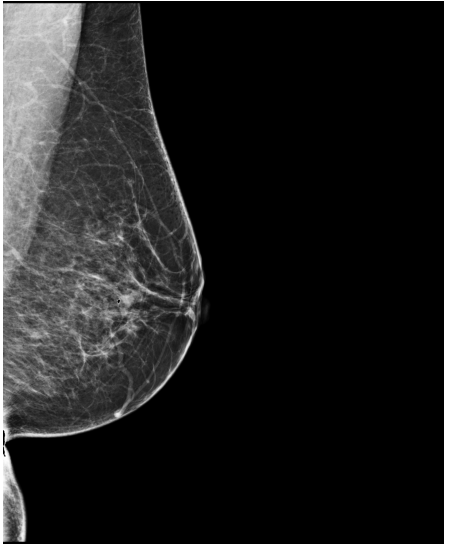}
    \end{subfigure}
      \begin{subfigure}{0.15\textwidth}
        \includegraphics[width=\textwidth]{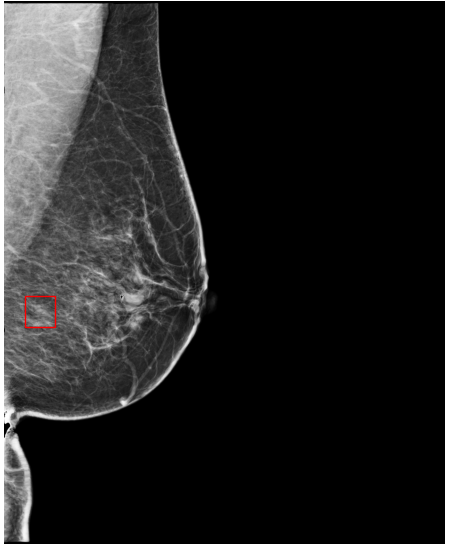}
    \end{subfigure}
      \begin{subfigure}{0.15\textwidth}
        \includegraphics[width=\textwidth]{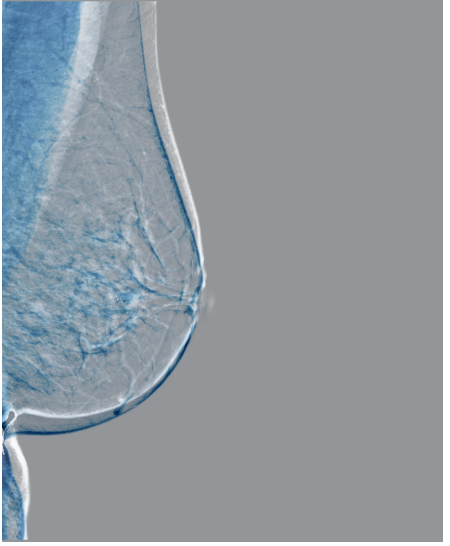}
    \end{subfigure}
      \begin{subfigure}{0.15\textwidth}
        \includegraphics[width=\textwidth]{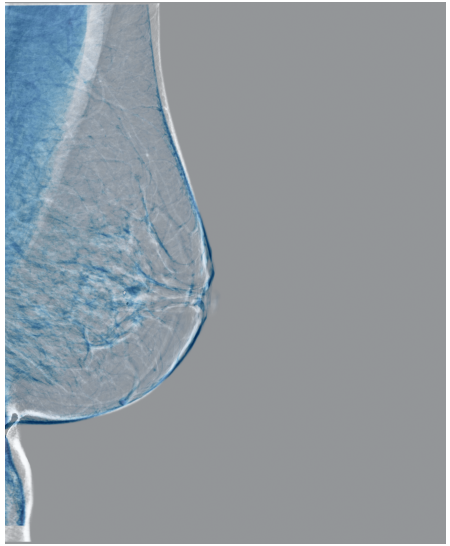}
    \end{subfigure}
      \begin{subfigure}{0.15\textwidth}
        \includegraphics[width=\textwidth]{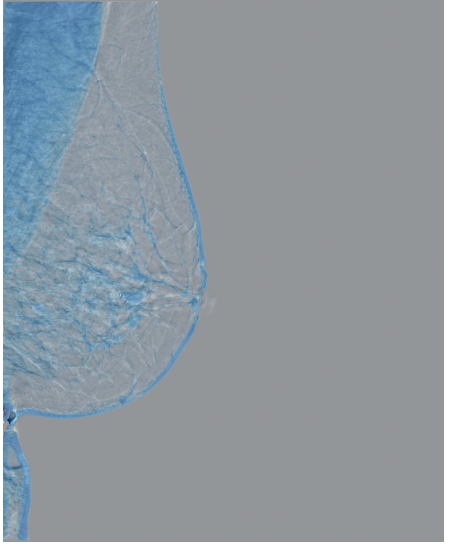}
    \end{subfigure} \\
    \begin{subfigure}{0.15\textwidth}
        \includegraphics[width=\textwidth]{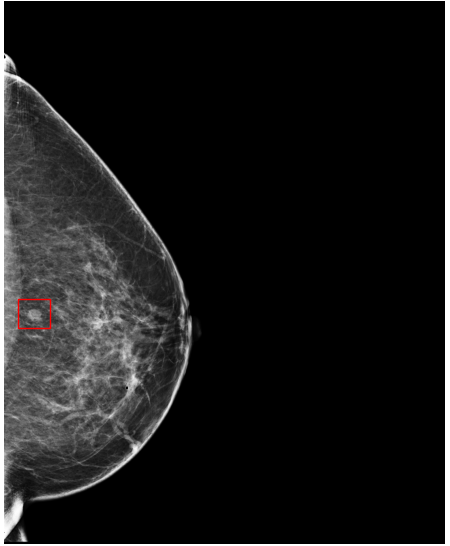}
    \end{subfigure}
      \begin{subfigure}{0.15\textwidth}
        \includegraphics[width=\textwidth]{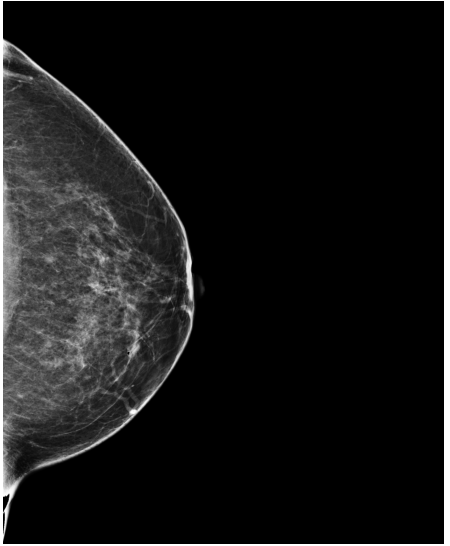}
    \end{subfigure}
      \begin{subfigure}{0.15\textwidth}
        \includegraphics[width=\textwidth]{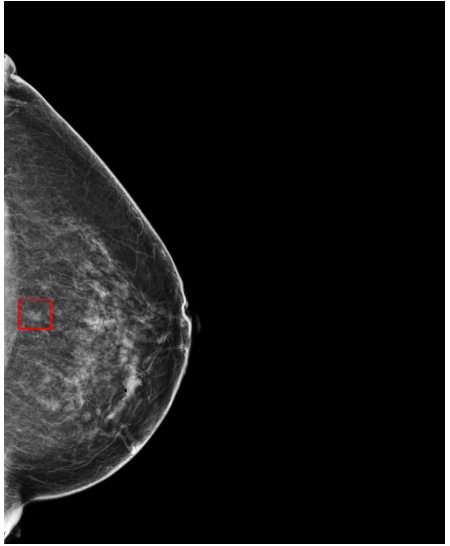}
    \end{subfigure}
      \begin{subfigure}{0.15\textwidth}
        \includegraphics[width=\textwidth]{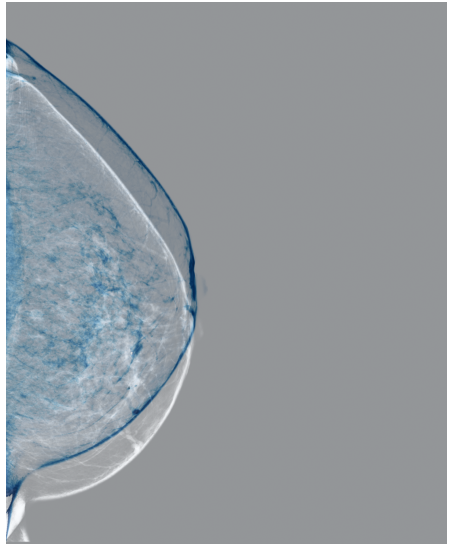}
    \end{subfigure}
      \begin{subfigure}{0.15\textwidth}
        \includegraphics[width=\textwidth]{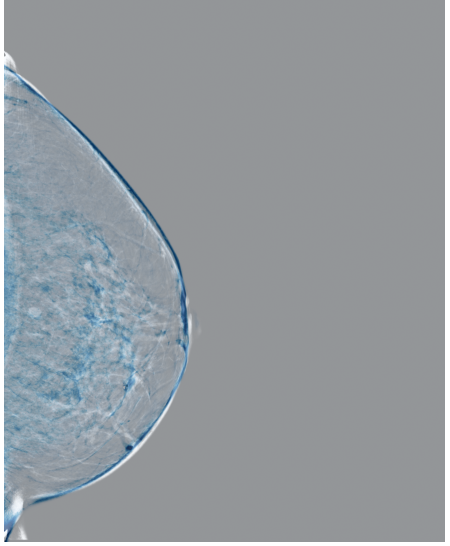}
    \end{subfigure}
      \begin{subfigure}{0.15\textwidth}
        \includegraphics[width=\textwidth]{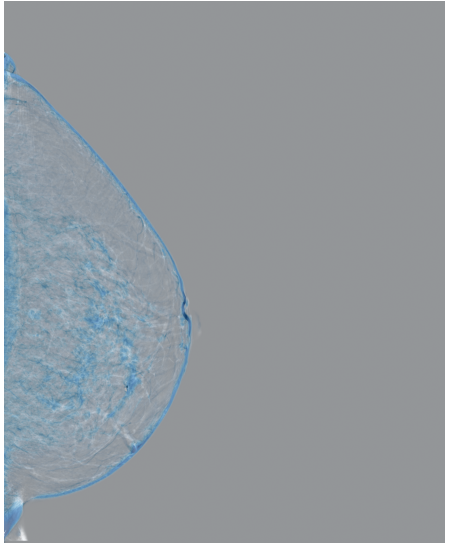}
    \end{subfigure} \\
     \begin{subfigure}{0.15\textwidth}
        \includegraphics[width=\textwidth]{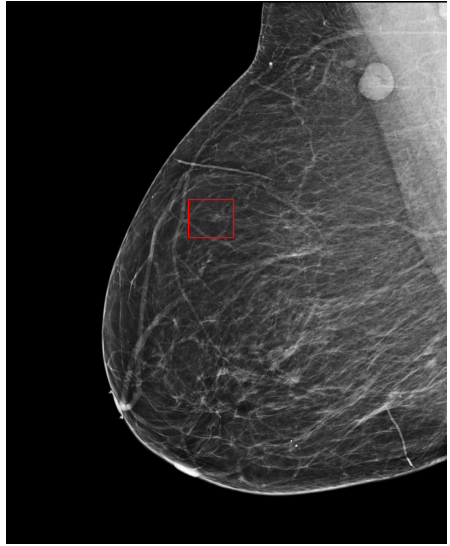}
    \end{subfigure}
      \begin{subfigure}{0.15\textwidth}
        \includegraphics[width=\textwidth]{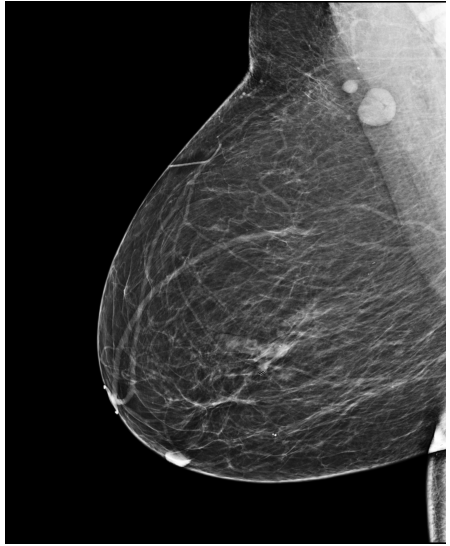}
    \end{subfigure}
      \begin{subfigure}{0.15\textwidth}
        \includegraphics[width=\textwidth]{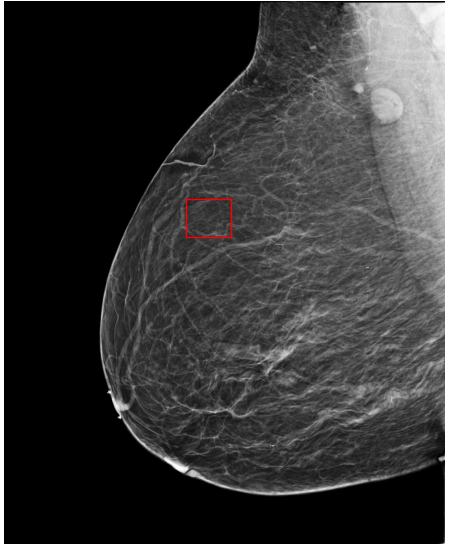}
    \end{subfigure}
      \begin{subfigure}{0.15\textwidth}
        \includegraphics[width=\textwidth]{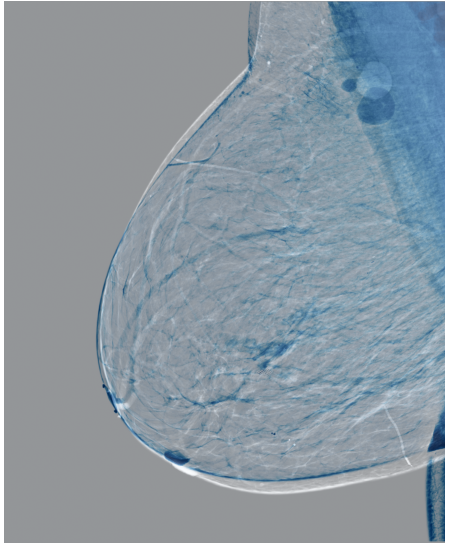}
    \end{subfigure}
      \begin{subfigure}{0.15\textwidth}
        \includegraphics[width=\textwidth]{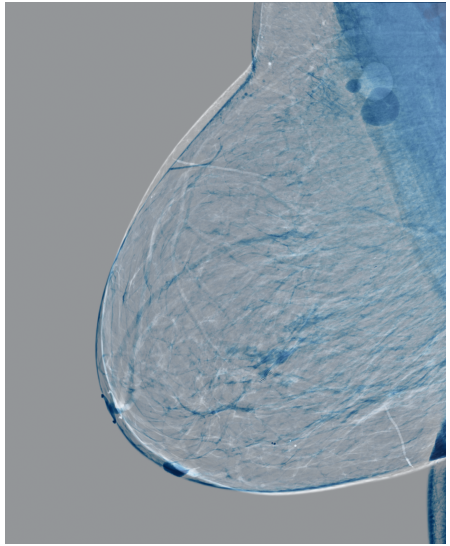}
    \end{subfigure}
      \begin{subfigure}{0.15\textwidth}
        \includegraphics[width=\textwidth]{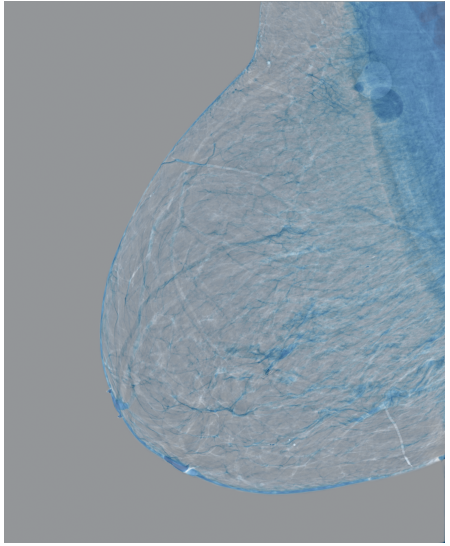}
    \end{subfigure} \\ 
     \begin{subfigure}{0.15\textwidth}
        \includegraphics[width=\textwidth]{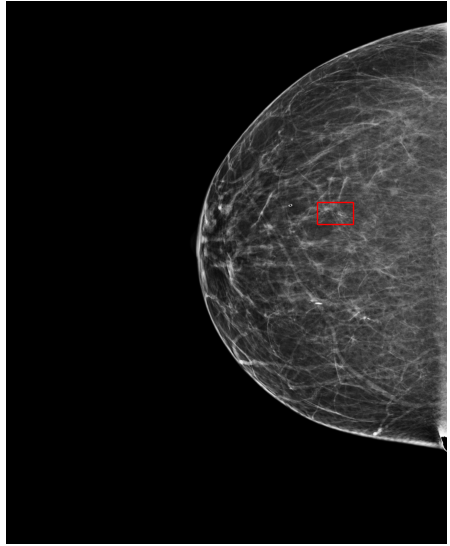}
    \caption{}
    \end{subfigure}
      \begin{subfigure}{0.15\textwidth}
        \includegraphics[width=\textwidth]{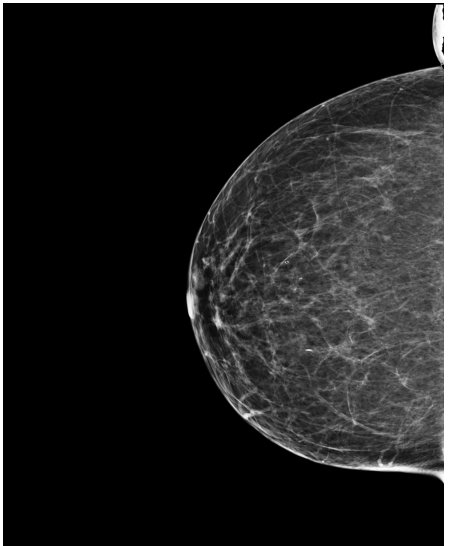}
     \caption{}
    \end{subfigure}
      \begin{subfigure}{0.15\textwidth}
        \includegraphics[width=\textwidth]{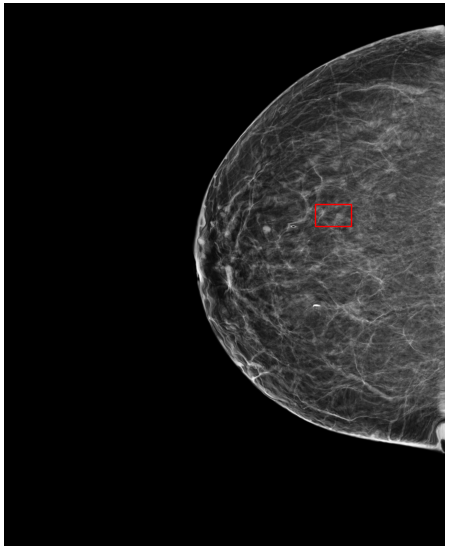}
            \caption{}    
    \end{subfigure}
      \begin{subfigure}{0.15\textwidth}
        \includegraphics[width=\textwidth]{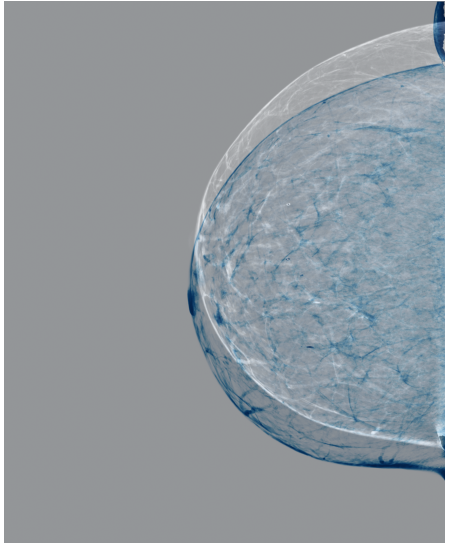}
            \caption{}    
    \end{subfigure}
      \begin{subfigure}{0.15\textwidth}
        \includegraphics[width=\textwidth]{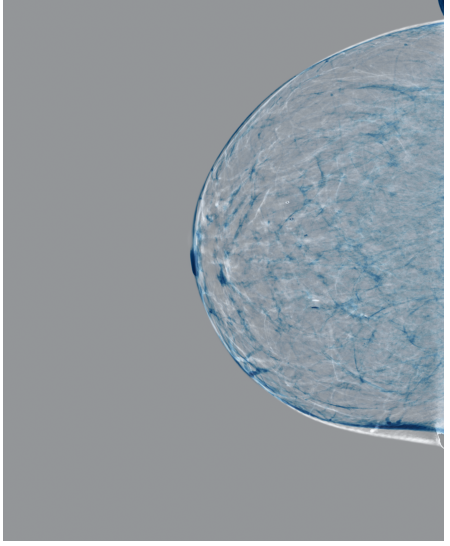}
        \caption{}    
    \end{subfigure}
      \begin{subfigure}{0.15\textwidth}
        \includegraphics[width=\textwidth]{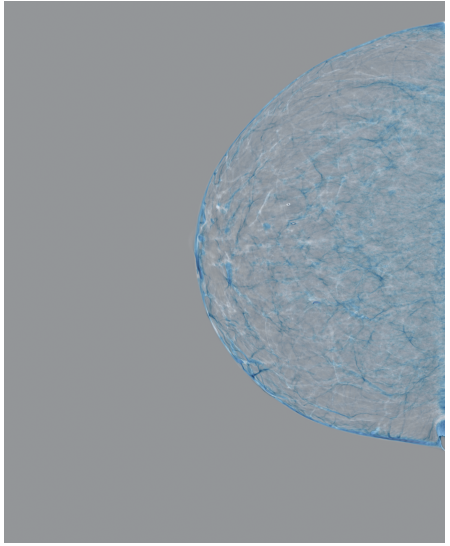}
             \caption{}   
    \end{subfigure}
  \caption{Registration result of MammoRegNet. Visual results of the MammoRegNet registration performance for patients with cancer. The tumor region is highlighted by a bounding box on the current mammogram and the aligned prior mammogram.  (a) Current mammogram, (b) prior mammogram, (c) aligned prior mammogram, (d) current mammogram overlaid with the prior mammogram, (e) current mammogram overlaid with the affine-transformed prior mammogram, and (f) current mammogram overlaid with the final transformed prior mammogram.}
     \label{fig:regmodel2}
 \end{figure}

Figures~\ref{fig:regmodel1} and~\ref{fig:regmodel2} demonstrate the registration performance of MammoRegNet across a variety of cases, including examples from patients with and without cancer. For the cancerous images, the location of the malignancy is indicated by a radiologist-annotated region of interest (ROI). In each example, the model successfully aligns the prior mammogram to the current view using a combination of affine and deformable transformations. The aligned prior mammograms (c) closely match the anatomical structures of the current mammograms (a), demonstrating MammoRegNet’s precision in spatial alignment. Overlay visualizations further reveal the alignment process: substantial misalignment is evident in the initial overlay (d), partially corrected with affine registration (e), and fully refined through deformable registration (f). Notably, MammoRegNet performs consistently across both MLO and CC views and for both left and right breasts, confirming its robustness across diverse clinical imaging scenarios. These results highlight MammoRegNet’s capacity to produce precise, anatomically consistent alignments, enabling high-fidelity image registration for longitudinal mammography analysis.

 \begin{table}[h]
\centering
\scriptsize
\caption{MammoRegNet registration performance (95\% CI in parentheses).}
\label{tab:mammoregnet2}
\begin{adjustbox}{center}
\begin{tabular}{lcccc}
\toprule
\textbf{Dataset} & \textbf{NCC before (\%) $\uparrow$} & \textbf{NCC affine (\%) $\uparrow$} & \textbf{NCC final (\%) $\uparrow$} & \textbf{NJD (\%) $\downarrow$} \\
\midrule
EMBED & \makecell{70.7 \\ (70.1–71.3)} & \makecell{78.7 \\ (78.3–79.1)} & \makecell{92.2 \\ (92.0–92.4)} & \makecell{0.0013 \\ (0.0012–0.0015)} \\
CSAW-CC & \makecell{76.0 \\ (75.3–76.6)} & \makecell{81.6 \\ (81.1–82.0)} & \makecell{92.8 \\ (92.5–93.0)} & \makecell{0.0008 \\ (0.0006–0.0009)} \\
\bottomrule
\end{tabular}
\end{adjustbox}
\end{table}

\end{document}